%% file: 00_main.tex
\documentclass{article}

 \usepackage[preprint]{neurips_2026}

\usepackage[utf8]{inputenc} % allow utf-8 input
\usepackage[T1]{fontenc}    % use 8-bit T1 fonts
\usepackage{hyperref}       % hyperlinks
\usepackage{url}            % simple URL typesetting
\usepackage{booktabs}       % professional-quality tables
\usepackage{amsfonts}       % blackboard math symbols
\usepackage{nicefrac}       % compact symbols for 1/2, etc.
\usepackage{microtype}      % microtypography
\usepackage{xcolor}         % colors

\usepackage{url}
\usepackage{graphicx}
\usepackage{float} 
\usepackage{subfigure}
\usepackage{hyperref}
\usepackage{makecell}
\usepackage{colortbl}
\usepackage{enumitem}
\usepackage{multirow}
\usepackage{multicol}
\usepackage{appendix}
\usepackage{fontawesome5}
\usepackage{tcolorbox}
\usepackage[table,dvipsnames]{xcolor}
\usepackage{pifont}
\usepackage{algorithm}
\usepackage{algorithmic}
\usepackage{wrapfig}

\usepackage{amsmath}
\usepackage{amssymb}
\usepackage{mathtools}
\usepackage{amsthm}

\definecolor{cite}{HTML}{6494EA}

\hypersetup{
    hypertex=true,
    colorlinks=true,
    linkcolor=red,
    citecolor=cite,
}

\theoremstyle{plain}
\newtheorem{theorem}{Theorem}[section]
\newtheorem{proposition}[theorem]{Proposition}
\newtheorem{lemma}[theorem]{Lemma}

\theoremstyle{definition}
\newtheorem{definition}[theorem]{Definition}

\theoremstyle{remark}

  \newcommand{\worldcuplogo}{\raisebox{-0.42em}{\includegraphics[height=1.5em]{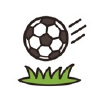}}~\hspace{-10pt}~\texttt{}}
  \title{\worldcuplogo{} WorldCup Sampling for Multi-bit LLM Watermarking}

\author{Yidan Wang\textsuperscript{1,2}, Yubing Ren\textsuperscript{1,2}\thanks{Corresponding author.}, Yanan Cao\textsuperscript{1,2}, Li Guo\textsuperscript{1,2} \\
\textsuperscript{1}Institute of Information Engineering, Chinese Academy of Sciences, Beijing, China \\ 
\textsuperscript{2}School of Cyber Security, University of Chinese Academy of Sciences, Beijing, China \\
\texttt{\{wangyidan, renyubing\}@iie.ac.cn}}

\begin{document}

\maketitle

\begin{abstract}
As large language models (LLMs) generate increasingly human-like text, watermarking has emerged as a promising solution for reliable attribution beyond mere detection. While multi-bit watermarking enables richer provenance encoding, existing approaches typically extend zero-bit watermarking schemes by introducing static logit perturbations and counting-based decoding strategies, which can degrade text quality and compromise decoding robustness as the payload increases. In this paper, we propose WorldCup, a multi-bit watermarking framework for LLMs that models the sampling process as a structured communication channel and embeds message bits through a hierarchical competition mechanism guided by complementary signals. Moreover, WorldCup incorporates entropy-aware modulation to preserve generation quality and enables robust message recovery via confidence-aware decoding that accounts for token-level reliability. Comprehensive experiments demonstrate that WorldCup achieves a strong balance across message capacity, detectability, robustness, text quality, and decoding efficiency, consistently outperforming prior baselines. We believe that this work establishes a scalable and principled foundation for future research on multi-bit watermarking in LLMs.
\end{abstract}

\input{01_Introduction}

\input{02_Preliminaries}

\input{03_Methodology}

\input{04_Experiments}

\input{05_Analysis}

\input{06_Conclusion}

\bibliography{reference}
\bibliographystyle{plain}

%%%%%%%%%%%%%%%%%%%%%%%%%%%%%%%%%%%%%%%%%%%%%%%%%%%%%%%%%%%%

\appendix

\input{07_Appendix.tex}

%%%%%%%%%%%%%%%%%%%%%%%%%%%%%%%%%%%%%%%%%%%%%%%%%%%%%%%%%%%%

\newpage

\end{document}

%% file: 01_Introduction.tex
\section{Introduction}

Large language models (LLMs)~\citep{guo2025deepseek, comanici2025gemini, achiam2023gpt} have shown remarkable performance across a wide range of real-world applications, including creative writing, code generation and AI agent~\citep{luo2025large}, rendering LLM-generated text progressively indistinguishable from human-written text~\citep{mitchell2023detectgpt}. While these advances significantly boost productivity, they also amplify serious risks such as misinformation dissemination, academic plagiarism, and phishing attacks~\citep{tang2024science}. As a consequence, LLM watermarking has emerged as a promising technique to mitigate these concerns by embedding imperceptible yet verifiable signals into generated text, thereby enabling reliable attribution, detection and traceability~\citep{bengio2025international}.

Inference-time LLM watermarking can be broadly categorized into zero-bit and multi-bit schemes based on embedding capacity~\citep{liu2024survey}. While zero-bit watermarking focuses solely on detection, multi-bit watermarking further enables the extraction of rich metadata, such as model identity or generation timestamps. The prevailing multi-bit watermarking paradigm builds upon zero-bit schemes by implicitly embedding message into the stochastic generation process. Typically, a pseudo-random seed is derived by hashing a predefined secret key with the context and is associated with a target bit string to steer an underlying zero-bit watermarking mechanism, thereby establishing a verifiable link between the message and the generated text. Decoding reverses this process by aggregating token-level evidence to identify the message that best aligns with the expected watermark signal.

However, despite their effectiveness, existing multi-bit watermarking still inherits key limitations from their zero-bit foundations. In particular, most approaches rely on static logit perturbations to inject watermark signals, resulting in an implicit and weakly controlled interaction with the original model distribution. Such a design can interfere with the natural generation dynamics, especially at higher payloads, thereby degrading text quality and fluency. Meanwhile, these methods typically adopt counting-based decoding, where tokens are discretely mapped to binary outcomes and contribute equally. This creates a mismatch between soft probabilistic embedding and hard aggregation, leading to suboptimal trade-offs between message capacity and robustness.

To overcome these limitations, we revisit multi-bit watermarking from a more direct and principled perspective. Rather than relying on static logit biasing, we treat the sampling process itself as a structured communication channel and integrate message encoding more tightly into token selection. This perspective is loosely inspired by Google’s SynthID Text~\citep{dathathri2024scalable}, which demonstrates that token-ranking perturbations can leave statistically verifiable traces without degrading text quality, albeit in a detection-only setting. Building on this insight, we extend the idea to a full-fledged multi-bit regime and introduce a multi-round, hierarchical sampling strategy that couples message embedding with the competitive selection of candidate tokens.

Specifically, we propose \textbf{WorldCup}, a multi-bit watermarking framework for LLMs that leverages the inherent redundancy of autoregressive token generation to support robust and scalable information embedding. At the \textbf{encoding} stage, WorldCup conceptualizes watermarking as a structured competition process, in which multiple complementary signals jointly guide token selection and induce stable statistical separation, while preserving sufficient flexibility for high-quality natural language generation. To balance detectability and fluency, the framework further incorporates an entropy-aware modulation mechanism that adaptively adjusts watermark strength in response to local uncertainty in the model output distribution. At the \textbf{decoding} stage, WorldCup moves beyond simple counting-based detection and adopts a soft confidence-aware aggregation strategy that weights token-level evidence by its statistical reliability, thereby mitigating the disproportionate influence of low-entropy tokens. These design choices enable fine-grained control over the divergence between clean and watermarked distributions, and become increasingly advantageous as the embedded payload grows, yielding substantial improvements in decoding accuracy and efficiency at scale.

To this end, comprehensive experiments across multiple LLMs and downstream tasks demonstrate that WorldCup consistently outperforms prior baselines, delivering a strong and well-balanced trade-off among multi-bit capacity, watermark detectability, robustness, text quality, and decoding efficiency. In summary, our contributions are threefold:

\begin{itemize}[leftmargin=*]
    \item We propose WorldCup, a versatile multi-bit watermarking framework that is rigorously validated through both theoretical analysis and extensive empirical evaluation.
    \item We introduce a confidence-aware decoding paradigm that moves beyond prior counting-based detectors, substantially improving both decoding accuracy and efficiency.
    \item We conduct a systematic analysis of key design components, examining different hyperparameter choices and offering actionable insights to guide future research.
\end{itemize}

%% file: 02_Preliminaries.tex
\section{Preliminaries}
\paragraph{Notations.} Consider an autoregressive LLM $\Theta$ based on transformer~\citep{vaswani2017attention}, let $\mathcal{V}$ denote the vocabulary set of all tokens with size $|\mathcal{V}|$. Given a prompt $p_0$ and previously contextual tokens $\mathbf{x}_{<t}=(x_1,x_2,\ldots,x_{t-1})$, the LLM generates an imminent token $x_t\in\mathcal{V}$ sequentially from the conditional distribution $P_\Theta(x_t\mid p_0,\mathbf{x}_{<t}) \in\Delta_{\mathcal{V}}$. The process repeats until either a predefined maximum length is reached or an end-of-sequence token is produced. 

\subsection{SynthID-Text Zero-bit Watermarking}

\paragraph{Watermark Embedding.} For zero-bit watermarking SynthID-Text~\citep{dathathri2024scalable}, at each generation step $t$, a random seed $r_t$ is derived by hashing the previously generated text together with a secret watermark key $\xi \in \Xi$. The seed is then provided to $m$ independent pseudo-random functions (PRFs) $\mathbf{g} = (g_1, \ldots, g_m)$~\citep{goldreich1986construct}, each of which assigns a binary value $g_\ell(x_i, r_t) \in \{0,1\}$ to every token $x_i \in \mathcal{V}$, where $\ell\in\{1,\ldots,m\}$. Token selection proceeds via an $m$-layer tournament-based sampling scheme. Concretely, $N^m$ candidate tokens are independently sampled from the model distribution $P_{\Theta}(\cdot \mid p_0, \mathbf{x}_{<t})$ to form the leaves of a complete $N$-ary tournament tree of depth $m$. At each layer $\ell$, candidates are grouped in sets of $N$ and compared using the corresponding PRF scores $g_\ell(\cdot,r_t)$. In practice, $N$ is set to $2$, which corresponds to pairwise comparisons. The highest-scoring token in each group advances to the next layer, with ties broken at random. This elimination process continues until ultimately a single token remains, which is emitted as the output token $x_t$.

\begin{figure*}[tp]
\centering
\includegraphics[width=\linewidth]{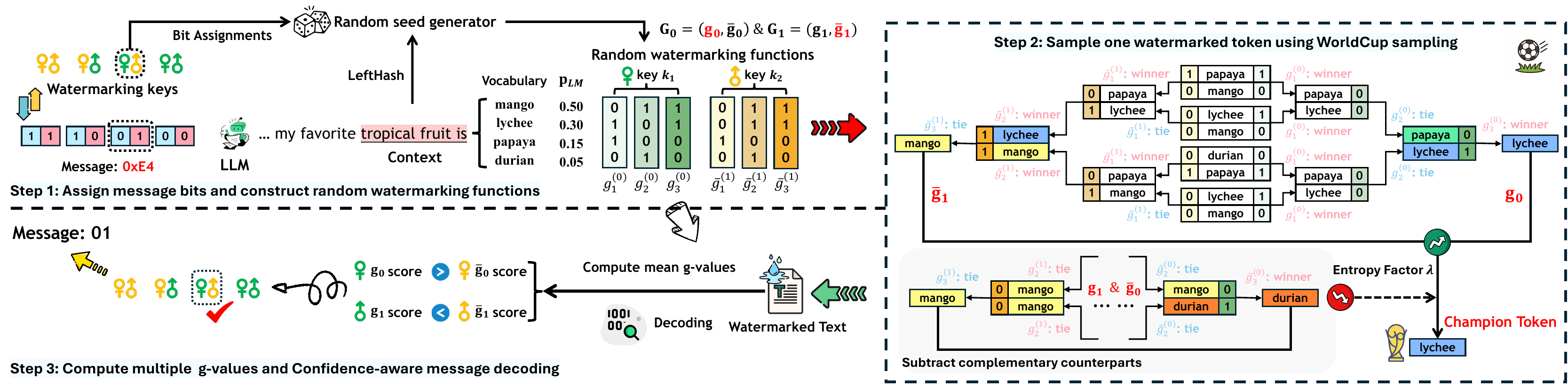}
\caption{An overview of our multi-bit watermarking framework \textbf{WorldCup} for LLMs.}
\label{Fig: WorldCup_Framework}
\end{figure*}

\paragraph{Watermark Detection.} By design, tournament sampling biases the generation process toward tokens that attain higher scores under the watermarking functions $\mathbf{g}$. Given a generated text $\mathbf{y} = (y_1,y_2,\ldots, y_T)$, watermark detection measures how well the text $\mathbf{y}$ aligns with the functions $\mathbf{g}$. Formally, the detection statistic is defined as the average watermark score over the sequence:
\begin{equation}
    s(\mathbf{y};\mathbf{g}) = \frac{1}{mT} \sum_{t=1}^{T} \sum_{\ell=1}^{m} g_\ell(y_t, r_t). \label{Equation: mean g-value}
\end{equation}
Since watermarked tokens tend to yield higher $g$-values, watermarked texts attain higher scores $s(\mathbf{y};\mathbf{g})$ than unwatermarked text, enabling reliable statistical zero-bit watermark detection.

\subsection{MPAC Multi-bit Watermarking}
% Multi-bit Watermarking via Position Allocation (MPAC) \citep{yoo-etal-2024-advancing} extends zero-bit watermarking to the multi-bit setting by allocating generated tokens to message positions and encoding message bits through a classic zero-bit watermarking method KGW \citep{pmlr-v202-kirchenbauer23a}. 

\paragraph{Message Encoding.} Let $\mathbf{m}\in\sum^b$ be the message over an $s$-ary alphabet $\sum=\{0,\ldots,s-1\}$ and $b$ is its length. At each step $t$, MPAC uses a PRF with seed $r_t$ to sample a position $p\in\{0,\ldots,b-1\}$ and retrieves $\mathbf{m}[p]$. The vocabulary is randomly shuffled and partitioned into $s$ disjoint subsets $\mathcal{V}_t=[C_0,\ldots,C_{s-1}]$; the subset indexed by $\mathbf{m}[p]$ is treated as the green list and given a positive logits bias $\delta$, increasing its sampling probability and thus encoding the message symbol.

% Let $\mathbf{m}\in\sum^b$ denote the message to be embedded, where $\sum=\{0,\ldots,s-1\}$ is an $s$-ary alphabet (e.g. $s=2$ for binary messages) and $b$ is the message length. At each generation step $t$, MPAC employs a PRF with random seed $r_t$ to sample a message position $p\in\{0,\ldots,b-1\}$ and retrieves the corresponding message symbol $\mathbf{m}[p]$. Meanwhile, the vocabulary is randomly shuffled and partitioned into $s$ disjoint subsets $\mathcal{V}_t=[C_0,\ldots,C_{s-1}]$. According to the value of $\mathbf{m}[p]$, the corresponding subset is designated as the green list and is assigned a positive logits bias $\delta$. This bias increases the likelihood that LLM samples tokens from the designated subset, thereby embedding the message symbol during generation.

\paragraph{Message Decoding.} MPAC extracts the embedded multi-bit messages from text via a majority voting matrix $\mathbf{M}\in \mathbb{R}^{b\times s}$ initialized to zero. For each token $y_t$, the decoder reconstructs the message position $p$ and the vocabulary partitions $\mathcal{V}_t$ using the same PRF as in the encoding stage. If $y_t \in C_j$, the corresponding entry $\mathbf{M}[p][j]$ is incremented by one. After processing the entire text, each message symbol is recovered via $\hat{\mathbf{m}}[p]=\arg\max_{j\in\{0,\ldots,s-1\}}(\mathbf{M}[p][j])$. Under the assumption that watermarked text contains more green-list tokens than non-green-list tokens, the aggregate majority count $\sum_{p=0}^{b-1}\max_j(\mathbf{M}[p][j])$ serves as the total green tokens for computing the $z$-score statistic, which subsequently determines whether the text is watermarked.

%% file: 03_Methodology.tex
\section{Methodology}

\paragraph{Overview.} As illustrated in Fig.~\ref{Fig: WorldCup_Framework}, WorldCup framework comprises two central stages: message embedding (Section~\ref{Section: Binary WorldCup Watermarking}, Section~\ref{Section: Complementary $g$-value Functions} and Section~\ref{Section: Generalized WorldCup Watermarking}) and message decoding (Section~\ref{Section: Confidence-aware Message Decoding}). 

% We introduce the embedding methodology progressively. Section~\ref{Section: Binary WorldCup Watermarking} presents a binary instantiation that embeds one bit per token using arbitrary $g$-value functions. Section~\ref{Section: Complementary $g$-value Functions} shows that complementary $g$-value functions achieve theoretically optimal performance, forming the core components of the framework. Building on this insight, Section~\ref{Section: Generalized WorldCup Watermarking} generalizes this formulation to embed $k\ge2$ bits per token. For decoding, Section~\ref{Section: Confidence-aware Message Decoding} introduces a confidence-aware algorithm that aggregates fine-grained token-level evidence for robust and reliable message recovery.

\subsection{Binary WorldCup Watermarking} \label{Section: Binary WorldCup Watermarking}

\begin{definition}
(random seed generator).
Given a security parameter $\kappa$, define the random seed space as $\mathcal{R}=\{0,1\}^\kappa$. Let $h(\cdot)$ be a hash function and $\xi$ a watermark key. At generation step $t$, the random seed is computed as
\begin{equation}
r_t = h(x_{t-c}, \ldots, x_{t-1}, \xi) \in \mathcal{R},
\end{equation}

where $c$ denotes the sliding-window size. We assume that $r_t \sim \operatorname{Unif}(\mathcal{R})$ for any $\mathbf{x}_{<t}$ where $\operatorname{Unif}(\cdot)$ represents the uniform distribution.
\end{definition}

\begin{definition} \label{Definition: $g$-value}
($g$-value). Given a token $x\in\mathcal{V}$, a random seed $r\in\mathcal{R}$, and a layer index $\ell\in\{1,\ldots,m\}$, a $g$-value function is a pseudo-random mapping $g_\ell:\mathcal{V}\times\mathcal{R}\rightarrow \mathbb{R}$. The $g$-value of token $x$ at layer $\ell$ is the random variable $g_\ell(x,r)$.
\end{definition}

To implement binary WorldCup watermarking, we introduce two independent families of $g$-value functions $\mathbf{g}_0=(g_1^{(0)},g_2^{(0)},\ldots,g_m^{(0)})$ and $\mathbf{g}_1=(g_1^{(1)},g_2^{(1)},\ldots,g_m^{(1)})$, to encode a binary message  $\mathbf{m}\in\{0,1\}^b$. At each generation step $t$, we first sample $N^m$ (typically $2^m$) candidate tokens $\{y_0,y_1,\ldots,y_{N-1}\}$ with replacement from the model distribution $P_{\Theta}(\cdot \mid p_0,\mathbf{x}_{<t})$. Then a cryptographic hash function $h$ is applied to randomly assign the current token $x_t$ to a target message bit, determining which bit is to be embedded.

Conditioned on the selected bit value, tournament sampling is carried out as follows. If the target bit equals $0$, candidates are scored using the function family $\mathbf{g}_0$: at each tournament layer $\ell$, the score $g_\ell^{(0)}(\cdot, r_t)$ is evaluated and the higher-scoring token advances to the next layer. This elimination process continues for $m$ rounds, ultimately yielding a single winner $x_t$, which is emitted as the watermarked token. If the target bit equals $1$, the same tournament structure is applied using the alternative family $\mathbf{g}_1$, as illustrated in Algorithm~\ref{Algorthim: Binary WorldCup Watermarking}. 

\begin{algorithm}[tp]
    \caption{Binary WorldCup Watermarking}
    \label{Algorthim: Binary WorldCup Watermarking}
    \begin{algorithmic}[1] % [1] adds line numbers
    
    \STATE \textbf{Input:} LLM distribution $P_\Theta(\cdot\mid p_0,\mathbf{x}_{<t})$, \text{random seed} $r_t$, layers number $m$, leaves number $N\geq 2$, message bit $\mathbf{m}[p]$, $g$-value function families $\mathbf{g}_0, \mathbf{g}_1$
    
    \STATE Draw $N^m$ i.i.d tokens $y_0^0,\ldots,y_{N^m-1}^0$$\sim$$P_\Theta(\cdot\mid p_0,\mathbf{x}_{<t})$
        \STATE Initialize $(g_1,\ldots,g_m)\gets\mathbf{g}_0\ \operatorname{if}\ \mathbf{m}[p]=0\ \operatorname{else}\ \mathbf{g}_1$

        \FOR{$1\leq\ell\leq m$}
            \FOR{$0\leq j\leq N^{m-\ell} - 1$}
            \STATE $Y:=[y_{Nj}^{\ell-1},\ldots,y_{Nj+N+1}^{\ell-1}]$ {\quad\textcolor{gray}{// may contain repeats}}
            \STATE $Y^*:=[y\in Y:g_\ell(y,r_t)=\max_{y'\in Y}g_\ell(y',r_t)]$
            {\quad\textcolor{gray}{// may contain repeats}}
            \STATE \text{Sample} $y_j^\ell\sim\operatorname{Unif}(Y^*)$
            \ENDFOR
        \ENDFOR 
        
    \STATE \textbf{Return} $x_t\leftarrow y_0^m$
    \end{algorithmic}
\end{algorithm}

For efficiency, instead of explicitly running the tournament in Algorithm~\ref{Algorthim: Binary WorldCup Watermarking}, we use an equivalent vectorized formulation to sample from the resulting watermarked distribution\footnote{In general, we assume $2^m \gg |\mathcal{V}|$.}:

\begin{definition} \label{Definition: watermarked distribution}
(watermarked distribution).
Given a probability distribution $P_\Theta \in \Delta_{\mathcal{V}}$, a random seed $r_t \in \mathcal{R}$, layers number $m \geq 1$, leaves number $N\geq 2$ and $g$-value functions $\mathbf{g}_0,\mathbf{g}_1$. For message $\mathbf{m}$, position $p$, the watermarked distribution of the winner in Algorithm~\ref{Algorthim: Binary WorldCup Watermarking} is defined as:
\begin{equation}
q(x_t) = \mathbb{P}[\text{Alg.}~\ref{Algorthim: Binary WorldCup Watermarking}(P_\Theta, r_t, m, N, \mathbf{m}[p], \mathbf{g}_0, \mathbf{g}_1) \Rightarrow x_t].
\end{equation}
\end{definition}
Details and equivalence proofs are given in Appendix~\ref{Appendix: Vectorized WorldCup Sampling}.

\subsection{Complementary $g$-value Functions} \label{Section: Complementary $g$-value Functions}
Thus far, the two $g$-value families have been treated as arbitrary pseudo-random constructions. We now consider a principled design in which $\mathbf{g}_0$ and $\mathbf{g}_1$ are constructed as complementary pairs, and show that this choice is optimal in terms of statistical discriminability.
\begin{definition}
(complementary $g$-value).
Given a $g$-value $g_\ell(x,r)$ as defined in Definition~\ref{Definition: $g$-value}, its complementary $g$-value $\bar{g}_\ell(x,r)$ is defined as follows:
\begin{equation}
\bar{g}_\ell(x,r) \triangleq 1 - g_\ell(x,r).
\end{equation}
\end{definition}
In this paper, we focus on the case where the $g$-value follows a Bernoulli distribution $g_\ell(x,r) \sim \operatorname{Bernoulli}(0.5)$. Under this setting, both $g_\ell(x,r)$ and $\bar{g}_\ell(x,r)$ take values in $\{0,1\}$.

Intuitively, this complementary construction induces perfect anti-correlation between $\mathbf{g}_0$ and $\mathbf{g}_1$: tokens favored under $\mathbf{g}_0$ are deterministically disfavored under $\mathbf{g}_1$. As a result, the two message hypotheses are pushed to opposite extremes of the $g$-value spectrum, yielding the maximum possible separation between their induced embedding distributions. From a decoding perspective, this symmetry directly maximizes the decision margin and improves robustness against noise.

\begin{wrapfigure}{r}{8cm}
\centering
\includegraphics[width=\linewidth]{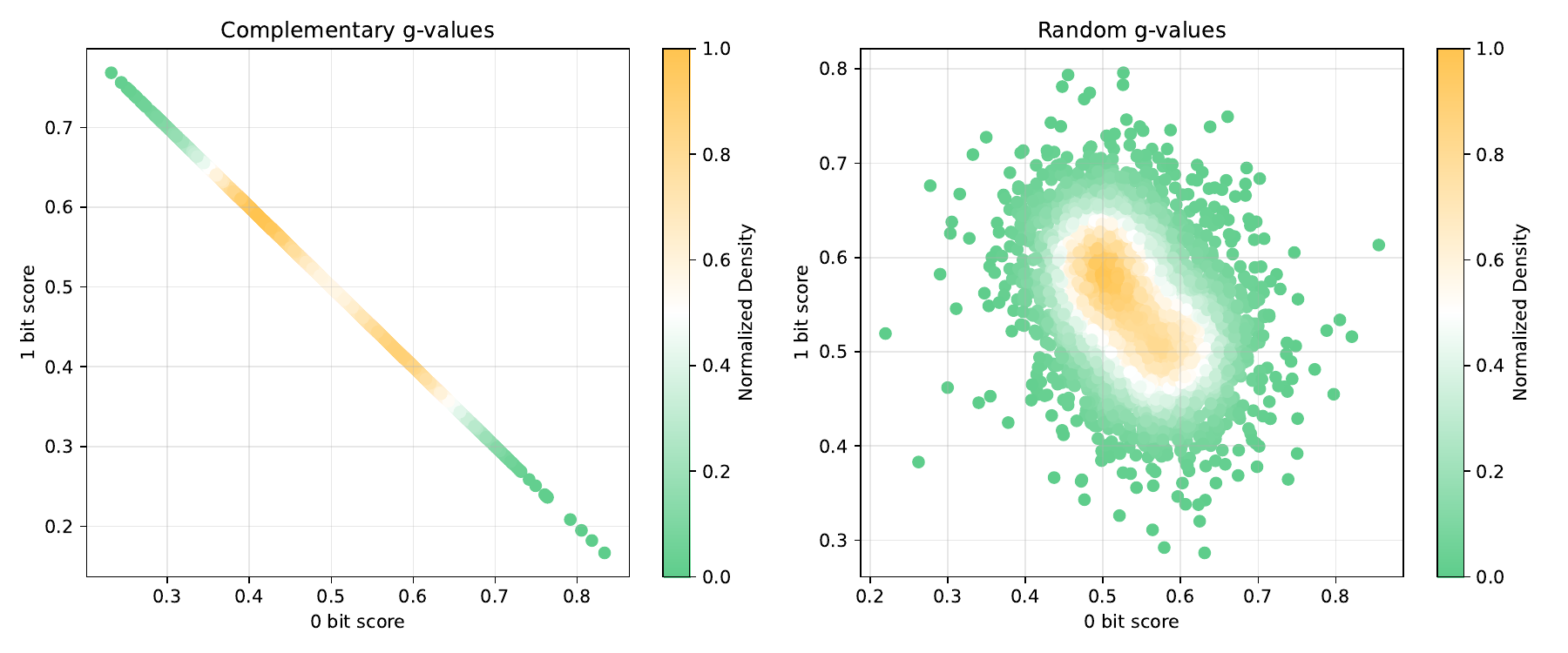}
\caption{Complementary $g$-values vs. random $g$-values.}
\label{Fig: G_values}
\end{wrapfigure}

Figure~\ref{Fig: G_values} visualizes this effect. Compared to independently sampled (random) $g$-value functions, complementary $g$-values exhibit a strict linear relationship, making the two message states ($0$ and $1$) substantially easier to distinguish. This observation motivates the following proposition, which formalizes the optimality of complementary $g$-value functions in the multi-bit watermarking setting.

\begin{proposition} \label{Proposition: g-value}
Let $g_\ell(x,r)$ be a Bernoulli $g$-value and $\bar{g}_\ell(x,r)$ its complementary counterpart. Encoding bits $0$ and $1$ by selecting between $g_\ell$ and $\bar{g}_\ell$ yields two embedding distributions whose statistical discriminability is maximized.
\end{proposition}

A formal proof based on the maximization of the expected squared difference $\mathbb{E}[(\mathbf{g}_1 - \mathbf{g}_0)^2]$ and an explicit correlation analysis are deferred to Appendix~\ref{Appendix: Proof of Proposition}.

% \begin{figure}[H]
% \centering
% \begin{minipage}[b]{0.45\textwidth}
% \centering
% \includegraphics[width=\textwidth]{img/G_values.pdf}
% \caption{Binary complementary $g$-values vs. random $g$-values.}
% \label{Fig: G_values}
% \end{minipage}
% \begin{minipage}[b]{0.45\textwidth}
% \centering
% \includegraphics[width=\textwidth]{img/Decoding.pdf}
% \caption{The comparison of multi-bit message decoding.}
% \label{Figure: Message Decoding}
% \end{minipage}
% \end{figure}

\subsection{Generalized WorldCup Watermarking} \label{Section: Generalized WorldCup Watermarking}
While binary WorldCup sampling enables one bit per token, the information allocation becomes increasingly imbalanced as the message length grows \citep{qu2025provably}. Although the nominal per-token capacity remains fixed, the global utilization of token-level degrees of freedom is suboptimal. To fully exploit the information-carrying potential of each token, we generalize the WorldCup framework by introducing $k$ groups of $g$-value function families, allowing each token to simultaneously encode $k$ bits.

We consider the simplest but nontrivial setting where each token carries two bits (i.e. $k=2$). To this end, we sample two independent families of $g$-value functions, denoted by $\mathbf{G}_{0}=(\mathbf{g}_{0},\mathbf{\bar{g}}_0)$ and $\mathbf{G}_1=(\mathbf{g}_1,\mathbf{\bar{g}}_1)$. The first bit is encoded using $\mathbf{G}_0$, following the binary setting: when the bit equals $0$, the watermark distribution $q_0$ (defined in Definition~\ref{Definition: watermarked distribution}) is derived from $\mathbf{g}_0$; otherwise, its complementary function $\mathbf{\bar{g}_0}$ is applied, yielding the distribution $\bar{q}_0$. The second bit is encoded analogously using $\mathbf{G}_1$, producing either $q_1$ or $\bar{q}_1$.

Conceptually, a token that correctly embeds both bits should score highly under both corresponding $g$-value functions. We thus aggregate their distributions additively. However, $\mathbf{G}_0$ and $\mathbf{G}_1$ are independent rather than mutually exclusive, so naive aggregation yields limited separability between bit patterns. To enhance discriminability, we subtract contributions from complementary distributions, thereby sharpening the contrast between competing message hypotheses. In terms of tournament sampling, this strategy favors tokens that rank highly under the intended $g$-value functions while ranking poorly under their complementary counterparts.

Although this more aggressive separation improves detectability, it can also distort the underlying language model distribution and degrade text quality, as demonstrated in Section~\ref{Section: Ablation Study}. To strike a better balance between detectability and fluency, we introduce an entropy-aware dynamic adjustment factor. Concretely, for a two-bit message $\mathbf{m}'$, we define the updated watermarked scores as follows:
\begin{equation} \label{Equation: 2-bit WorldCup}
    P_{\Theta,\mathbf{m}}=\left\{
            \begin{array}{lr}
            (q_0+q_1)-\lambda({\overline{q_0}} + {\overline{q_1}}), & \mathbf{m}'=00 \\
            (q_0+\overline{q_1})-\lambda(\overline{q_0} + {q_1}), & \mathbf{m}'=01 \\
            (\overline{q_0} + {q_1})-\lambda(q_0+\overline{q_1}), & \mathbf{m}'=10 \\
            ({\overline{q_0}} + \overline{q_1})-\lambda(q_0 + {q_1}), & \mathbf{m}'=11 \\
            \end{array}.
    \right.
\end{equation}

The scores are first made non-negative via clipping, then log-transformed for numerical stability and normalized via softmax, from which the \textbf{champion token} is ultimately sampled. Here, the coefficient $\lambda$ is adaptively determined by the entropy of the base LLM distribution:
\begin{equation}
    \lambda=\alpha\cdot\sigma(\sum -P_\Theta\log P_\Theta) \label{Equation: entropy factor},
\end{equation}
where $\alpha$ is a hyperparameter that controls the watermark strength, and $\sigma$ denotes the activation function, with their specific choices reported in Appendix~\ref{Appendix: Activation Function}. Under this formulation, high-entropy (low-confidence) token positions permit stronger separation, while low-entropy tokens favor more conservative modulation to preserve generation quality.

This construction can be naturally generalized to $k>2$ groups of $g$-value functions, allowing each token to embed $k$ bits. Detailed formulations are provided in Appendix~\ref{Appendix: $k$-ray WorldCup Watermarking}.

\subsection{Confidence-aware Message Decoding} \label{Section: Confidence-aware Message Decoding}

\begin{wrapfigure}{r}{8cm}
\centering
\includegraphics[width=\linewidth]{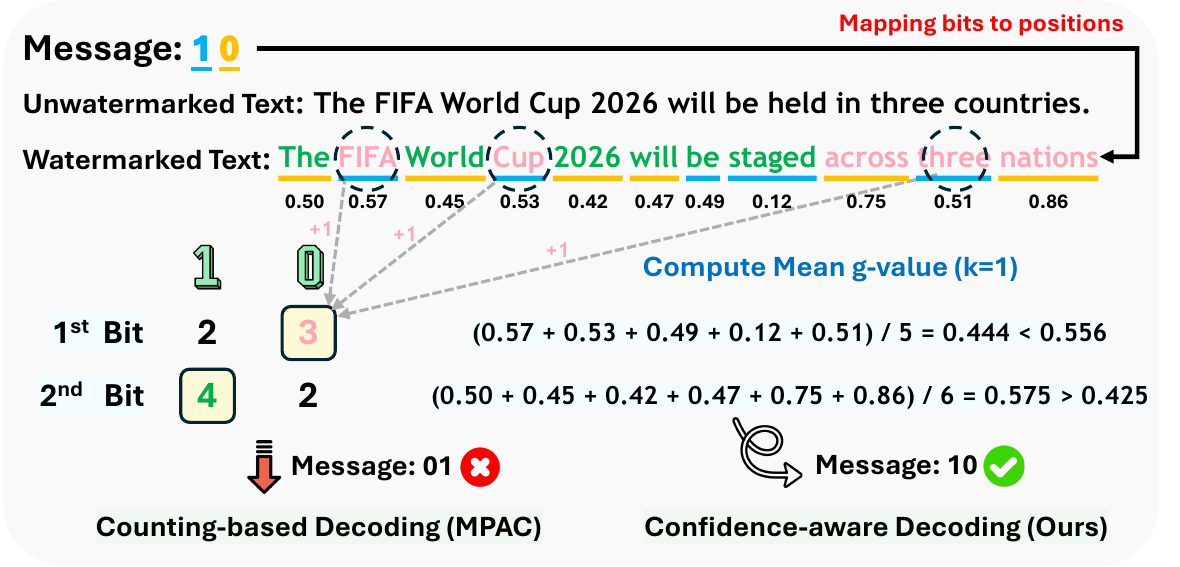}
\caption{The comparison of multi-bit message decoding: counting-based decoding vs. confidence-aware decoding.}
\label{Figure: Message Decoding}
\end{wrapfigure}

In contrast to prior counting-based decoding, our method employs a confidence-aware decoding strategy that aggregates fine-grained token-level scores across groups associated with the same message position, rather than making hard binary decisions per token. This design yields substantially more stable detection statistics, as illustrated in Fig.~\ref{Figure: Message Decoding}.

\paragraph{Problem setup.}
Let $\mathbf{y}_{1:T}=(y_1,y_2,\ldots,y_T)$ denote a generated text sequence of length $T$. As in the encoding stage, we consider $k$ independent groups of $g$-value function families, denoted by $\{\mathbf{G}_j=(\mathbf{g}_j,\mathbf{\bar g}_j)\}_{j=1}^k$. These $k$ families jointly encode a $k$-bit message string at each token position. Accordingly, a binary message in $\{0,1\}^b$ is reparameterized as a sequence of $2^k$-ary message $\mathbf{m} = (m_0,\ldots,m_{B-1})$, where $m_p \in \{0,\ldots,2^k-1\}$ for $p\in\{0,\ldots,B-1\}$, and $B = b/k$ denotes the total number of message symbols.

\paragraph{1) Recovering message positions.} We first employ the same hash function and watermarking key used during message embedding stage to identify the message symbol position $p$ associated with each token $y_t$.  All tokens assigned to the same index $p$ are then grouped together for joint decoding of the corresponding $2^k$-ary symbol $m_p$.

\paragraph{2) Computing confidence scores.} For each message position $p$ and each group $j \in \{1,\ldots,k\}$, we compute the empirical mean $g$-values over the corresponding token group, following Eq.~\ref{Equation: mean g-value}. In particular, we obtain $s^p_j$ and $\bar{s}^p_j$ corresponding to $\mathbf{g}_j$ and $\mathbf{\bar{g}}_j$, respectively. These quantities serve as calibrated confidence scores indicating whether the $j$-th bit of the $2^k$-ary message symbol at position $p$ favors bit $0$ or bit $1$.

\paragraph{3) Extracting message symbols.}
For each message position $p$, we independently infer the $j$-th bit of the embedded symbol by comparing $s^p_j$ and $\bar{s}^p_j$. The recovered $2^k$-ary message symbol $\hat m_p \in \{0, \ldots, 2^k - 1\}$ is then obtained as
\begin{equation}
\hat m_p = \sum_{j=1}^k 2^{k-j}\,\mathbb{I}\!\left(
s^p_j < \bar{s}^p_j\right).
\label{Equation: k-ray decoding}
\end{equation}

\paragraph{Computing the watermark $z$-score.}
We cast watermark detection as a hypothesis test under the null hypothesis that the text is unwatermarked. Detection relies on a standardized $z$-score (see Appendix~\ref{Appendix: Compute Watermark Z-score} for its definition and derivation). Under the null hypothesis, unwatermarked texts yield low $z$-scores, whereas watermarked texts produce systematically higher values. Thus, a sufficiently large $z$-score leads to rejection of the null hypothesis, indicating the presence of a watermark.

% We formulate watermark detection as a hypothesis test under the null hypothesis that the text is unwatermarked. Detection is based on a standardized $z$-score, whose definition and derivation are provided in Appendix~\ref{Appendix: Compute Watermark Z-score}. Under the null hypothesis, unwatermarked texts yield low $z$-score, whereas watermarked texts produce systematically larger values. Consequently, a sufficiently large $z$-score leads to rejection of the null hypothesis, indicating the presence of a watermark.

%% file: 04_Experiments.tex
\input{tables/message_embedding}

\section{Experiments} \label{Section: Experiments}
\paragraph{Overview.} We conduct a comprehensive empirical evaluation of WorldCup along three fundamental dimensions: \textbf{message embedding capacity} (Section~\ref{Experiment: Message Embedding Capacity}) including message decoding accuracy and watermark detectability, \textbf{text quality} (Section~\ref{Experiment: Robustness}), and \textbf{robustness} (Section~\ref{Experiment: Robustness}). 

% \paragraph{Overview.} We comprehensively evaluate WorldCup along three fundamental dimensions: \textbf{message embedding capacity} (Section~\ref{Experiment: Message Embedding Capacity}), \textbf{text quality} (Section~\ref{Experiment: Robustness}), and \textbf{robustness} (Section~\ref{Experiment: Robustness}). 

\subsection{Experimental Setup} \label{Experimental Setup}
\paragraph{Baselines.} We compare WorldCup with representative public, model-agnostic, training-free multi-bit watermarking methods, including BiMark \citep{feng2025bimark}, MPAC \citep{yoo-etal-2024-advancing}, SegMark \citep{qu2025provably}, StealthInk \citep{pmlr-v267-jiang25j}. Message lengths are set to 16, 24, 32, and 48 bits. Detailed settings are in Appendix~\ref{Appendix: Baselines}.

\paragraph{Datasets.} Following prior work~\citep{pmlr-v202-kirchenbauer23a, zhao2024provable}, we randomly sample 200 prompts from the C4~\citep{raffel2020exploring} and OpenGen~\citep{krishna2023paraphrasing} datasets for text generation. We further evaluate downstream performance on four representative tasks with varying input and output lengths, including machine translation, text summarization, question answering, and math reasoning. More details are in Appendix~\ref{Appendix: Datasets}.

\paragraph{Metrics.} We primarily use Bit Accuracy to evaluate message decoding correctness. Detectability is measured using AUC and Best F1 score, while text quality is assessed with Median Perplexity (PPL), ROUGE-L, BLEU, Pass@1, and the GPT-4 Score. Decoding efficiency is measured using decoding time (s). Robustness is evaluated using AUROC curves. Metric definitions are in Appendix~\ref{Appendix: Metrics}.

\paragraph{Models.} For detectability and robustness evaluation, we use LLaMA3-8B-Base \cite{grattafiori2024llama} and Gemma2-9B-Base \cite{team2024gemma}. For downstream tasks, we additionally evaluate their Instruct-tuned variants including LLaMA3.1-8B-Instruct, Gemma2-9B-Instruct and the latest Ministral-8B-Instruct \citep{liu2026ministral}. For fairness, we compute PPL using the larger Vicuna-13B-v1.5 model \citep{vicuna2023}. The details are in Appendix~\ref{Appendix: Models}.

\paragraph{Implementation Details.} Our method is implemented in Python 3.10.0 with PyTorch 2.6.0. All experiments are conducted on a single NVIDIA A100 80 GB GPU. We use a default setup with $g$-value function number $k=2$, temperature = 1.0, top-k = 50, top-p = 0.95, no\_repeat\_ngram\_size = 4, layers number $m=30$, and $\alpha=1.2$, together with lefthash and window size $c = 2$.

\subsection{Message Embedding Capacity}  \label{Experiment: Message Embedding Capacity}

\paragraph{Multi-bit watermarking scenario.}
We evaluate decoding performance with maximum sequence lengths of 128 and 256 tokens and a minimum length of 64. Table~\ref{Table: Bit Acc} shows that WorldCup consistently outperforms all baselines across both backbones. On LLaMA3-8B, it improves bit accuracy over StealthInk by \textbf{3.2\%}, \textbf{4.8\%}, \textbf{5.5\%}, and \textbf{5.2\%} for 16-bit, 24-bit, 32-bit, and 48-bit messages, respectively. Overall, these results indicate the effectiveness of WorldCup, where incorporating multiple $g$-functions improves per-token information utilization and leads to more reliable message recovery. We also observe that BiMark, MPAC, SegMark, and WorldCup exhibit lower average decoding accuracy on Gemma2-9B than LLaMA3-8B, with drops of \textbf{7.1\%}, \textbf{5.3\%}, \textbf{8.0\%}, and \textbf{5.2\%}, respectively. This gap is likely due to token-level uncertainly and details are in Appendix~\ref{Appendix: Entropy Analysis}.

\paragraph{Zero-bit watermarking scenario.} Using the method described in Section~\ref{Section: Confidence-aware Message Decoding} to compute the $z$-score as the detection statistic, WorldCup achieves an average AUC of approximately \textbf{99.7\%} across all scenarios. Moreover, 256-token sequences consistently outperform 128-token sequences, highlighting that longer contexts naturally accumulate stronger statistical evidence. These results not only confirm that detectability remains stable under large payloads, but also demonstrate that WorldCup gracefully subsumes the zero-bit watermarking regime without additional mechanisms.

\input{tables/text_quality}

\subsection{Text Quality Preservation} \label{Experiment: Text Quality}

\paragraph{Perplexity (PPL).} As shown in Table~\ref{Table: Bit Acc} and Fig.~\ref{Figure: PPL}, WorldCup consistently achieves lower perplexity than existing baselines, indicating better fluency preservation. We also observe that longer generated sequences tend to yield lower PPL. For instance, compared to 128-token outputs, 256-token sequences generated by WorldCup reduce PPL by \textbf{2.01} and \textbf{1.82} on LLaMA3-8B and Gemma2-9B, respectively. 

% Notably, WorldCup ($k=1$) even attains a PPL lower than that of natural text, suggesting that it implicitly regularizes generation and effectively mitigates distributional shifts from original LLM. Additional hyperparameter effects on PPL, such as $\textit{no\_repeat\_ngram\_size}$, are reported in Appendix~\ref{Appendix: Key Generation Hyperparameters}.

\begin{wrapfigure}{r}{8cm}
\centering
\includegraphics[width=\linewidth]{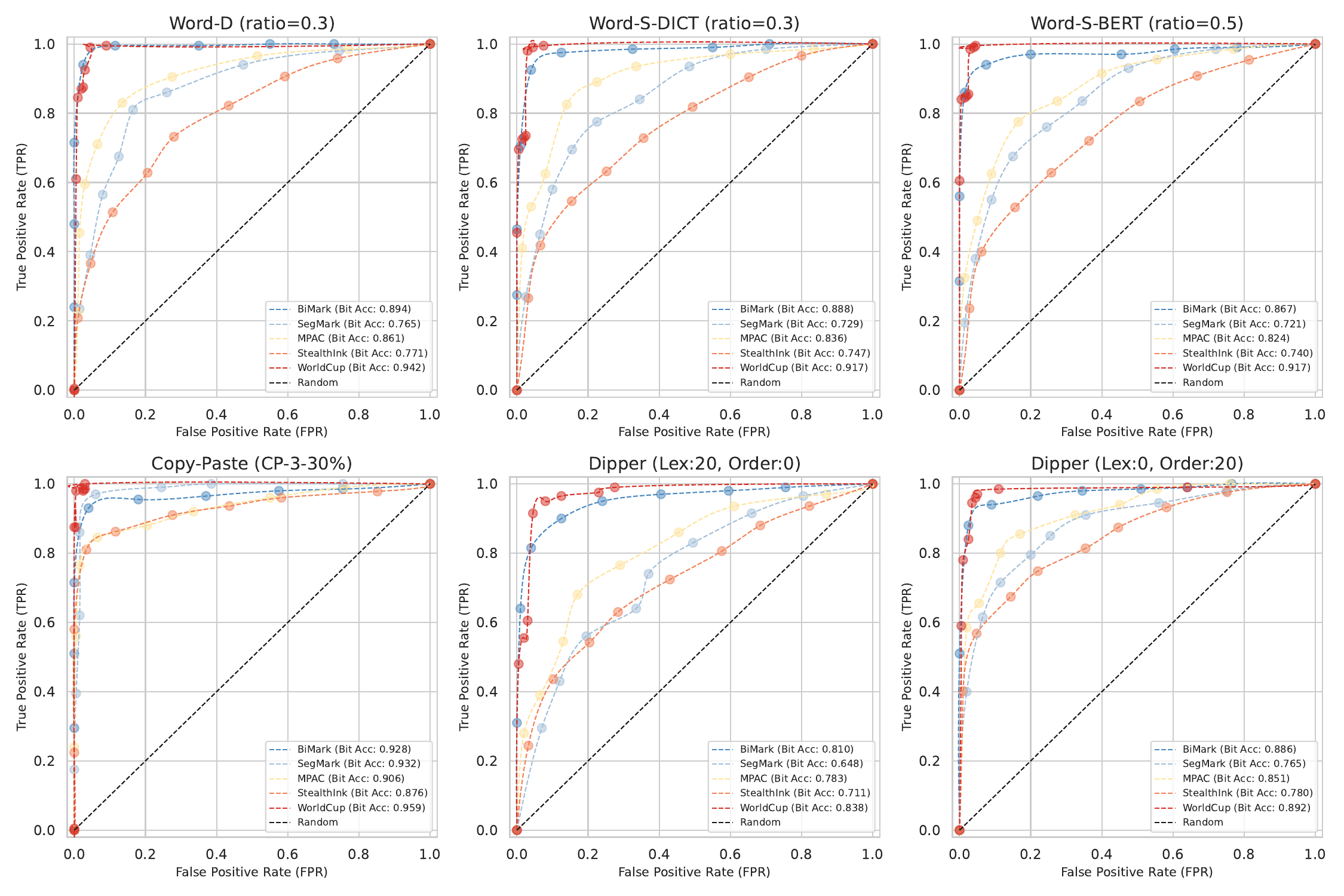}
\caption{The AUROC under various attacks including word-level and sentence-level on LLaMA3-8B-Base when embedding a 16-bit message with a 256-token sequence.}
\label{Figure: Robustness}
\end{wrapfigure}

% Prior studies typically did not enforce the $\textit{no\_repeat\_ngram\_size}$ constraint during evaluation. Although this setting can reduce PPL, it also increases the likelihood of repetitive and degenerate outputs (see Appendix~\ref{Appendix: Key Generation Hyperparameters}). To better mimic realistic deployment scenarios and maintain text quality, we enable this constraint to prevent repetition. As reported in Table~\ref{Table: Bit Acc} and Fig.~\ref{Figure: PPL}, WorldCup consistently reports lower perplexity than existing baselines, demonstrating strong fluency preservation. Notably, WorldCup ($k=1$) even achieves a PPL lower than natural text, suggesting that our sampling strategy implicitly regularizes generation and effectively limits distributional shift from the model’s native sampling behavior.

\paragraph{Downstream Task Performance.} As summarized in Table~\ref{Table: Downstream Task}, WorldCup achieves the best overall performance across diverse tasks. This advantage stems from its multi-round tournament sampling, which effectively preserves text quality. Meanwhile, WorldCup attains the highest multi-bit decoding accuracy and outperforms baselines on several tasks, especially those involving longer outputs. For example, in the long-form QA task, it surpasses MPAC by an average of \textbf{2.6\%}, \textbf{2.7\%}, and \textbf{3.0\%} on Ministral, LLaMA3.1, and Gemma2, respectively. These results demonstrate that our method ensures both reliable decoding and high text quality, enabling flexible trade-offs for different tasks.

\subsection{Robustness to Various Attacks} 
\label{Experiment: Robustness}

\paragraph{Attack Settings.} We assess robustness under three common attacks~\citep{kirchenbauer2024on}: editing (word deletion and synonym substitution), copy–paste, and paraphrasing using Dipper~\citep{krishna2023paraphrasing} using 256-token watermarked texts. Full settings are provided in Appendix~\ref{Appendix: Attack Settings}. 

As shown in Fig.~\ref{Figure: Robustness}, Fig.~\ref{Figure: Llama3-Robustness-Appendix}, and Fig.~\ref{Figure: Gemma2-Robustness-Appendix}, WorldCup consistently outperforms existing watermarking methods across diverse attack scenarios, achieving higher AUC and decoding accuracy. As illustrated in Fig.~\ref{Figure: Robustness}, WorldCup attains an average AUC of \textbf{98.9\%} and a decoding accuracy of \textbf{91.1\%}, surpassing BiMark, MPAC, StealthInk and SegMark by \textbf{3.2\%}, \textbf{6.7\%}, \textbf{14\%} and \textbf{15.1\%}, respectively. This robustness stems from its more effective use of per-token redundancy to encode multiple bits, preserving sufficient watermark signal under strong perturbations. Intuitively, embedding more bits requires longer sequences for reliable extraction and detection; otherwise, fewer tokens per bit can reduce robustness.
We also study the effect of hash window size in Appendix~\ref{Appendix: Key Generation Hyperparameters} and find that larger windows reduce robustness by weakening the locality of watermark signal aggregation.

% \begin{figure}[tp]
% \centering
% \includegraphics[width=\linewidth]{img/robustness_c4_Llama3-8B-Base_16_bits_main.pdf}
% \caption{The AUROC curves under different attacks on LLaMA3.}
% \label{Figure: Robustness}
% \end{figure}

% As shown in Fig.~\ref{Figure: Robustness}, Fig.~\ref{Figure: Llama3-Robustness-Appendix} and Fig.~\ref{Figure: Gemma2-Robustness-Appendix}, WorldCup ($k=2$) consistently surpasses existing watermarking methods across diverse attack settings, yielding higher AUC scores and decoding accuracy. This robustness stems from its ability to more fully exploit per-token redundancy to encode multiple bits, ensuring that sufficient watermark signal survives even under substantial perturbations. Nonetheless, we observe that achieving strong robustness for multi-bit watermarking remains challenging. Unlike zero-bit watermarking, which aggregates watermark evidence from all tokens across the entire text, bit-assignment–based decoding relies only on tokens associated with each specific message bit. Consequently, it exhibits heightened vulnerability under constrained sequence lengths or high payloads, where fewer tokens contribute to decoding each bit. We further analyze the influence of the hash window size in Appendix~\ref{Appendix: Key Generation Hyperparameters}, showing that increasing the window size weakens robustness due to reduced locality in watermark signal aggregation.

%% file: tables/message_embedding.tex
\definecolor{top1}{HTML}{ed583a}
\definecolor{top2}{HTML}{69c9b1}

% \definecolor{top1}{HTML}{ff8ba1}
% \definecolor{top2}{HTML}{3ecdb2}

\begin{table*}[tp]
    \centering
    \caption{The comparison of different multi-bit watermark performance on LLaMA3-8B-Base and Gemma2-9B-Base, where \faToggleOff\ denotes the representative baselines, \faToggleOn\ denotes our watermark method.}
    \resizebox{\textwidth}{!}{\begin{tabular}{c|l|cccccccccccccccc}
    \toprule[1.5pt]
    \multirow{3}{*}{\textbf{Bit Length}} & \multicolumn{1}{c|}{\multirow{3}{*}{\textbf{Watermark}}} & \multicolumn{8}{c}{\textsc{\textbf{Llama3-8B-Base}}} & \multicolumn{8}{c}{\textsc{\textbf{Gemma2-9B-Base}}}\\

    & & \multicolumn{4}{c}{\textsc{max 128 tokens}} & \multicolumn{4}{c}{\textsc{max 256 tokens}} & \multicolumn{4}{c}{\textsc{max 128 tokens}} & \multicolumn{4}{c}{\textsc{max 256 tokens}}\\
    
    \cmidrule(lr){3-6}
    \cmidrule(lr){7-10} 
    \cmidrule(lr){11-14}
    \cmidrule(lr){15-18}
    
    & & AUC $\uparrow$& Bit Acc $\uparrow$ & PPL $\downarrow$ & Time (s) $\downarrow$
    & AUC $\uparrow$& Bit Acc $\uparrow$ & PPL $\downarrow$ & Time (s) $\downarrow$
    & AUC $\uparrow$& Bit Acc $\uparrow$ & PPL $\downarrow$ & Time (s) $\downarrow$
    & AUC $\uparrow$& Bit Acc $\uparrow$ & PPL $\downarrow$ & Time (s) $\downarrow$ \\
    \midrule
    
    % - & - & - & - & - & - & - & - & - & - & - & - & - & - & - & - & - & - \\
    % \midrule
    
    \multirow{4}{*}{16 bits} 
    & \faToggleOff \ MPAC & 0.999 & 0.960 & 16.25 & \textbf{\color{top2}0.014} & 0.996 & 0.980 & 13.56 & \textbf{\color{top2}0.027} & 0.980 & 0.920 & 13.69 & 0.015 & 0.985 & 0.940 & 12.00 & \textbf{\color{top2}0.030} \\
    & \faToggleOff \ SegMark & 0.993 & 0.948 & 15.88 & 0.551 & 0.995 & \textbf{\color{top1}0.999} & 12.94 & 0.839 & 0.979 & 0.895 & 13.19 & 1.234 & 0.999 & \textbf{\color{top1}0.977} & 10.94 & 2.402 \\
    & \faToggleOff \ StealthInk & 0.992 & 0.937 & 14.69 & 0.016 & 0.997 & 0.972 & 13.28 & 0.038 & 0.951 & 0.873 & 11.72 & \textbf{\color{top2}0.014} & 0.978 & 0.918 & \textbf{\color{top2}9.500} & 0.031 \\
    & \faToggleOff \ BiMark & 1.000 & \textbf{\color{top2}0.977} & \textbf{\color{top2}14.34} & 0.026 & 1.000 & 0.987 & \textbf{\color{top2}11.25} & 0.044 & 1.000 & \textbf{\color{top2}0.929} & \textbf{\color{top2}11.06} & 0.031 & 0.999 & 0.955 & 9.625 & 0.039 \\
    % & \faToggleOn \textbf{ Ours ($k=1$)} & 1.000 & 0.975 & \textbf{\color{top1}8.000} & \textbf{\color{top1}0.008} & 1.000 & 0.986 & \textbf{\color{top1}6.703} & \textbf{\color{top1}0.012} & 0.993 & 0.909 & \textbf{\color{top1}6.750} & \textbf{\color{top1}0.008} & 1.000 & 0.951 & \textbf{\color{top1}5.719} & \textbf{\color{top1}0.011} \\
    & \faToggleOn \textbf{ WorldCup } & 0.999 & \textbf{\color{top1}0.982} & \textbf{\color{top1}12.94} & \textbf{\color{top1}0.009} & 1.000 & \textbf{\color{top2}0.990} & \textbf{\color{top1}11.25} & \textbf{\color{top1}0.015} & 0.998 & \textbf{\color{top1}0.931} & \textbf{\color{top1}10.84} & \textbf{\color{top1}0.008} & 0.998 & \textbf{\color{top2}0.956} & \textbf{\color{top1}8.906} & \textbf{\color{top1}0.012} \\
    \midrule

    \multirow{4}{*}{24 bits}
    & \faToggleOff \ MPAC & 0.996 & 0.916 & 16.75 & \textbf{\color{top2}0.012} & 0.997 & 0.955 & 14.00 & \textbf{\color{top2}0.016} & 0.972 & 0.875 & 13.69 & \textbf{\color{top2}0.017} & 0.959 & 0.905 & 12.00 & \textbf{\color{top2}0.035} \\
    & \faToggleOff \ SegMark & 0.947 & 0.839 & 17.13 & 0.721 & 0.992 & \textbf{\color{top1}0.992} & 13.38 & 0.804 & 0.849 & 0.734 & 13.19 & 1.670 & 0.974 & \textbf{\color{top2}0.922} & 10.13 & 2.853 \\
    & \faToggleOff \ StealthInk & 0.989 & 0.885 & 15.16 & 0.018 & 0.995 & 0.935 & 13.19 & 0.022 & 0.917 & 0.830 & 11.81 & 0.019 & 0.975 & 0.877 & 10.13 & 0.039 \\
    & \faToggleOff \ BiMark & 1.000 & \textbf{\color{top2}0.941} & \textbf{\color{top2}14.34} & 0.027 & 1.000 & 0.970 & \textbf{\color{top1}11.34} & 0.036 & 0.998 & \textbf{\color{top2}0.875} & \textbf{\color{top2}11.63} & 0.030 & 0.990 & 0.921 & \textbf{\color{top2}9.938} & 0.043 \\
    & \faToggleOn \textbf{ WorldCup } & 1.000 & \textbf{\color{top1}0.943} & \textbf{\color{top1}13.69} & \textbf{\color{top1}0.009} & 1.000 & \textbf{\color{top2}0.972} & \textbf{\color{top2}11.53} & \textbf{\color{top1}0.014} & 0.998 & \textbf{\color{top1}0.894} & \textbf{\color{top1}10.44} & \textbf{\color{top1}0.008} & 0.999 & \textbf{\color{top1}0.925} & \textbf{\color{top1}8.625} & \textbf{\color{top1}0.012} \\
    \midrule

    \multirow{4}{*}{32 bits}
    & \faToggleOff \ MPAC & 0.997 & 0.890 & 16.75 & \textbf{\color{top2}0.021} & 0.996 & 0.939 & 14.00 & \textbf{\color{top2}0.020} & 0.947 & \textbf{\color{top2}0.837} & 13.81 & \textbf{\color{top2}0.016} & 0.940 & \textbf{\color{top2}0.869} & 11.63 & \textbf{\color{top2}0.034} \\
    & \faToggleOff \ SegMark & 0.909 & 0.802 & 16.63 & 0.642 & 0.989 & \textbf{\color{top1}0.964} & 12.94 & 1.053 & 0.824 & 0.691 & 12.75 & 1.751 & 0.955 & 0.868 & 10.25 & 3.362 \\
    & \faToggleOff \ StealthInk & 0.990 & 0.845 & 14.44 & 0.024 & 0.994 & 0.919 & 12.75 & 0.029 & 0.943 & 0.797 & 12.00 & 0.018 & 0.962 & 0.850 & 10.13 & 0.036 \\
    & \faToggleOff \ BiMark & 1.000 & \textbf{\color{top2}0.890} & \textbf{\color{top2}13.91} & 0.026 & 1.000 & 0.947 & \textbf{\color{top2}12.38} & 0.040 & 0.996 & 0.806 & \textbf{\color{top2}12.00} & 0.031 & 0.989 & 0.865 & \textbf{\color{top2}9.938} & 0.041 \\
    % & \faToggleOn \textbf{ Ours ($k=1$)} & 0.999 & \textbf{\color{top2}0.893} & \textbf{\color{top1}7.813} & \textbf{\color{top1}0.009} & 1.000 & 0.940 & \textbf{\color{top1}6.688} & \textbf{\color{top1}0.013} & 0.990 & \textbf{\color{top2}0.842} & \textbf{\color{top1}7.094} & \textbf{\color{top1}0.008} & 0.995 & \textbf{\color{top2}0.885} & \textbf{\color{top1}5.906} & \textbf{\color{top1}0.013} \\
    & \faToggleOn \textbf{ WorldCup } & 1.000 & \textbf{\color{top1}0.915} & \textbf{\color{top1}13.38} & \textbf{\color{top1}0.010} & 1.000 & \textbf{\color{top2}0.958} & \textbf{\color{top1}11.25} & \textbf{\color{top1}0.017} & 0.995 & \textbf{\color{top1}0.859} & \textbf{\color{top1}10.94} & \textbf{\color{top1}0.009} & 0.994 & \textbf{\color{top1}0.905} & \textbf{\color{top1}8.688} & \textbf{\color{top1}0.013} \\
    \midrule

    \multirow{4}{*}{48 bits}
    & \faToggleOff \ MPAC & 0.993 & \textbf{\color{top2}0.828} & 16.63 & \textbf{\color{top2}0.017} & 0.990 & \textbf{\color{top2}0.891} & 13.19 & \textbf{\color{top2}0.039} & 0.936 & \textbf{\color{top2}0.770} & 14.00 & 0.018 & 0.914 & \textbf{\color{top2}0.821} & 12.19 & \textbf{\color{top2}0.038} \\
    & \faToggleOff \ SegMark & 0.837 & 0.628 & 16.38 & 0.725 & 0.968 & 0.848 & 13.56 & 1.381 & 0.770 & 0.600 & 13.38 & 2.668 & 0.900 & 0.697 & 9.938 & 3.382 \\
    & \faToggleOff \ StealthInk & 0.977 & 0.804 & 14.69 & 0.020 & 0.983 & 0.870 & 12.94 & 0.041 & 0.904 & 0.738 & 11.81 & \textbf{\color{top2}0.017} & 0.932 & 0.797 & \textbf{\color{top2}9.938} & 0.039 \\

    & \faToggleOff \ BiMark & 1.000 & 0.783 & \textbf{\color{top2}14.25} & 0.027 & 0.999 & 0.880 & \textbf{\color{top2}11.63} & 0.037 & 0.985 & 0.688 & \textbf{\color{top2}11.53} & 0.029 & 0.975 & 0.767 & 10.03 & 0.040 \\
    % & \faToggleOn \textbf{ Ours ($k=1$)} & 0.997 & \textbf{\color{top2}0.838} & \textbf{\color{top1}7.938} & \textbf{\color{top1}0.011} & 1.000 & \textbf{\color{top2}0.897} & \textbf{\color{top1}6.594} & \textbf{\color{top1}0.014} & 0.987 & 0.768 & \textbf{\color{top1}6.875} & \textbf{\color{top1}0.009} & 0.994 & \textbf{\color{top2}0.822} & \textbf{\color{top1}5.656} & \textbf{\color{top1}0.014} \\
    & \faToggleOn \textbf{ WorldCup } & 1.000 & \textbf{\color{top1}0.862} & \textbf{\color{top1}13.69} & \textbf{\color{top1}0.011} & 1.000 & \textbf{\color{top1}0.916} & \textbf{\color{top1}11.63} & \textbf{\color{top1}0.017} & 0.993 & \textbf{\color{top1}0.800} & \textbf{\color{top1}10.34} & \textbf{\color{top1}0.010} & 0.989 & \textbf{\color{top1}0.851} & \textbf{\color{top1}9.063} & \textbf{\color{top1}0.015} \\
    
    \bottomrule[1.5pt]
    \end{tabular}}
    \label{Table: Bit Acc}
\end{table*}

%% file: tables/text_quality.tex
\definecolor{bit1}{HTML}{ff7d29}
\definecolor{bit2}{HTML}{ffbf78}
\definecolor{bit3}{HTML}{ffeea9}
\definecolor{bit4}{HTML}{feffd2}

\definecolor{score1}{HTML}{42abf4}
\definecolor{score2}{HTML}{a0d7fd}
\definecolor{score3}{HTML}{cdecfe}
\definecolor{score4}{HTML}{e5f5fc}

\begin{table*}[tp]
    \caption{The performance of various watermarking methods across four different downstream tasks on three instruct-tuned LLMs, including Ministral-8B-IT, LLaMA3.1-8B-IT and Gemma2-9B-IT.}
    \resizebox{\textwidth}{!}{
    \begin{tabular}{lcccccccccccc}
        \toprule[1.5pt]
        
        \large{Model} & \multicolumn{3}{c}{\textbf{Task 1: Short Q, Short A}} & \multicolumn{3}{c}{\textbf{Task 2: Long Q, Short A}} & \multicolumn{3}{c}{\textbf{Task 3: Short Q, Long A}} & \multicolumn{3}{c}{\textbf{Task 4: Long Q, Long A}} \\

        & \multicolumn{3}{c}{\textit{Machine Translation}} & \multicolumn{3}{c}{\textit{Text Summarization}} & \multicolumn{3}{c}{\textit{Long-form QA}}& \multicolumn{3}{c}{\textit{Math Reasoning}}\\

        % {\Large Model} & \multicolumn{4}{c}{{\textbf{T1: Factual Knowledge}}} & \multicolumn{4}{c}{\textbf{T2: Long-form QA}} & \multicolumn{4}{c}{\textbf{T3: Math Reasoning}} & \multicolumn{4}{c}{\textbf{T4: Code Generation}} \\

        \cmidrule(lr){2-4}
        \cmidrule(lr){5-7}
        \cmidrule(lr){8-10}
        \cmidrule(lr){11-13}
                
        \large{+ Watermark} & Best F1 $\uparrow$ & Bit Accuracy $\uparrow$ & BLEU $\uparrow$ & Best F1 $\uparrow$ & Bit Accuracy $\uparrow$ &  ROUGE-L $\uparrow$ & Best F1 $\uparrow$ & Bit Accuracy $\uparrow$ & GPT4 Score $\uparrow$ & Best F1 $\uparrow$ & Bit Accuracy $\uparrow$  & Pass@1 $\uparrow$\\
        
        \hline \specialrule{0em}{2pt}{0pt}
        
        \makecell[l]{\textbf{\textsc{Ministral-8B-IT}}} & - & - & 0.200 \small{± 0.002} & - & - & 0.221 \small{± 0.001} & - & - & 5.250 \small{± 0.073} & - & - & 0.340 \small{± 0.010} \\

        \specialrule{0em}{1pt}{1pt}

        + \faToggleOff \ MPAC & 0.745 \small{± 0.005} & \cellcolor{bit2!50} 0.886 \small{± 0.004} & \cellcolor{score4!50} 0.166 \small{± 0.001} & 0.951 \small{± 0.003} & \cellcolor{bit3!50} 0.929 \small{± 0.002} & \cellcolor{score2!50} 0.215 \small{± 0.001} & 0.946 \small{± 0.002} & \cellcolor{bit3!50} 0.879 \small{± 0.001} & \cellcolor{score2!50} 5.122 \small{± 0.029} & 0.785 \small{± 0.002} & \cellcolor{bit3!50} 0.879 \small{± 0.001} & \cellcolor{score4!50} 0.265 \small{± 0.000} \\
        
        + \faToggleOff \ BiMark & 0.776 \small{± 0.006} & \cellcolor{bit4!50} 0.803 \small{± 0.030} & \cellcolor{score2!50} 0.193 \small{± 0.003} & 0.977 \small{± 0.005} & \cellcolor{bit2!50} 0.943 \small{± 0.005} & \cellcolor{score3!50} 0.208 \small{± 0.002} & 0.992 \small{± 0.000} & \cellcolor{bit2!50} 0.894 \small{± 0.002} & \cellcolor{score4!50} 4.998 \small{± 0.088} & 0.767 \small{± 0.006} & \cellcolor{bit4!50} 0.877 \small{± 0.002} & \cellcolor{score2!50} 0.310 \small{± 0.050} \\

        + \faToggleOff \ StealthInk & 0.723 \small{± 0.001} & \cellcolor{bit3!50} 0.843 \small{± 0.010} & \cellcolor{score3!50} 0.181 \small{± 0.008} & 0.930 \small{± 0.003} & \cellcolor{bit4!50} 0.923 \small{± 0.006} & \cellcolor{score4!50} 0.206 \small{± 0.005} & 0.876 \small{± 0.011} & \cellcolor{bit4!50} 0.799 \small{± 0.006} & \cellcolor{score1!50} 5.213 \small{± 0.016} & 0.832 \small{± 0.003} & \cellcolor{bit2!50} 0.919 \small{± 0.010} & \cellcolor{score3!50} 0.305 \small{± 0.010} \\

        % + WorldCup ($k=1$) & 0.672 \small{± 0.000} & \cellcolor{bit4!50} 0.793 \small{± 0.000} & \cellcolor{score1!50} 0.221 \small{± 0.000} & 0.967 \small{± 0.008} & \cellcolor{bit2!50} 0.948 \small{± 0.002} & \cellcolor{score1!50} 0.231 \small{± 0.002} & 0.977 \small{± 0.003} & \cellcolor{bit4!50} 0.871 \small{± 0.003} & \cellcolor{score1!50} 5.319 \small{± 0.008} & 0.777 \small{± 0.002} & \cellcolor{bit4!50} 0.854 \small{± 0.002} & \cellcolor{score1!50} 0.355 \small{± 0.000} \\

        + \faToggleOn \textbf{ WorldCup } & 0.752 \small{± 0.004} & \cellcolor{bit1!50} 0.919 \small{± 0.009} & \cellcolor{score1!50} 0.202 \small{± 0.005} & 0.983 \small{± 0.001} & \cellcolor{bit1!50} 0.955 \small{± 0.002} & \cellcolor{score1!50} 0.216 \small{± 0.004} & 0.996 \small{± 0.001} & \cellcolor{bit1!50} 0.905 \small{± 0.002} & \cellcolor{score3!50} 5.207 \small{± 0.092} & 0.898 \small{± 0.000} & \cellcolor{bit1!50} 0.953 \small{± 0.000} & \cellcolor{score1!50} 0.315 \small{± 0.000} \\

        \midrule

        \makecell[l]{\textbf{\textsc{LLaMA3.1-8B-it}}} & - & - & 0.268 \small{± 0.004} & - & - & 0.249 \small{± 0.001} & - & - & 5.243 \small{± 0.069} & - & - & 0.500 \small{± 0.020}\\

        \specialrule{0em}{1pt}{1pt}

        + \faToggleOff \ MPAC & 0.700 \small{± 0.000} & \cellcolor{bit2!50} 0.805 \small{± 0.002} & \cellcolor{score4!50} 0.224 \small{± 0.002} & 0.880 \small{± 0.010} & \cellcolor{bit3!50} 0.869 \small{± 0.004} & \cellcolor{score1!50} 0.249 \small{± 0.002} & 0.940 \small{± 0.003} & \cellcolor{bit3!50} 0.855 \small{± 0.003} & \cellcolor{score3!50} 5.110 \small{± 0.052} & 0.856 \small{± 0.006} & \cellcolor{bit2!50} 0.927 \small{± 0.004} & \cellcolor{score4!50} 0.343 \small{± 0.007} \\
        
        + \faToggleOff \ BiMark & 0.718 \small{± 0.002} & \cellcolor{bit3!50} 0.768 \small{± 0.068} & \cellcolor{score2!50} 0.258 \small{± 0.005} & 0.960 \small{± 0.004} & \cellcolor{bit2!50} 0.898 \small{± 0.004} & \cellcolor{score4!50} 0.242 \small{± 0.001}& 0.981 \small{± 0.000} & \cellcolor{bit2!50} 0.872 \small{± 0.005} & \cellcolor{score4!50} 5.066 \small{± 0.110} & 0.868 \small{± 0.005} & \cellcolor{bit3!50} 0.922 \small{± 0.006} & \cellcolor{score2!50} 0.420 \small{± 0.020} \\

        + \faToggleOff \ StealthInk & 0.672 \small{± 0.001} & \cellcolor{bit4!50} 0.766 \small{± 0.018} & \cellcolor{score3!50} 0.256 \small{± 0.005} & 0.877 \small{± 0.021} & \cellcolor{bit4!50} 0.848 \small{± 0.003} & \cellcolor{score3!50} 0.243 \small{± 0.001} & 0.845 \small{± 0.005} & \cellcolor{bit4!50} 0.779 \small{± 0.001} & \cellcolor{score1!50} 5.391 \small{± 0.050} & 0.775 \small{± 0.007} & \cellcolor{bit4!50} 0.872 \small{± 0.008} & \cellcolor{score3!50} 0.350 \small{± 0.015} \\

        % + WorldCup ($k=1$) & 0.672 \small{± 0.000} & \cellcolor{bit3!50} 0.768 \small{± 0.000} & \cellcolor{score1!50} 0.315 \small{± 0.000} & 0.912 \small{± 0.006} & \cellcolor{bit4!50} 0.860 \small{± 0.004} & \cellcolor{score1!50} 0.261 \small{± 0.003} & 0.952 \small{± 0.006} & \cellcolor{bit4!50} 0.843 \small{± 0.004} & \cellcolor{score1!50} 5.258 \small{± 0.062} & 0.814 \small{± 0.009} & \cellcolor{bit4!50} 0.917 \small{± 0.004} & \cellcolor{score1!50} 0.468 \small{± 0.012} \\

        + \faToggleOn \textbf{ WorldCup } & 0.705 \small{± 0.003} & \cellcolor{bit1!50} 0.850 \small{± 0.025} & \cellcolor{score1!50} 0.272 \small{± 0.001} & 0.963 \small{± 0.004} & \cellcolor{bit1!50} 0.919 \small{± 0.006} & \cellcolor{score2!50} 0.247 \small{± 0.002} & 0.986 \small{± 0.000} & \cellcolor{bit1!50} 0.882 \small{± 0.005} & \cellcolor{score2!50} 5.149 \small{± 0.024} & 0.892 \small{± 0.003} & \cellcolor{bit1!50} 0.947 \small{± 0.001} & \cellcolor{score1!50} 0.445 \small{± 0.005} \\
        
        \midrule

        \makecell[l]{\textbf{\textsc{Gemma2-9B-it}}} & - & - & 0.407 \small{± 0.001} & - & - & 0.312 \small{± 0.002} & - & -  & 6.027 \small{± 0.093} & - & - & 0.650 \small{± 0.010}\\

        \specialrule{0em}{1pt}{1pt}

        + \faToggleOff \ MPAC & 0.673 \small{± 0.000} & \cellcolor{bit2!50} 0.686 \small{± 0.007} & \cellcolor{score3!50} 0.393 \small{± 0.002} & 0.671 \small{± 0.000} & \cellcolor{bit2!50} 0.619 \small{± 0.001} & \cellcolor{score4!50} 0.309 \small{± 0.001} & 0.838 \small{± 0.000} & \cellcolor{bit2!50} 0.770 \small{± 0.001} & \cellcolor{score4!50} 5.920 \small{± 0.003} & 0.670 \small{± 0.000} & \cellcolor{bit3!50} 0.670 \small{± 0.000} & \cellcolor{score2!50} 0.625 \small{± 0.000} \\

        + \faToggleOff \ BiMark & 0.675 \small{± 0.002} & \cellcolor{bit3!50} 0.606 \small{± 0.027} & \cellcolor{score2!50} 0.398 \small{± 0.002} & 0.669 \small{± 0.000} & \cellcolor{bit3!50} 0.598 \small{± 0.011} & \cellcolor{score1!50} 0.317 \small{± 0.003} & 0.883 \small{± 0.004} & \cellcolor{bit3!50} 0.762 \small{± 0.001} & \cellcolor{score3!50} 5.946 \small{± 0.027} & 0.676 \small{± 0.002} & \cellcolor{bit2!50} 0.706 \small{± 0.003} & \cellcolor{score3!50} 0.590 \small{± 0.000} \\

        + \faToggleOff \ StealthInk & 0.667 \small{± 0.000} & \cellcolor{bit4!50} 0.604 \small{± 0.001} & \cellcolor{score4!50} 0.379 \small{± 0.003} & 0.667 \small{± 0.003} & \cellcolor{bit4!50} 0.593 \small{± 0.006} & \cellcolor{score2!50} 0.316 \small{± 0.002} & 0.712 \small{± 0.002} & \cellcolor{bit4!50} 0.690 \small{± 0.004} & \cellcolor{score1!50} 6.233 \small{± 0.008} & 0.670 \small{± 0.001} & \cellcolor{bit4!50} 0.663 \small{± 0.007} & \cellcolor{score4!50} 0.562 \small{± 0.008} \\

        % + WorldCup ($k=1$) & 0.668 \small{± 0.000} & \cellcolor{bit3!50} 0.623 \small{± 0.000} & \cellcolor{score1!50} 0.417 \small{± 0.001} & 0.671 \small{± 0.000} & \cellcolor{bit2!50} 0.627 \small{± 0.001} & \cellcolor{score1!50} 0.318 \small{± 0.001} & 0.831 \small{± 0.004} & \cellcolor{bit4!50} 0.760 \small{± 0.001} & \cellcolor{score1!50} 6.182 \small{± 0.003} & 0.676 \small{± 0.002} & \cellcolor{bit4!50} 0.659 \small{± 0.005} & \cellcolor{score1!50} 0.630 \small{± 0.010} \\

        + \faToggleOn \textbf{ WorldCup } & 0.673 \small{± 0.002} & \cellcolor{bit1!50} 0.713 \small{± 0.002} & \cellcolor{score1!50} 0.401 \small{± 0.003} & 0.668 \small{± 0.001} & \cellcolor{bit1!50} 0.664 \small{± 0.001} & \cellcolor{score3!50} 0.312 \small{± 0.001} & 0.886 \small{± 0.001} & \cellcolor{bit1!50} 0.800 \small{± 0.002} & \cellcolor{score2!50} 6.136 \small{± 0.017} & 0.678 \small{± 0.005} & \cellcolor{bit1!50} 0.755 \small{± 0.010} & \cellcolor{score1!50} 0.645 \small{± 0.020} \\
        
        \bottomrule[1.5pt]
    \end{tabular}}
    \label{Table: Downstream Task}
\end{table*}

%% file: 05_Analysis.tex
\begin{figure}[htbp]
\centering
\begin{minipage}[htbp]{0.5\textwidth}
\centering
\includegraphics[width=\textwidth]{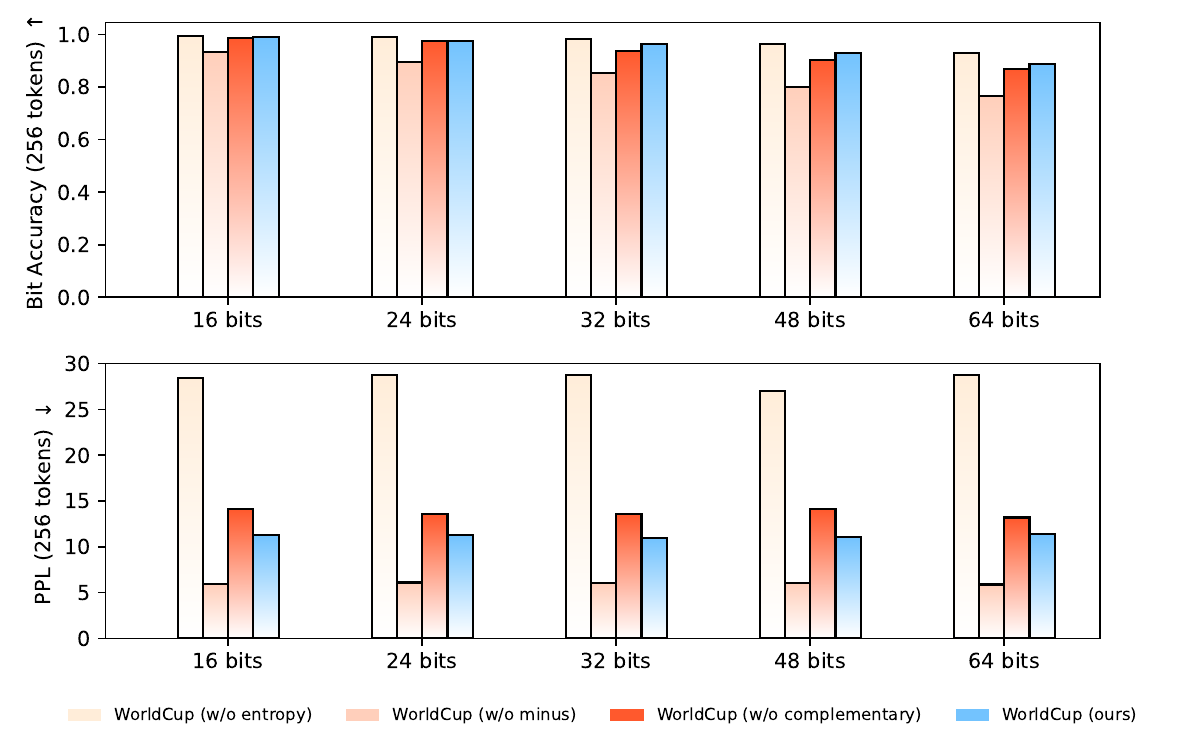}
\caption{The ablation study of WorldCup.}
\label{Figure: Ablation Study}
\end{minipage}
\begin{minipage}[htbp]{0.45\textwidth}
\centering
\includegraphics[width=\textwidth]{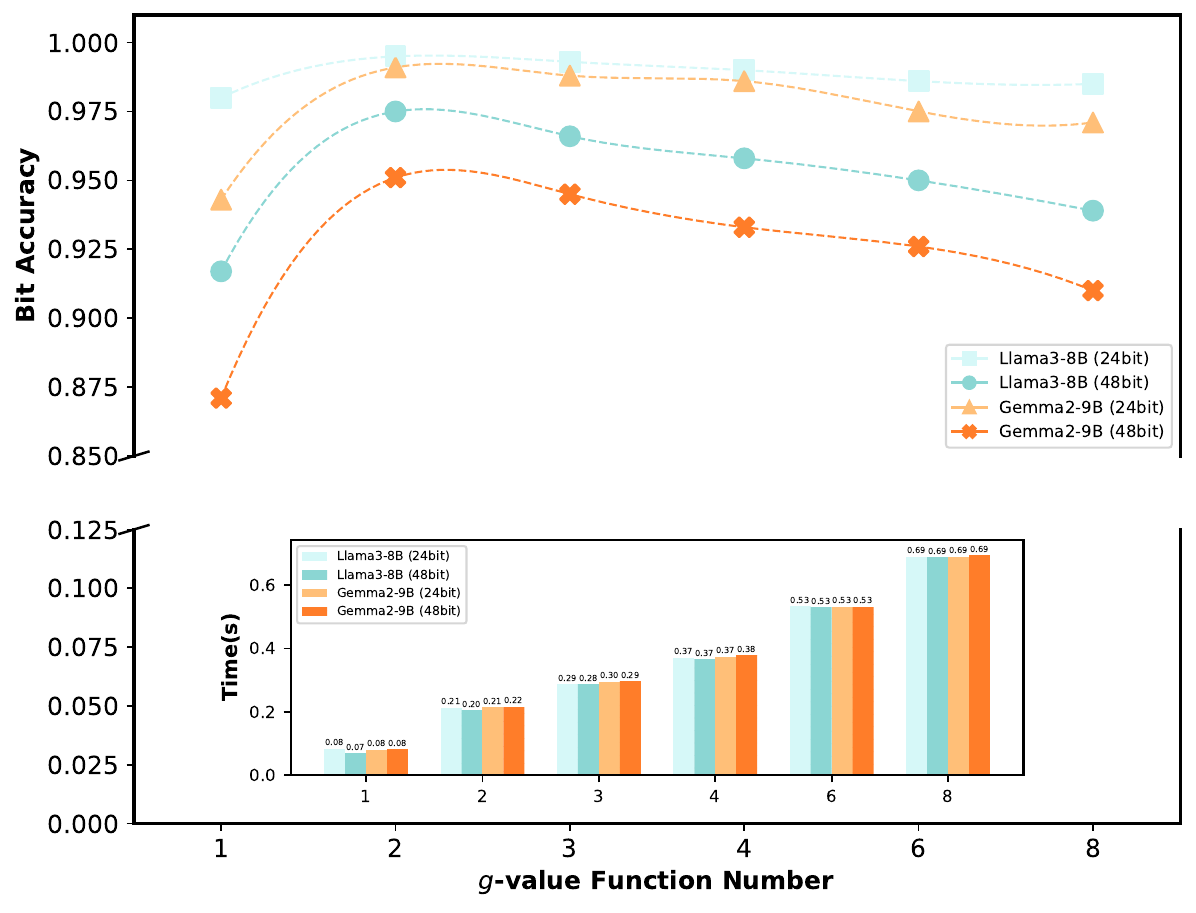}
\caption{The effect of varying $g$-value function number $k$ on accuracy and encoding time.}
\label{Figure: G-funtion Number}
\end{minipage}
\end{figure}

\section{Further Analysis} \label{Section: Further Analysis}

\subsection{Ablation Study} \label{Section: Ablation Study}

\paragraph{Settings.} To assess the contribution of each design in WorldCup, i.e., $q-\lambda \bar{q}$, where $q$ denotes the distribution induced by $k$ groups of $g$-value functions, we perform a component-wise ablation on LLaMA3-8B-Base. We evaluate three variants: (i) removing the entropy-aware factor $\lambda$ (\textbf{WorldCup w/o entropy}, i.e., $q-\bar{q}$); (ii) disabling the subtraction of complementary $g$-value distributions (\textbf{WorldCup w/o minus}, i.e., $q$); and (iii) replacing complementary $g$-value functions with randomly sampled ones (\textbf{WorldCup w/o complementary}, i.e., $q-\lambda q'$).

As shown in Fig.~\ref{Figure: Ablation Study}, the entropy-aware factor plays a crucial role in preserving text quality. Although it slightly reduces decoding accuracy, it substantially lowers the PPL of the generated text. In contrast, directly sampling from the original $g$-values without subtracting complementary ones yields relatively low PPL but poor discriminability, which significantly degrades decoding accuracy. Finally, when embedding either bit 0 or bit 1, using complementary $g$-value functions consistently outperforms random $g$-value functions in terms of both decoding accuracy and text quality. This observation is fully aligned with our theoretical analysis in Appendix~\ref{Appendix: Proof}. Overall, these results validate the necessity of each component and highlight their synergistic effect.

% \subsection{Token-level Information Capacity}
% We further investigate how the number of $g$-functions impacts performance on WorldCup under LLaMA3-8B and Gemma2-9B with 24-bit and 48-bit messages (Fig.~\ref{Figure: G-funtion Number}). Increasing $k$ does not monotonically improve performance: a single token cannot reliably encode arbitrarily many bits, and decoding accuracy reaches its optimum at $k=2$. Meanwhile, larger $k$ increases the number of concurrent sampling groups, inflating per-token generation cost. Striking a balance between accuracy and efficiency, we adopt $k=2$ as the default setting across all main experiments.

\subsection{Computational Cost Analysis} \label{Section: Computational Cost Analysis}

% For \textbf{encoding}, we evaluate the encoding time \emph{per generated token} and bit accuracy when embedding $k$ bits per token using $k$ groups of g-value functions, as shown in Fig.~\ref{Figure: G-funtion Number}. We observe that increasing $k$ leads to a roughly linear increase in encoding time. Specifically, generating a single token takes \textbf{0.08}s when $k = 1$, while the per-token time increases to \textbf{0.69}s when $k = 8$. Meanwhile, increasing $k$ does not monotonically improve decoding accuracy, as the information capacity of each token is inherently limited. In practice, the best performance is typically achieved at $k = 2$.

% For \textbf{decoding}, we quantify the \emph{per-sample decoding time}, defined as the time required to decode one complete generated text, including both z-score computation and multi-bit message recovery. The results are reported in Table~\ref{Table: Bit Acc}. WorldCup achieves the highest decoding efficiency, requiring approximately $\mathbf{1\sim2~\mathrm{ms}}$ on average to decode a single watermarked sample. This efficiency arises from confidence-aware decoding, which enables parallel aggregation of g-values and avoids token counting, allowing WorldCup to scale efficiently to longer sequences and higher payloads.

For \textbf{encoding}, we measure encoding time and bit accuracy when embedding $k$ bits \emph{per token} using $k$ groups of $g$-value functions (Fig.~\ref{Figure: G-funtion Number}). Encoding time increases roughly linearly with $k$: generating one token takes \textbf{0.08}s at $k=1$ and \textbf{0.69}s at $k=8$ for sequences of length 256 tokens. However, decoding accuracy does not improve monotonically with $k$, as each token has limited information capacity. In practice, the best performance is typically achieved at $k=2$. For \textbf{decoding}, we report the \emph{per-sample} decoding time, i.e., the time to decode a complete generated text, including $z$-score computation and multi-bit message recovery (Table~\ref{Table: Bit Acc}). WorldCup achieves the highest efficiency, requiring only  \textbf{0.01}$\sim$\textbf{0.02}s per sample on average. This gain comes from confidence-aware decoding, which enables fully parallel aggregation of $g$-values and eliminates token counting. As a result, WorldCup scales well to longer sequences and higher payloads, making it suitable for large-scale deployment.

%% file: 06_Conclusion.tex
\section{Conclusion}
% We presented WorldCup, a novel multi-bit watermarking framework for LLMs that inherits the strengths of tournament sampling. By combining complementary g-value functions with confidence-aware decoding, WorldCup effectively exploits token-level reliability to enable high-capacity and robust watermarking as well as great text quality. Empirical results across models and tasks demonstrate clear advantages over existing baselines. We hope that the insights behind WorldCup will support future research on scalable and reliable multi-bit LLM watermarking.

 % We introduced WorldCup, a novel multi-bit watermarking framework for LLMs that leverages the strengths of tournament sampling. By integrating complementary g-value functions with confidence-aware decoding, WorldCup effectively exploits token-level reliability to achieve high-capacity, robust watermarking without compromising text quality. Empirical evaluations across diverse models and tasks demonstrate its clear advantages over existing baselines. We hope that the principles underlying WorldCup will inspire future research on scalable and reliable multi-bit watermarking for LLMs and foster practical applications in AI content provenance and responsible model deployment.

We propose WorldCup, a multi-bit watermarking framework that views inference-time sampling as a communication channel for structured information embedding. By combining tournament-style sampling with entropy-aware modulation and confidence-aware decoding, it achieves high-capacity, robust watermarking without sacrificing text quality, outperforming prior methods in accuracy, efficiency, and robustness. We believe that WorldCup provides a principled foundation for scalable multi-bit watermarking and supports practical deployment for reliable LLM content attribution.

 % We introduced WorldCup, a novel multi-bit watermarking framework for LLMs that reframes inference-time sampling as a communication channel for structured information embedding. By combining tournament-style sampling with entropy-aware modulation and confidence-aware decoding, WorldCup enables high-capacity and robust watermarking without compromising text quality. Experiments across diverse models and tasks demonstrate clear advantages over existing baselines in accuracy, efficiency, and robustness. We believe that WorldCup provides a principled foundation for scalable multi-bit watermarking and supports practical deployment for reliable LLM content attribution.

%% file: 07_Appendix.tex
\section{Related Work} \label{Appendix: Related Work}
Existing LLM watermarking methods can be broadly categorized into zero-bit watermarking and multi-bit watermarking approaches~\citep{liu2024survey}, depending on whether message bits are explicitly embedded.

\subsection{Zero-bit Watermarking} 
The pioneering KGW method \citep{pmlr-v202-kirchenbauer23a} first introduced token-level watermarking by partitioning the vocabulary into "green" and "red" token lists and modifying the logits distribution during inference to embed a watermark signal. To enhance robustness, Unigram \citep{zhao2024provable} adopted a globally fixed vocabulary partition, while subsequent works \citep{liu2024a, ren-etal-2024-subtle, hou-etal-2024-k, hou-etal-2024-semstamp, verma2026watermarking} leveraged semantic and frequency-based features to defend against editing attacks. To mitigate text quality degradation, several studies proposed entropy-based watermarking schemes \citep{lu-etal-2024-entropy, lee-etal-2024-wrote, wang-etal-2025-trade}, while others explored unbiased reweighting strategies \citep{wu2023dipmark, hu2023unbiased, chen-etal-2025-improved}. Additionally, \cite{dathathri2024scalable, Aaronson, fu-etal-2024-gumbelsoft} designed alternative token sampling mechanisms that preserve the original logits distribution. Overall, zero-bit watermarking can only determine whether a text contains a watermark, limited in broader scenarios such as content provenance tracing.

\subsection{Multi-bit Watermarking}
Existing work \cite{fernandez2023three, jiang2025credidcrediblemultibitwatermark} embedded multi-bit message by establishing a mapping between predefined watermark keys and message bits. \cite{wang2024towards, cohen2025watermarking} divide both the text and message into multiple independent blocks, sequentially encoding each message segment into a corresponding text block. Although these methods achieve moderate decoding accuracy, they require enumerating all candidate messages during decoding, resulting in high exponential computational complexity with respect to message length (i.e., $O(2^b)$ for $b$ bits). To overcome this limitation, \cite{zhang2024remark, xu2025robust, lau-etal-2024-waterfall} explore training-based and post-hoc approaches, while \cite{yu2025saemark} proposes a black-box method for generating sentence-level watermarks. In contrast, MPAC~\citep{yoo-etal-2024-advancing} assigns distinct message bits to different tokens via hash functions and extends the KGW framework accordingly. Building upon this idea, BiMark~\citep{feng2025bimark} and StealthInk~\citep{pmlr-v267-jiang25j}, \cite{zamir2024excuse, boroujeny2024multi} developed distortion-free multi-bit watermark variants, while \cite{qu2025provably} introduced error-correcting codes \citep{chao2024watermarking, fairoze2023publicly, li2025efficientuniversalwatermarkingllmgenerated} to further enhance decoding robustness. Despite these advances, current multi-bit watermarking techniques still struggle to jointly optimize capacity, decoding accuracy, efficiency, and text quality.  
\newcommand{\compareyes}{\makebox[1.25em][c]{{\color{ForestGreen} \ding{52}}}}
\newcommand{\compareno}{\makebox[1.25em][c]{{\color{BrickRed} \ding{55}}}}
\newcommand{\comparepartially}{\makebox[1.25em][c]{{\color{Dandelion} \textsf{\textbf{P}}}}}

\begin{table}[htbp]
\caption{The comparison between \textbf{WorldCup} and related work: yes (\compareyes), partial (\comparepartially), or no (\compareno).}
\resizebox{\linewidth}{!}{%
\begin{tabular}{ccccccc}
\toprule
Multi-bit Watermarking & Detectability & Text Quality & Robustness & Token Capacity & Accuracy & Efficiency \\

\cmidrule(r){1-1} \cmidrule(lr){2-2} \cmidrule(lr){3-3} \cmidrule(lr){4-4} \cmidrule(lr){5-5} \cmidrule(lr){6-6} \cmidrule(lr){7-7}

MPAC~\citep{yoo-etal-2024-advancing} & \compareyes & \comparepartially & \compareno & \compareyes & \compareyes & \compareyes \\

BiMark~\citep{feng2025bimark} & \compareyes & \compareyes & \comparepartially & \comparepartially & \compareyes & \compareyes \\

SegMark~\citep{qu2025provably} & \compareyes & \comparepartially & \compareyes & \compareyes & \compareyes & \compareno \\

StealthInk~\citep{pmlr-v267-jiang25j} & \compareyes & \compareyes & \compareno & \compareyes & \comparepartially & \compareyes \\

\textbf{WorldCup (Ours)} & \compareyes & \compareyes & \compareyes & \compareyes & \compareyes & \compareyes \\

\bottomrule
\end{tabular}
}

\label{Table}
\end{table}

\section{Baselines} \label{Appendix: Baselines}
For each baseline, we follow the configurations in original papers. The key hyperparameter settings are as follows:

\begin{itemize}[leftmargin=*]
    \item \textbf{BiMark}~\citep{feng2025bimark}: The base scaling factor $\tilde{\delta}$ is set to $1.0$, and the number of layers $d$ is $10$. The proportion $\gamma$ of green lists is $0.5$ and the window size is 2. The values of $\text{c}\_\text{key}$ and $\text{bit}\_\text{idx}\_\text{key}$ are 530773 and 283519, respectively.
    \item \textbf{MPAC}~\citep{yoo-etal-2024-advancing}: We adopt the lefthash scheme. The window size is 2, and the proportion$ \gamma$ of green lists is 0.5. A bias $\delta = 2.0$ is added to the logit scores of green tokens. The hash key is 15485863.
    \item \textbf{SegMark}~\citep{qu2025provably}: We use the RSBH scheme (balanced segment assignment with ECC). The window size is 2, the proportion $\gamma$ of green lists is 0.5, and a bias $\delta=2.0$ is added to the logit of green tokens. The salt key is 35317.
    \item \textbf{StealthInk}~\citep{pmlr-v267-jiang25j}: We use "simple$\_$3" seeding scheme with $H=1$ message chunks. The window size is 2, and the hash key is 15485863.
\end{itemize}

\section{Datasets} \label{Appendix: Datasets}
We follow previous work~\citep{tu-etal-2024-waterbench} to evaluate our multi-bit watermark method on the following datasets:

\begin{itemize}[leftmargin=*]
    \item \textbf{C4}~\citep{raffel2020exploring} dataset is a large-scale, high-quality English pretraining corpus constructed by Google from Common Crawl. After extensive cleaning and filtering to remove non-linguistic, low-quality, and duplicate content, it yields roughly 750 GB of clean English text. We use the processed version in \url{https://huggingface.co/datasets/allenai/c4}.

    \item \textbf{OpenGen}~\citep{krishna2023paraphrasing} dataset contains 3,000 two-sentence text blocks drawn from the validation split of WikiText-103~\citep{merity2016pointer}, with the subsequent 300 tokens written by human. We sample 200 instances from this dataset for our experiments. The dataset is in in \url{https://github.com/XuandongZhao/Unigram-Watermark}.

    \item \textbf{WMT}~\citep{barrault-etal-2019-findings} dataset is a widely recognized benchmark in machine translation, containing parallel corpora from diverse sources and covering multiple language pairs. For our downstream evaluation, we primarily collect 200 samples from the WMT'19 De-En subset, with decoding parameters set to max\_new\_tokens = 64 and min\_new\_tokens = 16 (\textbf{short input, short output}). We embed \textbf{2-bit} message into each generated sample. It can be found at the following link: \url{https://huggingface.co/datasets/wmt/wmt19/viewer/de-en/validation}.

    \item \textbf{CNN\_DailyMail}~\citep{hermann2015teaching} dataset is a large-scale English news corpus containing over 300,000 unique articles written by journalists from CNN and the Daily Mail. The current release supports both extractive and abstractive summarization. We collect 200 samples and prompt the model to produce a one-sentence summary, with max\_new\_tokens = 64 and min\_new\_tokens = 32 (\textbf{long input, short output}), and embed a \textbf{16-bit} message into each generated summary. Details of the dataset can be found at: \url{https://huggingface.co/datasets/abisee/cnn\_dailymail}.
        
    \item \textbf{ELI5}~\citep{fan-etal-2019-eli5} dataset is a long-form QA dataset sourced from the Reddit community “Explain Like I’m Five.” It contains 270k diverse questions that require multi-sentence, explanatory answers supported by web evidence. We get 200 data points and set max\_new\_tokens = 256 and min\_new\_tokens = 64 (\textbf{short input, long output}), and embed a \textbf{32-bit} message into each generated answer. We use the processed subset available at \url{https://github.com/THU-KEG/WaterBench/blob/main/data/WaterBench/2-1\_longform\_qa.jsonl}.

    \item \textbf{GSM8K}~\citep{cobbe2021training} dataset consists of 8,500 high-quality grade-school math word problems requiring 2–8 steps of reasoning, with answers expressed in natural language. For downstream evaluation, we sample 200 instances and use an 8-shot setting with max\_new\_tokens = 256 and min\_new\_tokens = 64 (\textbf{long input, long output}), and embed a \textbf{4-bit} message into each generated solution. The dataset is available at: \url{https://huggingface.co/datasets/openai/gsm8k}.

\end{itemize}

\section{Metrics} \label{Appendix: Metrics}
All evaluation metrics used in our experiments are described in detail below:

\begin{itemize}[leftmargin=*]
    \item \textbf{Bit Accuracy (Bit Acc.)} measures the proportion of correctly extracted bits across all samples. Let $\mathbf{m}^{(j)} = (m^{(j)}_1, \dots, m^{(j)}_L)$ denote the embedded message of length $L$ for the $j$-th sample, and $\hat{\mathbf{m}}^{(j)} = (\hat{m}^{(j)}_1, \dots, \hat{m}^{(j)}_L)$ be the corresponding extracted message. The Bit Accuracy over $n$ samples is defined as
    
    \begin{equation}
        \mathrm{Bit\ Acc.} = \frac{1}{nL} \sum_{j=1}^{n} \sum_{i=1}^{L} \mathbb{I}\big[\hat{m}^{(j)}_i = m^{(j)}_i\big]
    \end{equation}
    
    where $\mathbb{I}[\cdot]$ denotes the indicator function. It reflects fine-grained bit-level decoding performance, as shown in Table~\ref{Table: Bit Acc}.

    \item \textbf{Message Extracted Rate (ME Rate)} quantifies the probability of perfectly recovering the entire embedded message. A message is considered successfully extracted if and only if all its bits are correctly recovered. Formally, the Message Extracted Rate over $n$ samples is defined as:

    \begin{equation}
        \mathrm{ME\ Rate} =\frac{1}{n} \sum_{j=1}^{n}
\mathbb{I}\big[\hat{\mathbf{m}}^{(j)} = \mathbf{m}^{(j)}\big].
    \end{equation}

    It is a strict metric that penalizes any bit error and reflects end-to-end message recovery reliability, as shown in Table~\ref{Table: ME Rate}.

    \item \textbf{F1 Score} is the harmonic mean of precision and recall:
    \begin{equation}
        \mathrm{F1} = 2 \cdot \frac{\mathrm{Precision} \cdot \mathrm{Recall}}{\mathrm{Precision}+\mathrm{Recall}}
    \end{equation}
    \textbf{Best F1 Score} denotes the maximum F1 score obtained over a threshold sweep, commonly used for evaluating binary classifiers without fixing a specific decision threshold.

    \item \textbf{AUROC Curve} (Receiver Operating Characteristic) plots \textbf{TPR} (True Positive Rate) against \textbf{FPR} (False Positive Rate) under varying decision thresholds. The area under this curve (\textbf{AUC}) summarizes the watermark detector’s ranking ability, with values closer to 1 indicating stronger discriminative performance.

    \item \textbf{Perplexity (PPL)} measures how well a language model predicts predicts a given text. For a sequence $x_{1:T}$:
    \begin{equation}
        \mathrm{PPL} = \exp\!\left(-\frac{1}{T}\sum_{t=1}^T \log p(x_t \mid x_{<t})\right)
    \end{equation} 
    Lower perplexity implies more confident and accurate language modeling. We report median PPL rather than mean PPL, as it provides more stable estimates and is less sensitive to extreme values~\citep{pmlr-v267-jiang25j}.

    % \item \textbf{BERTScore} \citep{zhang2019bertscore} computes similarity between generated and reference text using contextual embeddings from pretrained transformers. It aligns tokens by cosine similarity, offering a semantic-aware evaluation of generation quality. We use BERTScore-F1 to assess performance in both long-form QA and text summarization tasks.

    \item  \textbf{BLEU}~\citep{papineni2002bleu} (Bilingual Evaluation Understudy) is a standard automatic metric that quantifies lexical similarity by computing n-gram precision between machine-generated translations and human reference texts, with a brevity penalty to discourage overly short outputs.

    \item \textbf{Pass@K}~\citep{chen2021evaluating} measures the probability that at least one of the k generated solutions is correct. In this work, we follow the standard setting and report the results of GSM8K dataset using \textbf{Pass@1}.

    \item \textbf{Cosine similarity} Cosine similarity computes the cosine of the angle between two vectors to measure their semantic similarity. We use Sentence-BERT~\citep{reimers-gurevych-2019-sentence} to obtain sentence embeddings and apply cosine similarity to quantify the semantic closeness of natural texts and AI-generated texts.

    \item \textbf{ROUGE Score}~\citep{lin2004rouge} measures overlap between generated and reference text. Among its variants, we use ROUGE-L, which computes the longest common subsequence (LCS) between the candidate and reference, capturing sentence-level structural similarity beyond contiguous n-gram overlap.

    \item \textbf{Log Diversity}  quantifies textual diversity by measuring n-gram uniqueness. For each n-gram length $n\in\{2,3,4\}$, we compute a diversity score 
    and aggregate them by taking the product of the three adjusted scores:

    \begin{equation}
        \text{LogDiversity} = -\log\left(\max\left(1 - \prod_{n\in\{2,3,4\}}\left(1 - \left(1 - \frac{\text{unique}_n}{\text{total}_n}\right)\!/100\right),\, e^{-20}\right)\right)
    \end{equation}
    
    This log transformation stabilizes the metric and yields higher scores for more diverse, less repetitive text.

    \item \textbf{GPT4 Score}~\citep{zheng2023judging} leverages GPT-4~\citep{achiam2023gpt} as an evaluator. The model is prompted to rate the quality, correctness, or faithfulness of generated text relative to a reference or specification. This human-aligned evaluation correlates closely with expert judgments. The scoring template we use is as follows:

\tcbset{colframe = black, colbacktitle=white!80!gray, coltitle=black, colback=white, fonttitle = \bfseries}
\begin{tcolorbox}[title = {GPT-4 Judge Template}] 
You are an strict text quality evaluator. Your task is to compare a candidate text against a reference text (which serves as the ground truth) and produce a single final quality score.

Evaluation Criteria:

1. Fluency (Naturalness): How natural, grammatical, and readable the candidate text is.

2. Adequacy: How well the candidate preserves the meaning of the reference.

3. Coherence: How logically consistent and well-structured the candidate text is.

4. Relevance: How well the content matches the intent and key information of the reference.

5. Style Consistency: How closely the candidate matches the tone and style of the reference.

Scoring:

- For each criterion, assign a score from 1 to 10.

- Compute the final score as the average of all criterion scores.

- Output ONLY the final numerical score (e.g., 3.8). Do not explain, justify, or output intermediate scores.

Reference Text: xxx\ \ Candidate Text: xxx\ \ Score:

\end{tcolorbox}

\end{itemize}

\section{Backbone Models} \label{Appendix: Models}

We primarily employ the following backbone models in our experiments:

\begin{itemize}[leftmargin=*]
    \item \textbf{LLaMA3} family~\citep{grattafiori2024llama} is developed by Meta,  built upon an optimized Transformer architecture, which includes both pre-trained and instruction-tuned generative text models with sizes of 8B and 70B parameters. Both the 8B and 70B variants adopt Grouped Query Attention (GQA) to improve inference scalability. In this paper, we use the LLaMA3-8B Base and LLaMA3.1-8B-Instruct version, as details can be found in \url{https://huggingface.co/collections/meta-llama/meta-llama-3}.
    \item \textbf{Gemma2} family~\citep{team2024gemma} is a series of lightweight open-source models released by Google, developed using the same research foundations and technologies as the Gemini models. These models are text-to-text, decoder-only LLMs that currently support English, and are suitable for a wide range of text generation tasks. In this paper, we maily use the Gemma2-9B-Base and Gemma2-9B-Instruct version, as details can be found in \url{https://huggingface.co/google/gemma-2-9b}.
    \item \textbf{Ministral} family~\citep{liu2026ministral} belongs to Mistral AI’s latest third-generation models released in 2025, which includes three state-of-the-art small, dense models (3B, 8B, and 14B). These models support applications that understand text, images, and logic across 40+ languages, and can be used for coding, collaboration, or document analysis. In this paper, we utilize the Ministral-8B-Instruct, as detailed can be found in    \url{https://huggingface.co/mistralai/Ministral-8B-Instruct-2410}.
    % \item \textbf{Vicuna} family \cite{vicuna2023} is an open-source instruction-tuned LLM family based on the LLaMA architecture, fine-tuned on user-shared conversations from ShareGPT and released in 2023 by the LMSYS Org team. Evaluations using GPT-4 as a judge show that Vicuna-13B achieves over 90\% of the response quality of OpenAI ChatGPT and Google Bard, while outperforming other models such as LLaMA and Stanford Alpaca in more than 90\% of the comparisons. In this paper, we employ the Vicuna-13b-v1.5 to compute Perplexity, as detailed can be found in \url{https://huggingface.co/lmsys/vicuna-7b-v1.5}.
    
	% \item \textbf{Qwen3} family \cite{yang2025qwen3} is the latest generation in the Qwen series of LLMs, offering a comprehensive suite of both dense and Mixture-of-Experts (MoE) variants. Trained on extensive datasets, Qwen3 demonstrates strong performance in reasoning, instruction following, agentic capabilities, and multilingual support. In this paper, we mainly use the Qwen3-14B-Instruct models. Details can be found in \hyperlink{https://huggingface.co/collections/Qwen/qwen3}{ https://huggingface.co/collections/Qwen/qwen3}.
    
\end{itemize}

\section{Attack Settings} \label{Appendix: Attack Settings}
To evaluate the robustness of the watermark, we design the following various attack scenarios:

\begin{itemize}[leftmargin=*]
    \item \textbf{Word-D (ratio = $\rho$)}: Randomly deletes a proportion $\rho$ of words from the watermarked text.
    \item \textbf{Word-S-Dict (ratio = $\rho$)}: Randomly replaces a proportion $\rho$ of words with their synonyms based on the WordNet~\citep{miller1995wordnet} lexical dictionary.
    \item \textbf{Word-S-BERT (ratio = $\rho$)}: Randomly substitutes a proportion $\rho$ of words with context-aware synonyms generated by a BERT-based~\citep{devlin-etal-2019-bert} model.
    \item \textbf{Copy–Paste ($n-\rho$)}: Randomly splits the watermarked text into $n$ segments and inserts them into non-watermarked text, such that the inserted non-watermarked content accounts for a total proportion $\rho$.
    \item \textbf{Translation (en–zh)}: Translates the watermarked text from English to Chinese and then back to English using a fine-tuned T5 translation model: \url{https://huggingface.co/utrobinmv/t5_translate_en_ru_zh_small_1024}.
    \item \textbf{Rephrase (GPT-4o)}: Rewrites the watermarked text using the GPT-4o API with the $\tau=0.7$.
	\item \textbf{Rephrase (DIPPER-1)}: Rephrases the watermarked text using the DIPPER model with setting one ($\text{lex\_diversity=}\rho_1\text{ order\_diversity=}\rho_2\text{, max\_new\_tokens=256, sent\_interval=1, top\_p=0.75}$).
    \item \textbf{Rephrase (DIPPER-2)}: Rephrases the watermarked text using the DIPPER model with setting two ($\text{lex\_diversity=}\rho_1'\text{ order\_diversity=}\rho_2'\text{, max\_new\_tokens=256, sent\_interval=1, top\_p=0.75}$).
\end{itemize}

\input{07_Appendix-Proof}

\input{07_Appendix-More_Results} 

\input{07_Appendix-Additional_Analysis}

\input{07_Appendix-Impact_Statement}

\input{07_Appendix-Limitations}

\clearpage
\input{07_Appendix-Case_Study}

%% file: 07_Appendix-Proof.tex
\section{Theoretical Results} \label{Appendix: Proof}
\subsection{Proof of Proposition~\ref{Proposition: g-value}}
\label{Appendix: Proof of Proposition} 

\paragraph{Proof.} Let the random variables $G_0=\mathbf{g}_0(\mathbf{x},r)$ and
$G_1=\mathbf{g}_1(\mathbf{x},r)$ denote the $g$-values used to encode message bits $0$ and $1$, respectively. We assume that $G_0$ and $G_1$ share the same marginal distribution $F_g$, which ensures identical token-wise bias strength under both
hypotheses.

\noindent\textbf{Step 1: Discriminability at the scoring level.} We first quantify the separation between the two encoding hypotheses at the scoring level by the expected squared difference:

\begin{equation}
    D \triangleq \mathbb{E}\!\left[(G_1-G_0)^2\right].
\end{equation}

Expanding this expression yields:

\begin{equation} 
    D=\mathbb{E}(G_1^2)+\mathbb{E}(G_0^2)-2\mathbb{E}[G_1G_0] \label{Equation: expectation}
\end{equation}

Let $\mu_i=\mathbb{E}[G_i]$ and $\sigma_i^2=\operatorname{Var}(G_i)$ for $i\in\{0,1\}$.
Rewriting Eq.~\ref{Equation: expectation} in mean–variance form gives

\begin{equation}
    D=\sigma^2_1+\sigma_0^2-2\operatorname{Cov(G_1,G_0)}+(\mu_1-\mu_0)^2 \label{mean-variance}
\end{equation} 

Under the identical-marginal assumption ($\mu_0=\mu_1=\mu$, $\sigma_0^2=\sigma_1^2=\sigma^2$), Eq.~\ref{mean-variance} simplifies to

\begin{equation}
    D=2\sigma^2-2\operatorname{Cov}(G_1,G_0) \label{covariance}
\end{equation}

For fixed marginals, $D$ is maximized by minimizing the covariance between $G_0$ and $G_1$. By the Cauchy-Schwarz inequality:

\begin{equation}
    \operatorname{Cov}(G_1,G_0)\geq -\sqrt{\operatorname{Var}(G_1)\operatorname{Var}(G_0)}=-\sigma^2
\end{equation}

with equality if and only if $G_1$ and $G_0$ are perfectly anti-correlated. Substituting this bound into Eq.~\ref{covariance} yields:

\begin{equation}
    D_{\max}=2\sigma^2-2(-\sigma^2)=4\sigma^2
\end{equation}

\noindent\textbf{Step 2: Consistency with equal-mean constraint.} Perfect anti-correlation implies $G_1 = a - G_0$ almost surely. Enforcing $\mathbb{E}[G_1]=\mathbb{E}[G_0]=\mu$ gives $a=2\mu$, and hence:

\begin{equation}
    G_1 = 2\mu - G_0.
\end{equation}

Under the symmetric Bernoulli setting used in our experiments ($\mu=0.5$), this reduces to the complementary construction $G_1 = 1 - G_0$.

\noindent\textbf{Step 3: Implication for embedding distributions.} While the above analysis operates at the level of $g$-values, it has direct implications for the resulting embedding distributions induced by tournament sampling. Each tournament round selects the higher-scoring token according to the corresponding $g$-value function. Therefore, for any fixed candidate set, the probability that two encoding hypotheses produce different winners is a monotonically increasing function of the separation between their underlying scores. In particular, perfect anti-correlation maximizes the probability that a token favored under $G_0$ is disfavored under $G_1$, and vice versa. As a result, the induced token distributions under message bits $0$ and $1$ are pushed toward opposite extremes of the sampling decision boundary. This maximizes the distinguishability of the resulting embedding distributions in terms of any decision-based statistical distance (e.g., total variation or hypothesis-testing
error).

\noindent Consequently, complementary $g$-values achieve the maximum possible discriminability between embedding distributions under the tournament sampling mechanism, completing the proof.

\subsection{Vectorized WorldCup Sampling} \label{Appendix: Vectorized WorldCup Sampling} 
In the multi-bit WorldCup watermark, once a message bit is assigned to each token, the corresponding $g$-value function is uniquely determined. Consequently, conditioning on the assigned message bits, the generation of each token is distributionally equivalent to that of a zero-bit watermark. This observation allows us to directly adopt the analytical framework of SynthID~\citep{dathathri2024scalable}.

Let $p(\cdot)$ denote the base language-model distribution over the vocabulary $V$. For any token $x_t$, random seed $r$, and $g$-value function $g_\ell(\cdot,r)$ at layer $\ell$, we define:

\begin{equation}
    \begin{aligned}
    p\!\left(\mathcal{V}^{=g_{\ell}(x_t,r)}\right)
    &:= \sum_{x\in \mathcal{V}:\, g_{\ell}(x,r)=g_{\ell}(x_t,r)} p(x),\\
    p\!\left(\mathcal{V}^{<g_{\ell}(x_t,r)}\right)
    &:= \sum_{x\in \mathcal{V}:\, g_{\ell}(x,r)<g_{\ell}(x_t,r)} p(x),\\
    p\!\left(\mathcal{V}^{\le g_{\ell}(x_t,r)}\right)
    &:= \sum_{x\in \mathcal{V}:\, g_{\ell}(x,r)\le g_{\ell}(x_t,r)} p(x).
    \end{aligned}
\end{equation}

\begin{theorem} \label{Theorem: single-layer WorldCup sampling}
    (Vectorized form, single-layer WorldCup sampling). Given a probability distribution $P_\Theta$ over $\mathcal{V}$, random seed $r\in \mathcal{R}$, $g$-value distribution $\mathbf{g}_0$ and $\mathbf{g}_1$, and the number of leaves $N\geq 2$, message $\mathbf{m}$, the watermarked distribution $q(\cdot\mid P_\Theta, r_t, m, N, \mathbf{m}, p,\mathbf{g}_0,\mathbf{g}_1)$ for $m=1$ is given by:

    \begin{equation}
            q(\cdot\mid P_\Theta, r_t, m, N, \mathbf{m}, p,\mathbf{g}_0,\mathbf{g}_1) =\left\{
            \begin{array}{ll}
            p\left(x_{t}\right)\left(\frac{p\left(V^{\leq g(x_t,r)
            }\right)^{N} - p\left(V^{<g(x_t,r)}\right)^N}{p\left(V=g(x_t,r)\right)}\right) & \text { if } p(x_t)\neq0 \\
            0 & \text{ if } p(x_t)=0
            \end{array}\right.
    \end{equation}

    where $g=\mathbf{g}_0=g_1^{(0)}$ if $\mathbf{m}[p]=0$ else $g=\mathbf{g}_1=g_1^{(1)}$ when $m=1$
\end{theorem}

\paragraph{Proof.} We first note that if $p(x_t)=0$, then $\mathbb{P}(\text{Alg.~\ref{Algorthim: Binary WorldCup Watermarking} returns } x_t)=0$. Hence, we assume $p(x_t)>0$ in the following derivation. In a single-layer tournament, $N$ samples participate in the selection. Let $|Y^\ast|=j$ denote the number of tokens in the winning set, and suppose that $x_t$ appears $i$ times among these $j$ tokens (pairwise comparison is not required). The probability that outputs $x_t$ is:

\begin{equation}
    \begin{aligned}
    \mathbb{P}\!\left(\text{Alg.~\ref{Algorthim: Binary WorldCup Watermarking} returns } x_t\right)
    &=\sum_{j=1}^{N}\sum_{i=1}^{j}
    \mathbb{P}\!\left(
    |Y^\ast|=j,\,
    x_t \text{ appears } i \text{ times in } Y^\ast,\,
    \text{Alg.~\ref{Algorthim: Binary WorldCup Watermarking} returns } x_t
    \right)\\
    &=\sum_{j=1}^{N}\sum_{i=1}^{j}
    \binom{N}{j}
    p\!\left(V^{<g(x_t,r)}\right)^{N-j}
    \binom{j}{i}
    p(x_t)^i
    p\!\left(V^{=g(x_t,r)}\setminus x_t\right)^{j-i}
    \frac{i}{j}.
    \end{aligned}
\end{equation}

Rearranging the summations yields:

\begin{equation}
    \mathbb{P}\!\left(\text{Alg.~\ref{Algorthim: Binary WorldCup Watermarking} returns } x_t\right)=\sum_{j=1}^{N}\binom{N}{j}p\!\left(V^{<g(x_t,r)}\right)^{N-j}\sum_{i=1}^{j}\binom{j}{i}\frac{i}{j}p(x_t)^ip\!\left(V^{=g(x_t,r)}\setminus x_t\right)^{j-i}.
\end{equation}

\begin{lemma}
\begin{equation}
    \sum_{i=1}^{j}\binom{j}{i}\frac{i}{j}a^{i}b^{\,j-i}=a\,(a+b)^{j-1}
\end{equation} \label{Lemma}
\end{lemma}

Using the identity~\ref{Lemma}, and letting $a=p(x_t)$ and
$b=p\!\left(V^{=g(x_t,r)}\setminus x_t\right)$, we obtain:

\begin{equation}
    \sum_{i=1}^{j}\binom{j}{i}\frac{i}{j}p(x_t)^ip\!\left(V^{=g(x_t,r)}\setminus x_t\right)^{j-i}=p(x_t)\,p\!\left(V^{=g(x_t,r)}\right)^{j-1}.
\end{equation}

Substituting back gives:

\begin{equation}
\begin{aligned}
\mathbb{P}\!\left(\text{Alg.~\ref{Algorthim: Binary WorldCup Watermarking} returns } x_t\right)
&=
\sum_{j=1}^{N}
\binom{N}{j}
p\!\left(V^{<g(x_t,r)}\right)^{N-j}
p(x_t)\,
p\!\left(V^{=g(x_t,r)}\right)^{j-1}\\
&=
\frac{p(x_t)}{p\!\left(V^{=g(x_t,r)}\right)}
\sum_{j=1}^{N}
\binom{N}{j}
p\!\left(V^{<g(x_t,r)}\right)^{N-j}
p\!\left(V^{=g(x_t,r)}\right)^{j}.
\end{aligned}
\end{equation}

Applying the binomial theorem finally yields:

\begin{equation}
    \mathbb{P}\!\left(\text{Alg.~\ref{Algorthim: Binary WorldCup Watermarking} returns } x_t\right)=\frac{p(x_t)}{p\!\left(V^{=g(x_t,r)}\right)}\Big(p\!\left(V^{\le g(x_t,r)}\right)^N-p\!\left(V^{<g(x_t,r)}\right)^N\Big).
\end{equation}

In particular if $N=2$, then:

\begin{equation}
    \mathbb{P}\!\left(\text{Alg.~\ref{Algorthim: Binary WorldCup Watermarking} returns } x_t\right)=p(x_t)\left[p(x_t)+2p(V^{< g(x_t,r)})\right]
\end{equation}

When the $g$-value distribution $\mathbf{g}$ is binary (i.e., $g\in\{0,1\}$), the watermark distribution induced by a single-layer tournament with $N$ candidates is given by

\begin{equation}
    q(\cdot\mid P_\Theta, r_t, m, N, \mathbf{m}, p,\mathbf{g}_0,\mathbf{g}_1)=
    \begin{cases}
    p(x_t)\, p\!\left(V^{g=0}\right)^{N-1},
    & \text{if } g(x_t,r)=0,\\[6pt]
    p(x_t)\,\dfrac{1-p\!\left(V^{g=0}\right)^{N}}
    {p\!\left(V^{g=1}\right)},
    & \text{if } g(x_t,r)=1,
    \end{cases}
\end{equation}

where $p\!\left(V^{g=0}\right) :=\sum_{x\in V:\,g(x,r)=0}p(x)$ and $p\!\left(V^{g=1}\right) :=\sum_{x\in V:\,g(x,r)=1}p(x).$

By a straightforward induction on the number of layers, the above result extends directly to the $m$-layer WorldCup sampling:

\begin{theorem}
    (Vectorized form, multi-layer WorldCup sampling). Given a single-layer WorldCup sampling distribution $q(P_\Theta,g(\cdot),N)$:
    \begin{equation} \label{Equation: multi-layer WorldCup sampling}
        \begin{aligned}
        q^{(1)}(\cdot) &:=q(\cdot\mid P_\Theta, r_t, m, N, \mathbf{m}, p,\mathbf{g}_0,\mathbf{g}_1)\\
        q^{(2)}(\cdot) &:=q(\cdot\mid q^{(1)}, r_t, m, N, \mathbf{m}, p,\mathbf{g}_0,\mathbf{g}_1)\\
        \ldots \\
        q^{(m)}(\cdot) &:=q(\cdot\mid q^{(m-1)}, r_t, m, N, \mathbf{m}, p,\mathbf{g}_0,\mathbf{g}_1)\\
        \end{aligned}
    \end{equation}
    It follows that $q^{(m)}(\cdot)$ is equal to the $m$-layer WorldCup watermarked distribution $q(\cdot\mid Alg.~\ref{Algorthim: Binary WorldCup Watermarking} \ \operatorname{return}\ x_t)$
\end{theorem}

\paragraph{Proof.} The above theorem follows straightforwardly by induction on $m$. The case $m=1$ is ensured by Theorem~\ref{Theorem: single-layer WorldCup sampling}. Assume that the statement holds for $m-1$. For an $m$-layer tournament, we may equivalently first execute $N$-many ($m-1$)-layer tournaments and then apply a single-layer tournament to the resulting winners via $g_m(\cdot, r)$. By the induction assumption, the $N$ winners are drawn from $q^{(m-1)}(\cdot)$ as defined in Eq.~\ref{Equation: multi-layer WorldCup sampling}, and by Theorem~\ref{Theorem: single-layer WorldCup sampling} the winner of the single-layer tournament is given by $q(\cdot\mid q^{(m-1)}, r_t, m, N, \mathbf{m}, p,\mathbf{g}_0,\mathbf{g}_1)$.

% This establishes that, conditioned on the embedded message bits, the distribution induced by WorldCup sampling coincides with that of a zero-bit watermark, thereby providing a unified theoretical basis for subsequent detection and analysis.

\subsection{$k$-bit WorldCup Watermarking} \label{Appendix: $k$-ray WorldCup Watermarking}
We can naturally extend Equation~\ref{Equation: 2-bit WorldCup} to $k>2$ bits. Let $\mathbf{m}'\in\{0,1\}^k$ denote a $k$-bit message, and let $\{q_0, q_1, \ldots, q_{k-1}\}$ be the corresponding $g$-value functions $\mathbf{g}_i$ or probability subsets associated with each bit. Then, the unnormalized watermark distribution for a message $\mathbf{m}'$ can be written as:

\begin{equation}
    P_{\Theta,\mathbf{m}} = \sum_{i=0}^{k-1} \big( b_i \overline{q}_i + (1-b_i) q_i \big)
- \lambda \sum_{i=0}^{k-1} \big( b_i q_i + (1-b_i) \overline{q}_i \big)
\end{equation}

where $b_i$ is the $i$-th bit of $\mathbf{m}'$, and $\overline{q_i}$ is the distribution induced by the complementary $g$-value function $\bar{\mathbf{g}}_i$.

\subsection{Compute Watermark Z-score}  \label{Appendix: Compute Watermark Z-score} 
Let $\mathbf{y}=(y_1,\ldots,y_T)$ denote a generated text of length $T$ produced by an $m$-layer WorldCup sampling scheme. As in multi-bit watermarking, each token $y_t$ is deterministically mapped, via the shared hash function and watermarking key, to a message bit position $p_t\in\{1,\ldots,b\}$. For each token $y_t$ and layer $\ell\in\{1,\ldots,m\}$, we evaluate complementary $g$-value functions $g_\ell(y_t,r)\in\{0,1\}$ and $\bar g_\ell(y_t,r)=1-g_\ell(y_t,r)$, and define the signed token–layer score:
\begin{equation}
d_{t,\ell}\triangleq g_\ell(y_t,r)-\bar g_\ell(y_t,r)
=2g_\ell(y_t,r)-1
\end{equation}

Rather than aggregating all tokens simultaneously, we compute the detection statistic in a bit-wise, cumulative manner. Let $\mathcal{T}_p=\{t: p_t=p\}$ denote the set of tokens assigned to bit position $p$, and define the cumulative token set up to position $p$ as $\mathcal{C}_p=\bigcup_{i=1}^p \mathcal{T}_i$. The confidence-aware statistic after incorporating the first $p$ bit positions is defined as
\begin{equation}
S_p(\mathbf{y})
=\frac{1}{m|\mathcal{C}_p|}
\sum_{t\in\mathcal{C}_p}\sum_{\ell=1}^m d_{t,\ell}
\end{equation}
As $p$ increases, token–layer contributions are progressively accumulated.

Under the null hypothesis $H_0$ of unwatermarked text, the complementary $g$-value functions are symmetric and exchangeable, yielding $\mathbb{E}_{H_0}[g_\ell(y_t,r)]=\mathbb{E}_{H_0}[\bar g_\ell(y_t,r)]=0.5$. Consequently, $\mathbb{E}_{H_0}[d_{t,\ell}]=0$ and thus $\mathbb{E}_{H_0}[S_p(\mathbf{y})]=0$ for all $p$. Moreover, since $d_{t,\ell}\in\{-1,+1\}$ with equal probability under $H_0$, we have $\operatorname{Var}_{H_0}(d_{t,\ell})=1$. Assuming approximate independence across token–layer pairs $(t,\ell)$, the variance of the cumulative statistic is $\operatorname{Var}_{H_0}(S_p(\mathbf{y}))=\frac{1}{m|\mathcal{C}_p|}$.

We therefore define a position-dependent standardized detection statistic
\begin{equation}
z_p
=\frac{S_p(\mathbf{y})}{\sqrt{1/(m|\mathcal{C}_p|)}}
=\sqrt{m|\mathcal{C}_p|}~S_p(\mathbf{y}),
\end{equation}
which measures the watermark strength after cumulatively incorporating all tokens mapped to the first $p$ bit positions. In particular, $z_b$ corresponds to the final $z$-score computed over the entire text.

By the central limit theorem, $z_p$ converges in distribution to $\mathcal{N}(0,1)$ under the null hypothesis $H_0$ for each p. For watermarked text, tokens aligned with the embedded message bits contribute biased scores within each bit position, causing the cumulative statistic to deviate from zero as additional positions are incorporated.  In multi-bit watermarking with symmetric signed scores $d_{t,\ell}\in\{-1,+1\}$, the overall mean bias may cancel out when the numbers of embedded 0- and 1-bits are balanced, making the final signed statistic $z_b$ close to zero. However, the watermark does not induce a uniform mean shift; instead, it creates structured, position-dependent biases. Consequently, detection remains effective when using two-sided or position-aware statistics, such as $|z_p|$, $\max_p |z_p|$, or the energy $\sum_{p=1}^b z_p^2$, which are robust to sign cancellation.

In practice, we find that replacing $d_{t,\ell}$ with a max-based heuristic, $d_{t,\ell}\triangleq\max(g_\ell(y_t,r),\bar g_\ell(y_t,r))$, often yields stronger empirical detection performance. While this hard selection improves the effective signal-to-noise ratio by suppressing noisy contributions, it introduces a positive bias under $H_0$ and therefore does not yield a properly calibrated $z$-score.

% By the central limit theorem, $z_p$ converges in distribution to $\mathcal{N}(0,1)$ under $H_0$ for each $p$. For watermarked text, however, tokens aligned with the embedded message bits contribute biased scores, causing the cumulative z-score to increase as additional bit positions are incorporated. This progressive accumulation explicitly reflects the multi-bit structure of the watermark and avoids premature averaging over unrelated token groups. In practice, we find that replacing $d_{t,\ell}$ with a max-based heuristic, $d_{t,\ell}\triangleq\max(g_\ell(y_t,r),\bar g_\ell(y_t,r))$, often yields stronger empirical detection performance. While this hard selection improves the effective signal-to-noise ratio by suppressing noisy contributions, it introduces a positive bias under $H_0$ and therefore does not yield a properly calibrated z-score.

%% file: 07_Appendix-More_Results.tex
\section{More Experimental Results}
\definecolor{top1}{HTML}{ed583a}
\definecolor{top2}{HTML}{69c9b1}

% \definecolor{top1}{HTML}{ff8ba1}
% \definecolor{top2}{HTML}{3ecdb2}

\begin{table*}[htbp]
    \centering
    \caption{The Comparison of different multi-bit watermark performance with ME Rate.}
    \resizebox{\textwidth}{!}{\begin{tabular}{c|l|cccccccccccccccc}
    \toprule[1.5pt]
    \multirow{3}{*}{\textbf{Bit Length}} & \multicolumn{1}{c|}{\multirow{3}{*}{\textbf{Watermark}}} & \multicolumn{8}{c}{\textsc{\textbf{Llama3-8B-Base}}} & \multicolumn{8}{c}{\textsc{\textbf{Gemma2-9B-Base}}}\\

    & & \multicolumn{4}{c}{\textsc{max 128 tokens}} & \multicolumn{4}{c}{\textsc{max 256 tokens}} & \multicolumn{4}{c}{\textsc{max 128 tokens}} & \multicolumn{4}{c}{\textsc{max 256 tokens}}\\
    
    \cmidrule(lr){3-6}
    \cmidrule(lr){7-10} 
    \cmidrule(lr){11-14}
    \cmidrule(lr){15-18}
    
    & & AUC $\uparrow$& ME Rate $\uparrow$ & PPL $\downarrow$ & Time (s) $\downarrow$
    & AUC $\uparrow$& ME Rate $\uparrow$ & PPL $\downarrow$ & Time (s) $\downarrow$
    & AUC $\uparrow$& ME Rate $\uparrow$ & PPL $\downarrow$ & Time (s) $\downarrow$
    & AUC $\uparrow$& ME Rate $\uparrow$ & PPL $\downarrow$ & Time (s) $\downarrow$ \\
    \midrule
    
    % - & - & - & - & - & - & - & - & - & - & - & - & - & - & - & - & - & - \\
    % \midrule
    
    \multirow{4}{*}{16 bits} 
    & \faToggleOff \ MPAC & 0.999 & 0.520 & 16.25 & 0.049 & 0.996 & 0.785 & 13.56 & 0.087 & 0.980 & 0.305 & 13.69 & 0.045 & 0.985 & 0.500 & 12.00 & 0.070 \\
    & \faToggleOff \ SegMark & 0.993 & 0.815 & 15.88 & 0.551 & 0.995 & \textbf{\color{top1}0.995} & 12.94 & 0.839 & 0.979 & 0.510 & 13.19 & 1.234 & 0.999 & \textbf{\color{top2}0.905} & 10.94 & 2.402 \\
    & \faToggleOff \ BiMark & 1.000 & 0.750 & 14.34 & 0.026 & 1.000 & 0.860 & 11.25 & 0.044 & 1.000 & 0.410 & 11.06 & 0.031 & 0.999 & 0.620 & 9.625 & 0.039 \\
    & \faToggleOn \textbf{ Ours ($k=1$)} & 1.000 & \textbf{\color{top2}0.915} & \textbf{\color{top1}7.938} & \textbf{\color{top1}0.007} & 1.000 & 0.970 & \textbf{\color{top1}6.563} & \textbf{\color{top1}0.012} & 0.998 & \textbf{\color{top2}0.710} & \textbf{\color{top1}7.250} & \textbf{\color{top1}0.007} & 0.994 & 0.860 & \textbf{\color{top1}6.344} & \textbf{\color{top1}0.012} \\
    & \faToggleOn \textbf{ Ours ($k=2$)} & 0.998 & \textbf{\color{top1}0.940} & \textbf{\color{top2}12.94} & \textbf{\color{top2}0.008} & 1.000 & \textbf{\color{top2}0.990} & \textbf{\color{top2}10.94} & \textbf{\color{top2}0.013} & 1.000 & \textbf{\color{top1}0.850} & \textbf{\color{top2}10.94} & \textbf{\color{top2}0.009} & 1.000 & \textbf{\color{top1}0.965} & \textbf{\color{top2}9.563} & \textbf{\color{top2}0.012} \\
    \midrule

    \multirow{4}{*}{24 bits}
    & \faToggleOff \ MPAC & 0.996 & 0.175 & 16.75 & 0.052 & 0.997 & 0.445 & 14.00 & 0.074 & 0.972 & 0.085 & 13.69 & 0.044 & 0.959 & 0.190 & 12.00 & 0.070 \\
    & \faToggleOff \ SegMark & 0.947 & 0.450 & 17.13 & 0.721 & 0.992 & \textbf{\color{top1}0.975} & 13.38 & 0.804 & 0.849 & 0.190 & 13.19 & 1.670 & 0.974 & \textbf{\color{top2}0.695} & 10.13 & 2.853 \\
    & \faToggleOff \ BiMark & 1.000 & 0.280 & 14.34 & 0.027 & 1.000 & 0.700 & \textbf{\color{top2}11.34} & 0.036 & 0.998 & 0.085 & 11.63 & 0.030 & 0.990 & 0.415 & 9.938 & 0.043 \\
    & \faToggleOn \textbf{ Ours ($k=1$)} & 0.999 & \textbf{\color{top2}0.505} & \textbf{\color{top1}7.938} & \textbf{\color{top1}0.008} & 0.996 & 0.820 & \textbf{\color{top1}6.625} & \textbf{\color{top2}0.016} & 0.998 & \textbf{\color{top2}0.280} & \textbf{\color{top1}7.375} & \textbf{\color{top1}0.008} & 0.993 & 0.610 & \textbf{\color{top1}5.844} & \textbf{\color{top1}0.013} \\
    & \faToggleOn \textbf{ Ours ($k=2$)} & 1.000 & \textbf{\color{top1}0.710} & \textbf{\color{top2}12.84} & \textbf{\color{top2}0.008} & 1.000 & \textbf{\color{top2}0.885} & \textbf{\color{top2}10.75} & \textbf{\color{top1}0.013} & 0.999 & \textbf{\color{top1}0.500} & \textbf{\color{top2}10.44} & \textbf{\color{top2}0.008} & 1.000 & \textbf{\color{top1}0.815} & \textbf{\color{top2}9.125} & \textbf{\color{top2}0.016} \\
    \midrule

    \multirow{4}{*}{32 bits}
    & \faToggleOff \ MPAC & 0.997 & 0.035 & 16.75 & 0.051 & 0.996 & 0.255 & 14.00 & 0.090 & 0.947 & 0.005 & 13.81 & 0.046 & 0.940 & 0.025 & 11.63 & 0.074 \\
    & \faToggleOff \ SegMark & 0.909 & \textbf{\color{top2}0.240} & 16.63 & 0.642 & 0.989 & \textbf{\color{top1}0.840} & 12.94 & 1.053 & 0.824 & 0.025 & 12.75 & 1.751 & 0.955 & \textbf{\color{top2}0.465} & 10.25 & 3.362 \\
    & \faToggleOff \ BiMark & 1.000 & 0.055 & 13.91 & 0.026 & 1.000 & 0.425 & 12.38 & 0.040 & 0.996 & 0.005 & 12.00 & 0.031 & 0.989 & 0.115 & 9.938 & 0.041 \\
    & \faToggleOn \textbf{ Ours ($k=1$)} & 0.996 & 0.100 & \textbf{\color{top1}8.250} & \textbf{\color{top1}0.008} & 1.000 & 0.565 & \textbf{\color{top1}6.547} & \textbf{\color{top2}0.014} & 0.996 & \textbf{\color{top2}0.060} & \textbf{\color{top1}7.125} & \textbf{\color{top1}0.008} & 0.998 & 0.355 & \textbf{\color{top1}6.250} & \textbf{\color{top2}0.015} \\
    & \faToggleOn \textbf{ Ours ($k=2$)} & 0.998 & \textbf{\color{top1}0.390} & \textbf{\color{top2}13.28} & \textbf{\color{top2}0.011} & 1.000 & \textbf{\color{top2}0.800} & \textbf{\color{top2}10.56} & \textbf{\color{top1}0.013} & 1.000 & \textbf{\color{top1}0.105} & \textbf{\color{top2}10.66} & \textbf{\color{top2}0.009} & 1.000 & \textbf{\color{top1}0.565} & \textbf{\color{top2}9.500} & \textbf{\color{top1}0.013} \\
    \midrule

    \multirow{4}{*}{48 bits}
    & \faToggleOff \ MPAC & 0.993 & 0.000 & 16.63 & 0.057 & 0.990 & 0.010 & 13.19 & 0.078 & 0.936 & 0.000 & 14.00 & 0.048 & 0.914 & 0.000 & 12.19 & 0.076 \\
    & \faToggleOff \ SegMark & 0.837 & 0.000 & 16.38 & 0.725 & 0.968 & \textbf{\color{top1}0.340} & 13.56 & 1.381 & 0.770 & 0.000 & 13.38 & 2.668 & 0.900 & \textbf{\color{top2}0.030} & 9.938 & 3.382 \\
    & \faToggleOff \ BiMark & 1.000 & 0.000 & 14.25 & 0.027 & 0.999 & 0.060 & 11.63 & 0.037 & 0.985 & 0.000 & 11.53 & 0.029 & 0.975 & 0.000 & 10.03 & 0.040 \\
    & \faToggleOn \textbf{ Ours ($k=1$)} & 0.998 & 0.000 & \textbf{\color{top1}7.750} & \textbf{\color{top1}0.009} & 1.000 & 0.140 & \textbf{\color{top1}6.688} & \textbf{\color{top1}0.014} & 0.983 & 0.000 & \textbf{\color{top1}7.000} & \textbf{\color{top1}0.009} & 0.985 & 0.010 & \textbf{\color{top1}5.938} & \textbf{\color{top1}0.013} \\
    & \faToggleOn \textbf{ Ours ($k=2$)} & 0.999 & \textbf{\color{top1}0.025} & \textbf{\color{top2}12.84} & \textbf{\color{top2}0.009} & 1.000 & \textbf{\color{top2}0.310} & \textbf{\color{top2}11.44} & \textbf{\color{top2}0.014} & 0.996 & \textbf{\color{top1}0.015} & \textbf{\color{top2}10.50} & \textbf{\color{top2}0.009} & 0.998 & \textbf{\color{top1}0.130} & \textbf{\color{top2}9.313} & \textbf{\color{top2}0.014} \\
    
    \bottomrule[1.5pt]
    \end{tabular}}
    \label{Table: ME Rate}
\end{table*}

\begin{figure}[htbp]
\centering
\includegraphics[width=\linewidth]{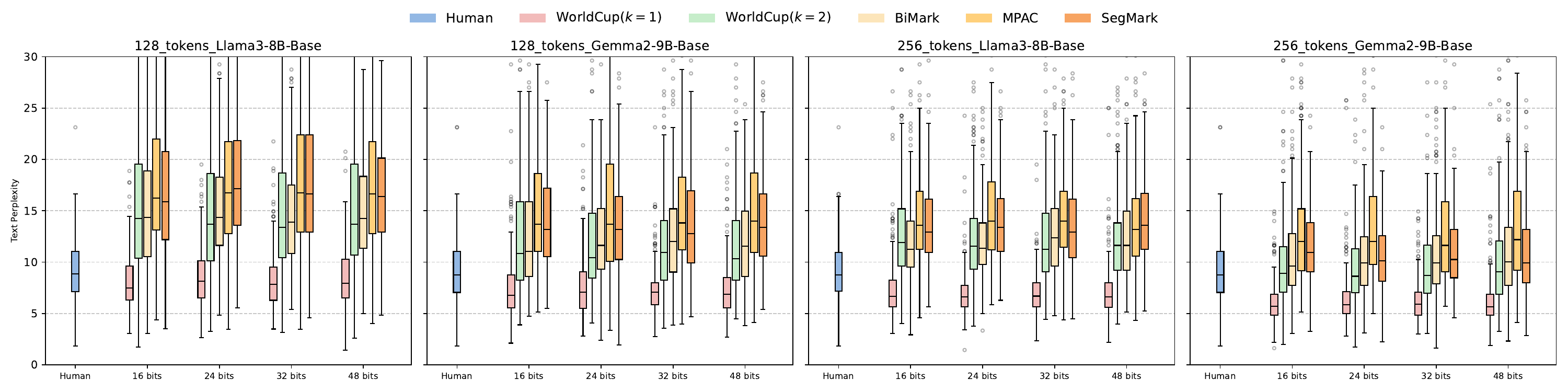}
\caption{The PPL of different multi-bit watermarking methods on LLaMA3-8B and Gemma2-9B.}
\label{Figure: PPL}
\end{figure}

\begin{figure}[htbp]
\centering
\includegraphics[width=\linewidth]{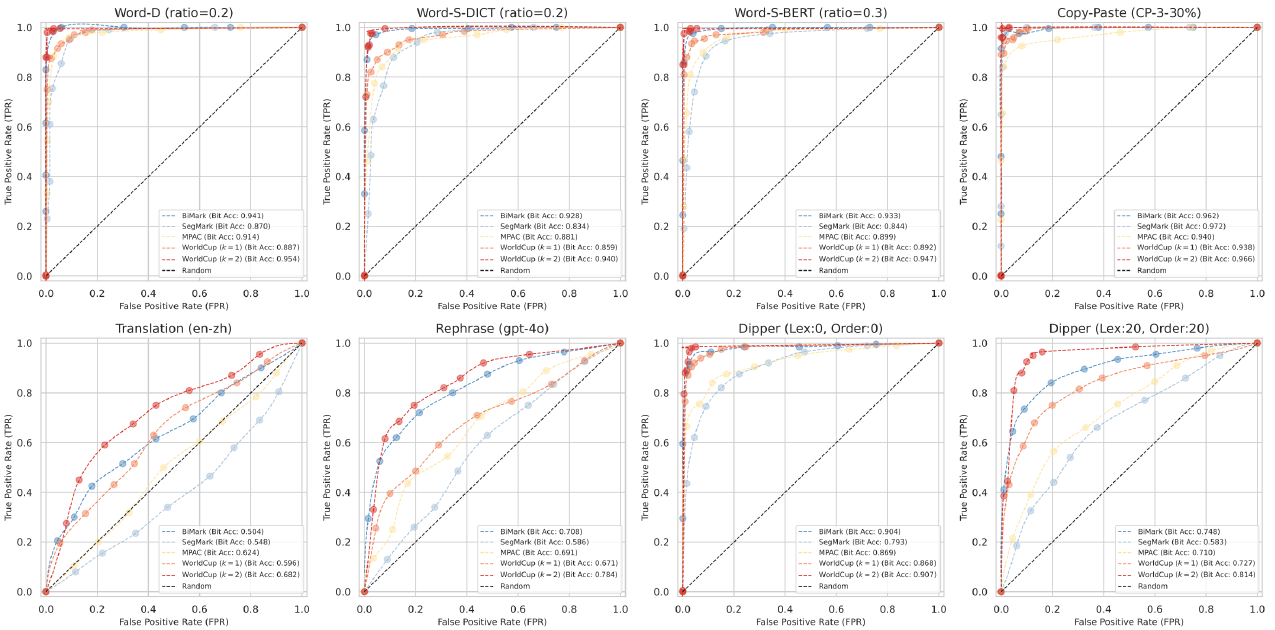}
\caption{The AUROC curves under different attacks on LLaMA3-8B-Base model (16 bits).}
\label{Figure: Llama3-Robustness-Appendix}
\end{figure}

\begin{figure}[htbp]
\centering
\includegraphics[width=\linewidth]{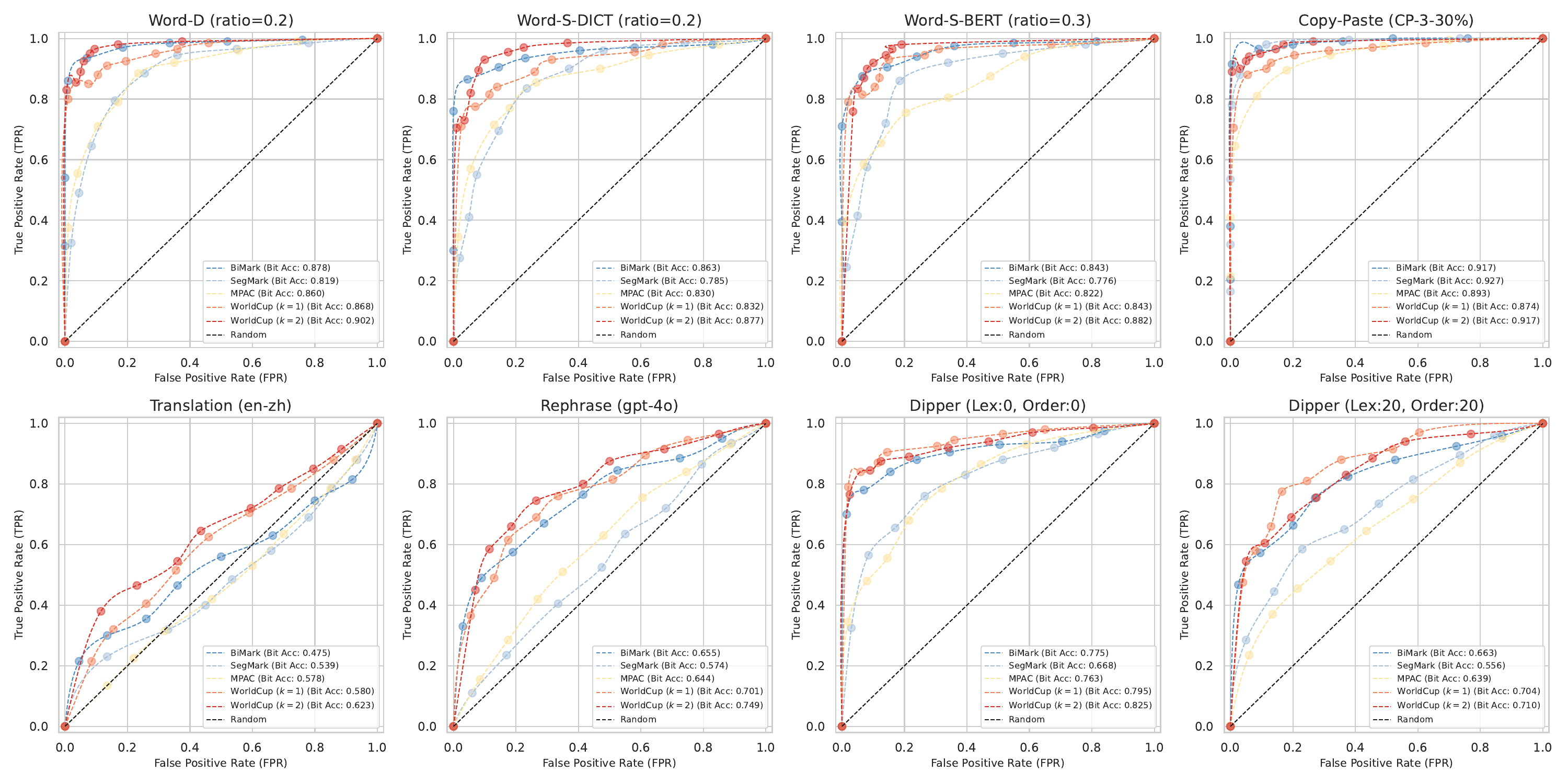}
\caption{The AUROC curves under different attacks on Gemma2-9B-Base model (16 bits).}
\label{Figure: Gemma2-Robustness-Appendix}
\end{figure}

To evaluate message-level decoding accuracy, we use the Message Extracted Rate (ME Rate). We further apply a Hamming error-correcting code~\citep{hamming1950error} to alleviate the sensitivity of ME Rate to a small number of bit errors. As reported in Table~\ref{Table: ME Rate}, WorldCup consistently outperforms existing baselines across most settings. Fig.\ref{Figure: PPL} shows that our method achieves more reliable detection while preserving better text quality. Additional robustness results (Fig.\ref{Figure: Llama3-Robustness-Appendix} and Fig.~\ref{Figure: Gemma2-Robustness-Appendix}) indicate that sentence-level attacks (e.g., back-translation and paraphrasing) are more harmful to multi-bit watermarking than word-level attacks (e.g., word deletion and synonym substitution). This is inherent to bit-allocation-based schemes, as sentence-level transformations disrupt the contextual alignment required to recover token-level bit information. Improving robustness to such attacks remains an open challenge.

% To measure message-level decoding accuracy, we evaluate the correctness of message extraction using the Message Extracted Rate (ME Rate). In addition, we incorporate a Hamming error-correcting code~\citep{hamming1950error}, which effectively mitigates the issue that a small number of bit errors in each sample can otherwise lead to a sharp degradation in ME Rate. As shown in Table~\ref{Table: ME Rate}, our WorldCup method consistently outperforms existing baseline approaches in most settings. Furthermore, the visualization of the perplexity (PPL) distribution in Fig.~\ref{Figure: PPL} clearly demonstrates that our method achieves higher detection reliability while maintaining relatively better text quality. Moreover, the results of additional robustness experiments are presented in Fig.~\ref{Figure: Llama3-Robustness-Appendix} and Fig.~\ref{Figure: Gemma2-Robustness-Appendix}. The results show that sentence-level attacks have a more pronounced impact on multi-bit watermarking, such as back-translation and paraphrasing, whereas word-level attacks, including word deletion and synonym substitution, are comparatively less destructive. This phenomenon is inherent to the bit-allocation-based paradigm: sentence-level transformations make it more difficult to correctly recover the bit information carried by each token from the surrounding context. Consequently, improving the robustness of multi-bit watermarking against sentence-level attacks remains a significant open challenge.

The numerical results corresponding to Fig.~\ref{Figure: Ablation Study} and Fig.~\ref{Figure: G-funtion Number} are reported in Table~\ref{Table: Ablation Study} and Table~\ref{Table: G-value Function Number}, respectively. Based on these experimental results, we select the hyperparameter $\alpha = 1.2$ and set the number of $g$-value functions to 2, as this configuration provides a favorable trade-off among bit accuracy, watermark detectability, and text quality.

\begin{table*}[htbp]
    \centering
    \caption{The numerical results of the ablation study on different components of WorldCup.}
    \resizebox{\textwidth}{!}{\begin{tabular}{l|ccccccccccccccc}
    \toprule[1.5pt]
    \multicolumn{1}{l}{\multirow{2}{*}{\textbf{Model}}} & \multicolumn{15}{c}{\textsc{\textbf{C4 Dataset}}} \\

    & \multicolumn{3}{c}{16 bits} & \multicolumn{3}{c}{24 bits} & \multicolumn{3}{c}{32 bits} & \multicolumn{3}{c}{48 bits} & \multicolumn{3}{c}{64 bits}\\
    
    \cmidrule(lr){2-4}
    \cmidrule(lr){5-7} 
    \cmidrule(lr){8-10}
    \cmidrule(lr){11-13}
    \cmidrule(lr){14-16}
    
    + Watermark & Best F1 $\uparrow$ & Bit Acc $\uparrow$ & PPL $\downarrow$ & Best F1 $\uparrow$ & Bit Acc $\uparrow$ & PPL $\downarrow$ & Best F1 $\uparrow$ & Bit Acc $\uparrow$ & PPL $\downarrow$ & Best F1 $\uparrow$ & Bit Acc $\uparrow$ & PPL $\downarrow$ & Best F1 $\uparrow$ & Bit Acc $\uparrow$ & PPL $\downarrow$ \\
    \midrule

    \textsc{\textbf{Llama3-8B-Base}} \\
    + WorldCup ($k=1$, ours) & 0.997 & 0.986 & 6.688 & 0.997 & 0.958 & 6.688 & 0.992 & 0.930 & 6.844 & 0.998 & 0.880 & 6.563 & 0.997 & 0.807 & 6.594 \\
    + WorldCup ($k=1$, w/o comp.) & 0.995 & 0.949 & 6.813 & 1.000 & 0.907 & 6.656 & 1.000 & 0.874 & 7.031 & 1.000 & 0.797 & 6.563 & 0.998 & 0.754 & 6.750 \\
    + WorldCup ($k=2$, w/o entropy) & 1.000 & 0.996 & 28.38 &  1.000 & 0.989 & 28.75 & 1.000 & 0.983 & 28.75 & 1.000 & 0.963 & 27.00 & 0.998 & 0.929 & 28.75 \\
    + WorldCup ($k=2$, w/o minus) & 0.940 & 0.933 & 5.906 & 0.946 & 0.895 & 6.094 & 0.910 & 0.853 & 6.063 & 0.884 & 0.802 & 6.063 & 0.870 & 0.765 & 5.875 \\
    + WorldCup ($k=2$, w/o comp.) & 0.995 & 0.986 & 14.13 & 0.995 & 0.975 & 13.56 & 0.995 & 0.937 & 13.56 & 0.992 & 0.904 & 14.13 & 0.987 & 0.868 & 13.19 \\
    + WorldCup ($k=2$, ours, $\lambda$) & 0.995 & 0.988 & 9.813 & 0.980 & 0.974 & 10.13 & 0.985 & 0.956 & 9.313 & 0.953 & 0.923 & 9.813 & 0.954 & 0.873 & 9.406 \\
    + WorldCup ($k=2$, ours, $1.2\lambda$) & 0.990 & 0.990 & 11.25 & 0.985 & 0.977 & 11.25 & 0.988 & 0.965 & 10.94 & 0.970 & 0.930 & 11.06 & 0.963 & 0.888 & 11.44 \\
    + WorldCup ($k=2$, ours, $1.5\lambda$) & 0.998 & 0.995 & 14.81 & 0.992 & 0.981 & 14.81 & 0.993 & 0.973 & 15.38 & 0.985 & 0.936 & 15.38 & 0.975 & 0.909 & 14.44 \\
    + WorldCup ($k=2$, ours, $2.0\lambda$) & 1.000 & 0.998 & 49.00 & 1.000 & 0.995 & 49.75 & 1.000 & 0.989 & 48.63 & 1.000 & 0.963 & 49.38 & 1.000 & 0.938 & 49.38 \\    
    \bottomrule[1.5pt]
    \end{tabular}}
    \label{Table: Ablation Study}
\end{table*}

\begin{table*}[htbp]
    \centering
    \caption{The effect of $g$-value function number on different LLMs.}
    \resizebox{\textwidth}{!}{\begin{tabular}{l|cccccccccccc}
    \toprule[1.5pt]
    \multicolumn{1}{c|}{\multirow{2}{*}{\textbf{Model}}} & \multicolumn{2}{c}{$g$\_function\_num=1} & \multicolumn{2}{c}{$g$\_function\_num=2} & \multicolumn{2}{c}{$g$\_function\_num=3} & \multicolumn{2}{c}{$g$\_function\_num=4} & \multicolumn{2}{c}{$g$\_function\_num=6} &
    \multicolumn{2}{c}{$g$\_function\_num=8} \\
    
    \cmidrule(lr){2-3}
    \cmidrule(lr){4-5} 
    \cmidrule(lr){6-7}
    \cmidrule(lr){8-9}
    \cmidrule(lr){10-11}
    \cmidrule(lr){12-13}
    
    & Bit Acc $\uparrow$ & Enc Time $\downarrow$ & Bit Acc $\uparrow$ & Enc Time $\downarrow$ & Bit Acc $\uparrow$ & Enc Time $\downarrow$ & Bit Acc $\uparrow$ & Enc Time $\downarrow$ & Bit Acc $\uparrow$ & Enc Time $\downarrow$ & Bit Acc $\uparrow$ & Enc Time $\downarrow$ \\
    \midrule

    \textsc{\textbf{Llama3-8B-Base}} & - & - & - & - & - & - & - & - & - & - & - & -\\

    WorldCup (24 bit) & 0.980 & 0.083 & 0.995 & 0.213 & 0.993 & 0.287 & 0.990 & 0.371 & 0.986 & 0.533 & 0.985 & 0.690 \\

    WorldCup (48 bit) & 0.917 & 0.069 & 0.975 & 0.203 & 0.966 & 0.285 & 0.958 & 0.366 & 0.950 & 0.530 & 0.939 & 0.689 \\

    \midrule

    \textsc{\textbf{Gemma2-9B-Base}} & - & - & - & - & - & - & - & - & - & - & - & -\\

    WorldCup (24 bit) & 0.943 & 0.079 & 0.991 & 0.214 & 0.988 & 0.296 & 0.986 & 0.374 & 0.975 & 0.532 & 0.971 & 0.691 \\

    WorldCup (48 bit) & 0.871 & 0.080 & 0.951 & 0.215 & 0.945 & 0.294 & 0.933 & 0.377 & 0.926 & 0.531 & 0.910 & 0.693 \\
    
    \bottomrule[1.5pt]
    \end{tabular}}
    \label{Table: G-value Function Number}
\end{table*}

%% file: 07_Appendix-Additional_Analysis.tex
\section{Additional Analysis} \label{Appendix: Additional Analysis}

\subsection{Entropy Analysis} \label{Appendix: Entropy Analysis}
\begin{figure}[htbp]
\centering
\begin{minipage}[t]{0.4\textwidth}
\centering
\includegraphics[width=\linewidth]{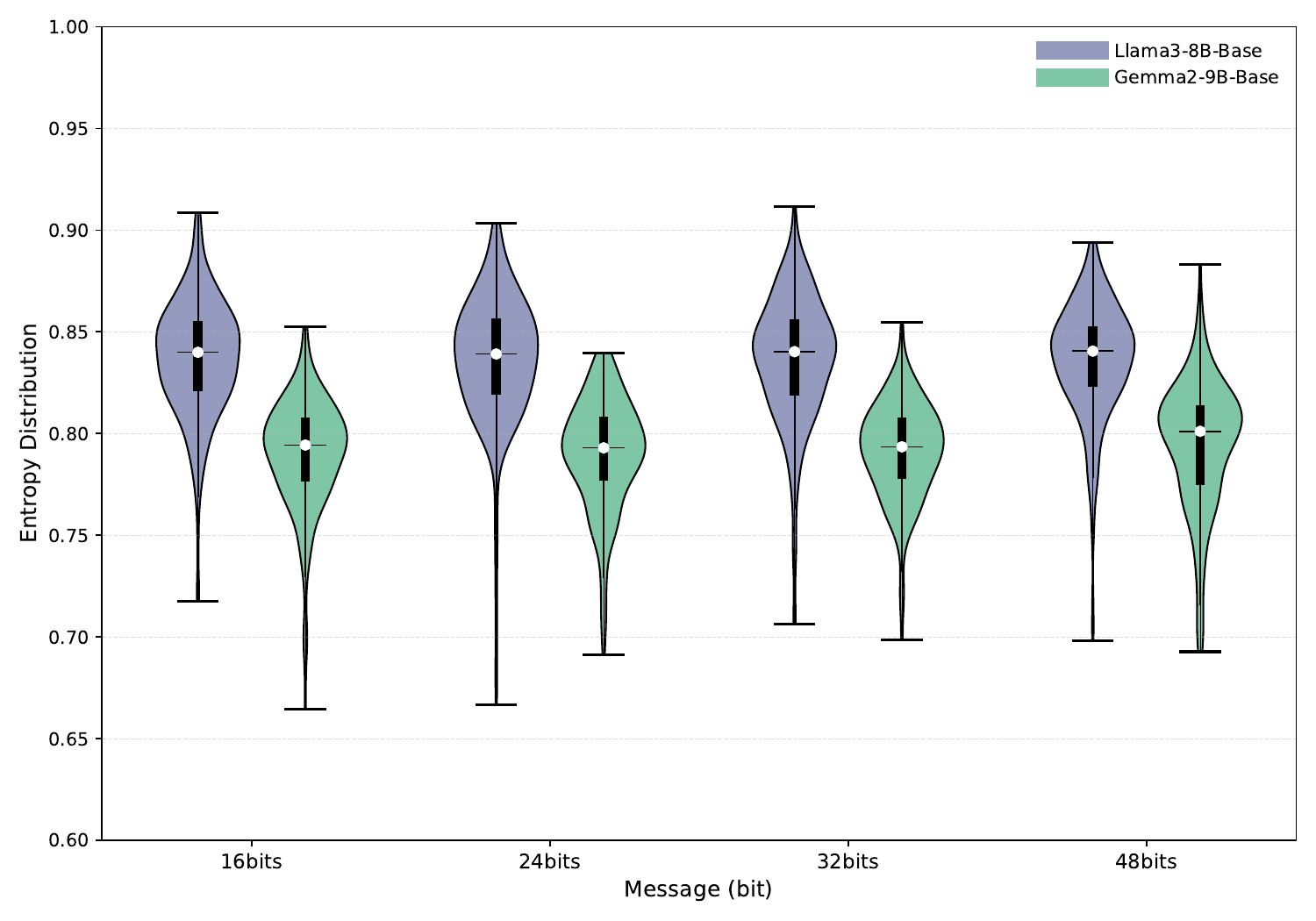}
\end{minipage}
\begin{minipage}[t]{0.4\textwidth}
\centering
\includegraphics[width=\linewidth]{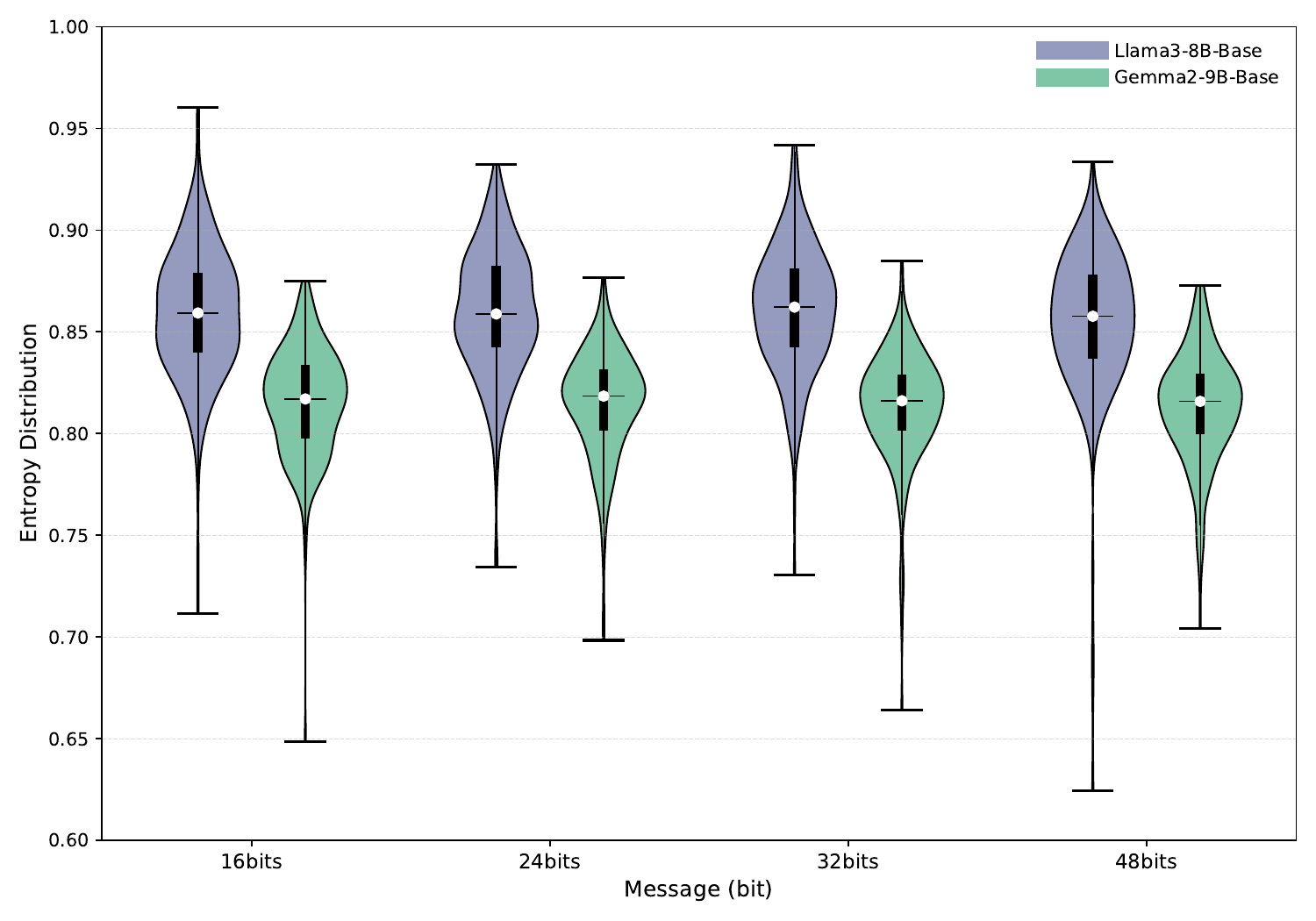}
\end{minipage}
\caption{The comparison of spike entropy between LLaMA3-8B-Base and Gemma2-9B-Base.}
\label{Figure: Entropy}
\end{figure}

From the results in Table~\ref{Table: Bit Acc}, we observe that existing multi-bit watermarking methods achieve substantially lower bit decoding accuracy on Gemma2-9B-Base than on LLaMA3-8B-Base. To better understand this phenomenon, we conduct an entropy-based analysis. Specifically, we visualize the average token-level entropy for each sample during text generation, as illustrated in Fig.~\ref{Figure: Entropy}. The results clearly indicate that Gemma2-9B-Base produces tokens with consistently lower entropy, aligning with prior findings that low-entropy text is inherently more challenging for watermark embedding. A plausible explanation is that Gemma2-9B-Base, owing to its larger parameter scale compared to LLaMA3-8B-Base, generates tokens with higher confidence, thereby reducing entropy and limiting the effective embedding capacity for multi-bit watermarks.

\subsection{Distortionary and Non-distortionary $g$-value Functions} \label{Appendix: Distortionary and Non-distortionary $g$-value Functions}
According to SynthID~\citep{dathathri2024scalable}, we define distortionary and non-distortionary $g$-value functions as follows:

\begin{definition}
    A sampling algorithm $\mathcal{S}:\Delta \mathcal{V}\times \mathcal{R}\rightarrow \mathcal{V}$ is (single-token) non-distortionary if for any probability distribution $p\in\Delta\mathcal{V}$ and token $x\in\mathcal{V}$:
    \begin{equation}
        \mathbb{E}_{r\sim\text{Unif}{(\mathcal{R})}}[\mathbb{P}(\mathcal{S}(p,r)=x)] = p(x)
    \end{equation}
    If $\mathcal{S}$ is not non-distortionary, we call it distortionary.
\end{definition}

\begin{figure}[htbp]
\centering
\begin{minipage}[t]{0.4\textwidth}
\centering
\includegraphics[width=\linewidth]{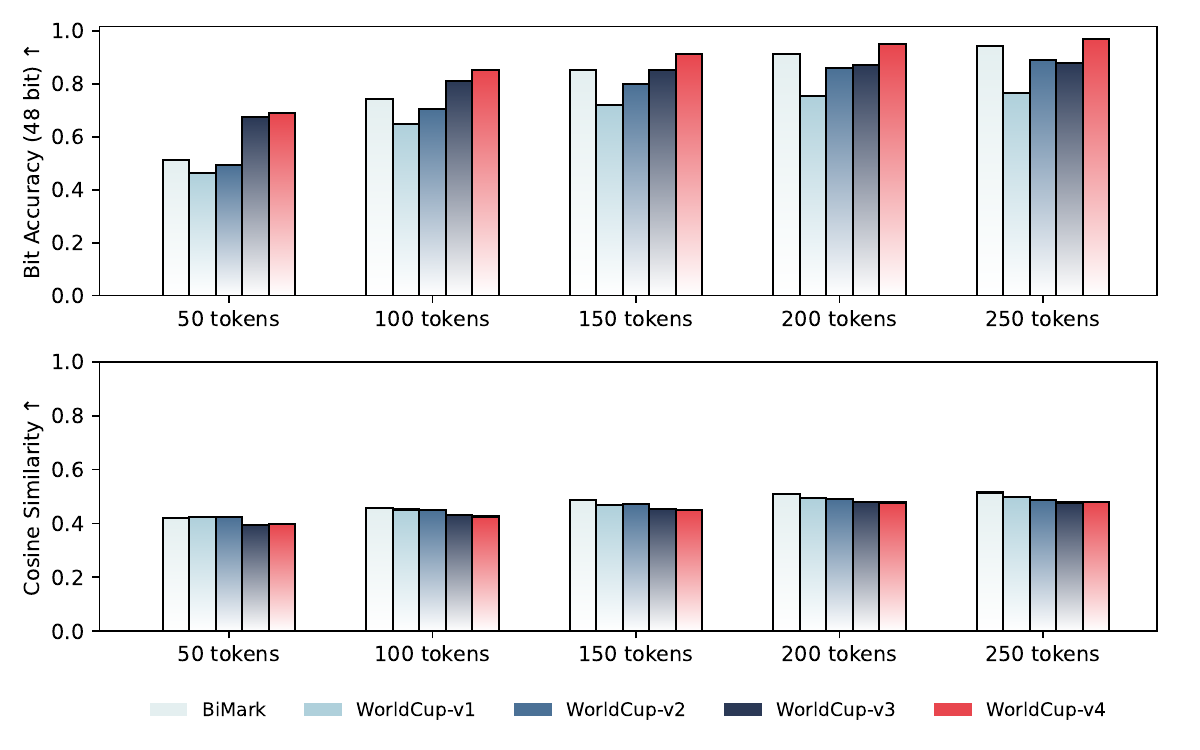}
\end{minipage}
\begin{minipage}[t]{0.4\textwidth}
\centering
\includegraphics[width=\linewidth]{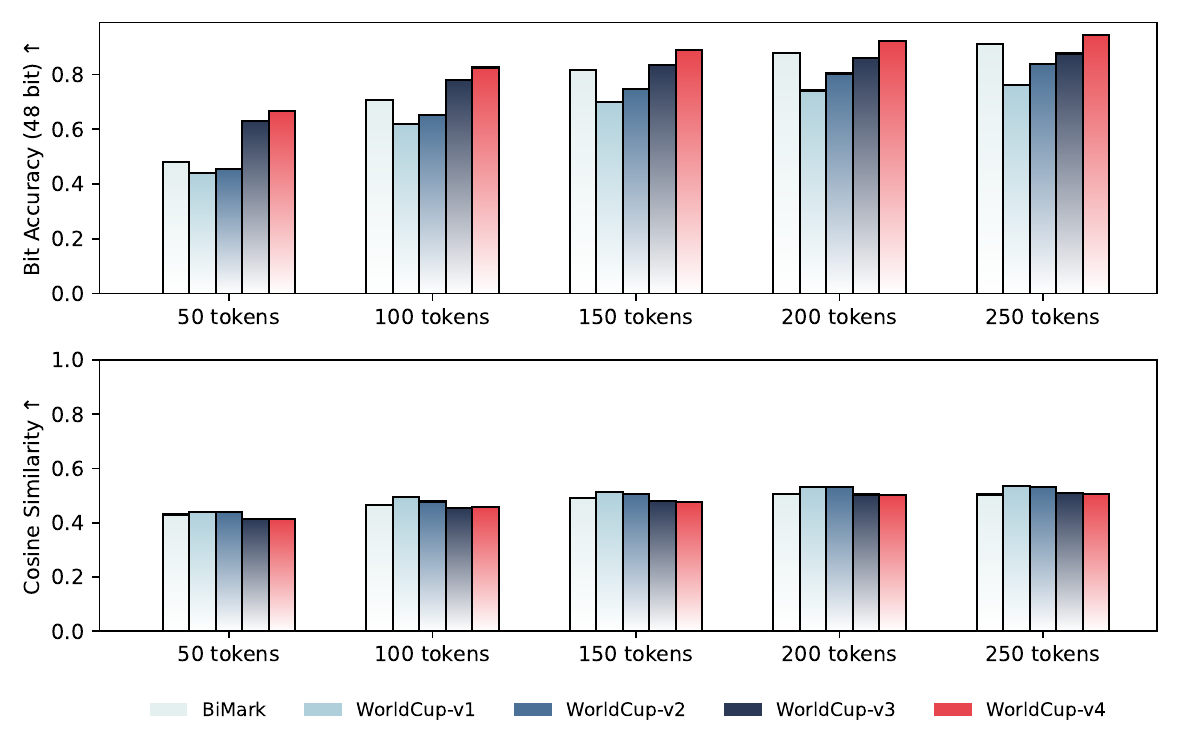}
\end{minipage}
\caption{The comparison of distortionary and non-distortionary $g$-value functions}
\label{Figure: Distortionary and Non-Distortionary}
\end{figure}

Therefore, following the design of non-distortionary and distortionary $g$-value functions in SynthID, we evaluate their performance in the multi-bit watermarking setting. The results are summarized in Fig.~\ref{Figure: Distortionary and Non-Distortionary} and Table~\ref{Table: Distortionary and Non-Distortionary}. Specifically, WorldCup-v1, v2, v3, and v4 in the figure correspond to the configurations ($k=1$, non-distortionary), ($k=1$, distortionary), ($k=2$, non-distortionary), and ($k=2$, distortionary) in the Table~\ref{Table: Distortionary and Non-Distortionary}, respectively. We observe that the non-distortionary setting yields higher cosine similarity between the generated watermarked text and the ground truth compared to the distortionary setting, at the cost of reduced bit accuracy. Consequently, in practical scenarios, the choice between these two settings can be made based on application-specific requirements. Unless otherwise stated, our experiments primarily adopt the distortionary configuration.

\begin{table*}[htbp]
    \centering
    \caption{The comparison of distortionary $g$-value function and non-distortionary $g$-value function.}
    \resizebox{\textwidth}{!}{\begin{tabular}{l|ccccccccccccccc}
    \toprule[1.5pt]
    \multicolumn{1}{l}{\multirow{2}{*}{\textbf{Model}}} & \multicolumn{15}{c}{\textsc{\textbf{C4 Dataset}}} \\

    & \multicolumn{3}{c}{50 tokens} & \multicolumn{3}{c}{100 tokens} & \multicolumn{3}{c}{150 tokens} & \multicolumn{3}{c}{200 tokens} & \multicolumn{3}{c}{250 tokens} \\
    
    \cmidrule(lr){2-4}
    \cmidrule(lr){5-7} 
    \cmidrule(lr){8-10}
    \cmidrule(lr){11-13}
    \cmidrule(lr){14-16}
    
    + Watermark & Best F1 $\uparrow$ & Bit Acc $\uparrow$ & Cos Sim $\uparrow$ & Best F1 $\uparrow$ & Bit Acc $\uparrow$ & Cos Sim $\uparrow$ & Best F1 $\uparrow$ & Bit Acc $\uparrow$ & Cos Sim $\uparrow$ & Best F1 $\uparrow$ & Bit Acc $\uparrow$ & Cos Sim $\uparrow$ & Best F1 $\uparrow$ & Bit Acc $\uparrow$ & Cos Sim $\uparrow$ \\
    \midrule

    \textsc{\textbf{Llama3-8B-Base}} \\
    + BiMark & 0.972 & 0.513 & 0.422 & 0.998 & 0.743 & 0.457 & 1.000 & 0.851 & 0.487 & 1.000 & 0.911 & 0.510 & 1.000 & 0.943 & 0.515 \\
    + WorldCup ($k=1$, non-distortionary) & 0.883 & 0.463 & 0.425 & 0.916 & 0.647 & 0.452 & 0.883 & 0.719 & 0.468 & 0.881 & 0.754 & 0.496 & 0.869 & 0.767 & 0.497 \\
    + WorldCup ($k=1$, distortionary) & 0.935 & 0.494 & 0.425 & 0.967 & 0.705 & 0.449 & 0.980 & 0.799 & 0.471 & 0.992 & 0.858 & 0.490 & 0.990 & 0.889 & 0.488 \\
    + WorldCup ($k=2$, non-distortionary) & 0.934 & 0.673 & 0.395 & 0.941 & 0.812 & 0.433 & 0.929 & 0.853 & 0.454 & 0.915 & 0.871 & 0.479 & 0.912 & 0.879 & 0.478 \\
    + WorldCup ($k=2$, distortionary) & 0.947 & 0.689 & 0.397 & 0.995 & 0.853 & 0.426 & 0.997 & 0.914 & 0.451 & 1.000 & 0.949 & 0.478 & 1.000 & 0.968 & 0.481 \\
    \midrule

    \textsc{\textbf{Gemma2-9B-Base}} \\
    + BiMark & 0.949 & 0.480 & 0.430 & 0.988 & 0.706 & 0.466 & 0.995 & 0.816 & 0.490 & 1.000 & 0.879 & 0.505 & 0.998 & 0.909 & 0.504 \\
    + WorldCup ($k=1$, non-distortionary) & 0.807 & 0.439 & 0.439 & 0.864 & 0.620 & 0.493 & 0.888 & 0.699 & 0.514 & 0.909 & 0.741 & 0.532 & 0.879 & 0.760 & 0.536 \\
    + WorldCup ($k=1$, distortionary) & 0.865 & 0.455 & 0.438 & 0.935 & 0.653 & 0.478 & 0.948 & 0.747 & 0.506 & 0.972 & 0.803 & 0.532 & 0.975 & 0.838 & 0.533 \\
    + WorldCup ($k=2$, non-distortionary) & 0.870 & 0.630 & 0.415 & 0.950 & 0.780 & 0.454 & 0.958 & 0.834 & 0.479 & 0.962 & 0.861 & 0.504 & 0.959 & 0.876 & 0.508 \\
    + WorldCup ($k=2$, distortionary) & 0.915 & 0.665 & 0.412 & 0.992 & 0.825 & 0.459 & 0.990 & 0.888 & 0.477 & 0.993 & 0.920 & 0.503 & 0.997 & 0.942 & 0.505 \\
    \bottomrule[1.5pt]
    \end{tabular}}
    \label{Table: Distortionary and Non-Distortionary}
\end{table*}

\subsection{Different Detectors} \label{Appendix: Different Detectors}
Following SynthID Text~\citep{dathathri2024scalable}, we consider two watermark detectors: mean detector (D1) and weighted-mean detector (D2).

The mean detector computes the average $g$-value across all tokens and all layers as
\begin{equation}
\text{MeanScore}(x) := \frac{1}{mT} \sum_{t=1}^{T} \sum_{\ell=1}^{m} g_{t,\ell},
\end{equation}
where $T$ denotes the number of tokens and $m$ the number of layers.

The weighted-mean detector assigns non-increasing weights $\alpha_1 \geq \cdots \geq \alpha_m \geq 0$ to different layers, with $\sum_{\ell=1}^{m} \alpha_\ell = m$:
\begin{equation}
\text{WeightedMeanScore}(x, \boldsymbol{\alpha}) := \frac{1}{mT} \sum_{t=1}^{T} \sum_{\ell=1}^{m} \alpha_\ell  g_{t,\ell}.
\end{equation}

Specifically, we set
$\alpha_1 = \tau,\quad
\alpha_2 = \tau - \frac{\tau - \mu}{m - 1},\quad
\alpha_3 = \tau - 2\frac{\tau - \mu}{m - 1},\ \ldots,\
\alpha_m = \mu$,
with $\tau = 10$ and $\mu = 1$, and then renormalize the weights such that $\sum_{\ell=1}^{m} \alpha_\ell = m$.

\begin{figure}[htbp]
\centering
\includegraphics[width=0.4\linewidth]{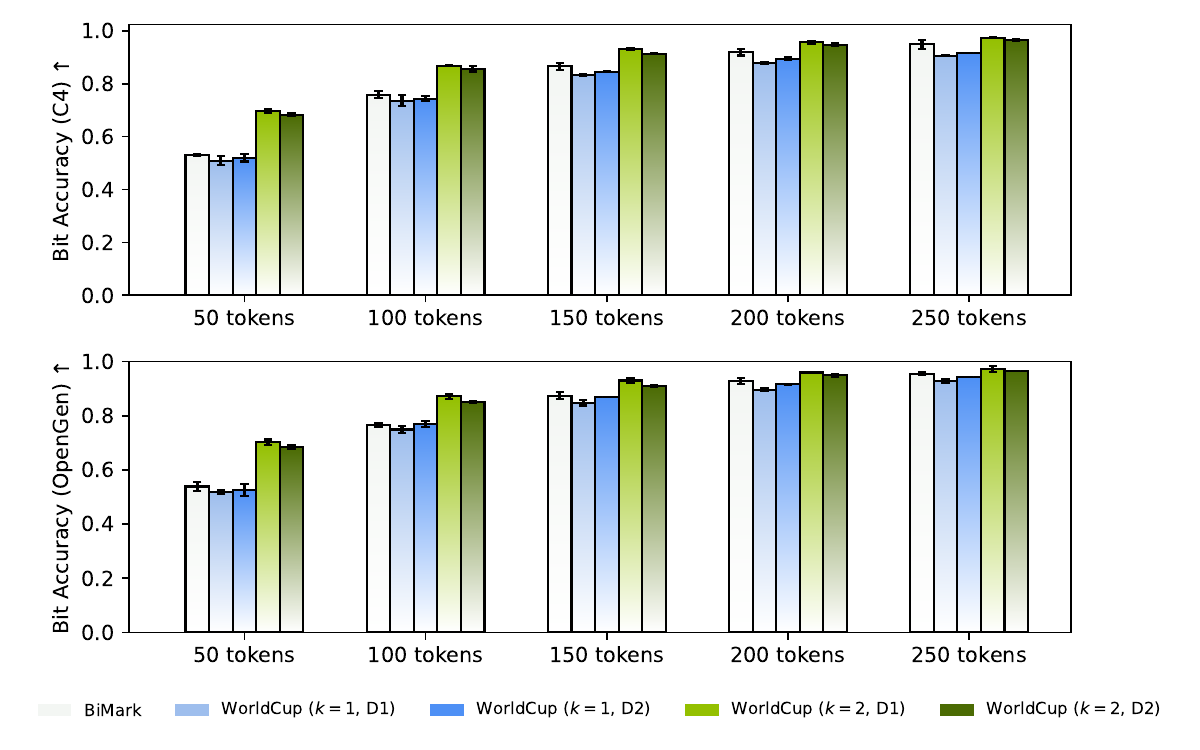}
\caption{The comparison of mean detector and weighted mean detector.}
\label{Figure: Weighted Mean Detector}
\end{figure}

We evaluate both detectors under $k=1$ and $k=2$ on the C4 and OpenGen datasets across varying token lengths. Figure~\ref{Figure: Weighted Mean Detector} shows the corresponding detection curves, and the quantitative results are summarized in Table~\ref{Table: Weighted Mean Detecor}. Our results indicate that for $k=1$, the weighted-mean detector consistently outperforms the mean detector, in agreement with the findings of SynthID. This is because the contribution of watermarking evidence from each layer diminishes as depth increases, making layer-wise weighting beneficial. However, for $k=2$, this trend reverses, and the mean detector achieves superior performance. A plausible explanation is that the use of multiple g-functions alleviates the attenuation of watermark signals across layers, thereby diminishing the advantage of layer weighting.

\begin{table*}[htbp]
    \centering
    \caption{The comparison of mean and weighted mean watermark detector on WorldCup.}
    \resizebox{\textwidth}{!}{\begin{tabular}{l|cccccccccc}
    \toprule[1.5pt]
    \multicolumn{1}{l}{\multirow{2}{*}{\textbf{Dataset}}} & \multicolumn{10}{c}{\textsc{\textbf{Llama3-8B-Base}}} \\

    & \multicolumn{2}{c}{50 tokens} & \multicolumn{2}{c}{100 tokens} & \multicolumn{2}{c}{150 tokens} & \multicolumn{2}{c}{200 tokens} & \multicolumn{2}{c}{250 tokens} \\
    
    \cmidrule(lr){2-3}
    \cmidrule(lr){4-5} 
    \cmidrule(lr){6-7}
    \cmidrule(lr){8-9}
    \cmidrule(lr){10-11}
    
    + Watermark & Best F1 $\uparrow$ & Bit Acc $\uparrow$ & Best F1 $\uparrow$ & Bit Acc $\uparrow$ & Best F1 $\uparrow$ & Bit Acc $\uparrow$ & Best F1 $\uparrow$ & Bit Acc $\uparrow$ & Best F1 $\uparrow$ & Bit Acc $\uparrow$ \\
    \midrule

    \textsc{\textbf{C4 Dataset}} & - & - & - & - & - & - & - & - & - & - \\

    + BiMark & 1.000 \small{± 0.000} & 0.530 \small{± 0.001} & 1.000 \small{± 0.000} & 0.758 \small{± 0.007} & 1.000 \small{± 0.000} & 0.866 \small{±0.007} & 1.000 \small{± 0.000} & 0.919 \small{± 0.006} & 1.000 \small{± 0.000} & 0.949 \small{± 0.008} \\

    + WorldCup ($k=1$, D1) & 0.931 \small{± 0.039} & 0.509 \small{± 0.009} & 0.980 \small{± 0.010} & 0.736 \small{±0.010} & 0.985 \small{± 0.015} & 0.832 \small{± 0.002} & 0.995 \small{± 0.005} & 0.879 \small{± 0.002} & 0.995 \small{± 0.005} & 0.906 \small{± 0.001} \\

    + WorldCup ($k=1$, D2) & 0.953 \small{± 0.005} & 0.520 \small{± 0.007} & 0.985 \small{± 0.005} & 0.744 \small{± 0.005} & 0.995 \small{± 0.005} & 0.846 \small{± 0.001} & 0.990 \small{± 0.010} & 0.895 \small{± 0.004} & 0.995 \small{± 0.005} & 0.918 \small{± 0.000} \\

    + WorldCup ($k=2$, D1) & 0.985 \small{± 0.015} & 0.697 \small{± 0.004} & 0.995 \small{± 0.005} & 0.869 \small{± 0.001} & 1.000 \small{± 0.000} & 0.930 \small{± 0.002} & 1.000 \small{± 0.000} & 0.957 \small{± 0.003} & 1.000 \small{± 0.000} & 0.974 \small{± 0.001} \\

    + WorldCup ($k=2$, D2) & 0.975 \small{± 0.015} & 0.683 \small{± 0.003} & 1.000 \small{± 0.000} & 0.857 \small{± 0.006} & 1.000 \small{± 0.000} & 0.914 \small{± 0.001} & 1.000 \small{± 0.000} & 0.948 \small{± 0.003} & 1.000 \small{± 0.000} & 0.966 \small{± 0.002} \\

    \midrule

    \textsc{\textbf{OpenGen Dataset}} \\

    + BiMark & 0.995 \small{± 0.005} & 0.538 \small{± 0.008} & 1.000 \small{± 0.000} & 0.765 \small{± 0.004} & 1.000 \small{± 0.000} & 0.774 \small{± 0.006} & 1.000 \small{± 0.000} & 0.927 \small{± 0.006} & 1.000 \small{± 0.000} & 0.955 \small{± 0.003} \\

    + WorldCup ($k=1$, D1) & 0.934 \small{± 0.006} & 0.517 \small{± 0.004} & 0.975 \small{± 0.015} & 0.749 \small{± 0.006} & 1.000 \small{± 0.000} & 0.848 \small{± 0.006} & 1.000 \small{± 0.000} & 0.896 \small{± 0.004} & 1.000 \small{± 0.000} & 0.927 \small{± 0.004} \\

    + WorldCup ($k=1$, D2) & 0.960 \small{± 0.020} & 0.526 \small{± 0.011} & 0.995 \small{± 0.005} & 0.768 \small{± 0.006} & 1.000 \small{± 0.000} & 0.868 \small{± 0.001} & 1.000 \small{± 0.000} & 0.915 \small{± 0.002} & 1.000 \small{± 0.000} & 0.943 \small{± 0.000} \\

    + WorldCup ($k=2$, D1) & 0.975 \small{± 0.005} & 0.702 \small{± 0.005} & 1.000 \small{± 0.000} & 0.871 \small{± 0.005} & 1.000 \small{± 0.000} & 0.930 \small{± 0.004} & 1.000 \small{± 0.000} & 0.959 \small{± 0.001} & 1.000 \small{± 0.000} & 0.972 \small{± 0.006} \\

    + WorldCup ($k=2$, D2) & 0.985 \small{± 0.005} & 0.684 \small{± 0.004} & 1.000 \small{± 0.000} & 0.850 \small{± 0.003} & 1.000 \small{± 0.000} & 0.909 \small{± 0.002} & 1.000 \small{± 0.000} & 0.949 \small{± 0.003} & 1.000 \small{± 0.000} & 0.966 \small{± 0.000} \\
            
    \bottomrule[1.5pt]
    \end{tabular}}
    \label{Table: Weighted Mean Detecor}
\end{table*}

\subsection{WorldCup Sampling Layers} \label{Appendix: WorldCup sampling layers}
To investigate the impact of the number of layers in WorldCup, we conducted experiments with different layer counts $m = 5,10,20,30,50$ on LLaMA3-8B-Base and Gemma2-9B-Base, with results summarized in Table~\ref{Table: Layers Number}. Empirically, a layer count of $m = 30$ yields relatively optimal performance, consistent with findings from SynthID-Text~\citep{dathathri2024scalable}. Lower layer counts generally lead to decreased decoding accuracy, while increasing the number of layers exhibits a diminishing returns effect. The number of layers, however, has no significant impact on perplexity (PPL). Therefore, in this work, we primarily adopt $m = 30$ for our experiments.

\begin{table*}[h]
    \centering
    \caption{The result of Worldcup sampling across different number of layers $m$.}
    \resizebox{\textwidth}{!}{\begin{tabular}{l|cccccccccccccccccccc}
    \toprule[1.5pt]
    \multicolumn{1}{l}{\multirow{2}{*}{\textbf{Model}}} & \multicolumn{19}{c}{\textsc{\textbf{C4 Dataset}}} \\

    & \multicolumn{4}{c}{layers $m=5$} & \multicolumn{4}{c}{layers $m=10$} & \multicolumn{4}{c}{layers $m=20$} & \multicolumn{4}{c}{layers $m=30$} & \multicolumn{4}{c}{layers $m=50$} \\
    
    \cmidrule(lr){2-5}
    \cmidrule(lr){6-9} 
    \cmidrule(lr){10-13}
    \cmidrule(lr){14-17}
    \cmidrule(lr){18-21}
    
    + Watermark & Best F1 $\uparrow$ & Bit Acc $\uparrow$ & PPL $\downarrow$ & Time (s) $\downarrow$ & Best F1 $\uparrow$ & Bit Acc $\uparrow$ & PPL $\downarrow$ & Time (s) $\downarrow$ & Best F1 $\uparrow$ & Bit Acc $\uparrow$ & PPL $\downarrow$ & Time (s) $\downarrow$ & Best F1 $\uparrow$ & Bit Acc $\uparrow$ & PPL $\downarrow$ & Time (s) $\downarrow$ & Best F1 $\uparrow$ & Bit Acc $\uparrow$ & PPL $\downarrow$ & Time (s) $\downarrow$ \\
    \midrule

    \textsc{\textbf{Llama3-8B-Base}} \\
    + WorldCup ($k=1$) & 0.954 & 0.931 & 5.688 & 0.011 & 0.998 & 0.972 & 6.188 & 0.011 & 0.995 & 0.985 & 6.625 & 0.011 & 1.000 & \textbf{0.987} & 6.563 & 0.011 & 0.998 & 0.984 & 7.063 & 0.011 \\
    + WorldCup ($k=2$) & 0.995 & 0.990 & 9.250 & 0.011 & 1.000 & 0.994 & 9.250 & 0.012 & 0.997 & 0.991 & 10.13 & 0.012 & 1.000 & \textbf{0.995} & 10.94 & 0.012 & 0.997 & 0.985 & 16.00 & 0.012 \\
    \midrule

    \textsc{\textbf{Gemma2-9B-Base}} \\
    + WorldCup ($k=1$) & 0.932 & 0.883 & 5.719 & 0.011 & 0.975 & 0.933 & 5.938 & 0.011 & 0.993 & 0.957 & 6.078 & 0.012 & 0.997 & \textbf{0.972} & 6.125 & 0.013 & 0.980 & 0.963 & 6.031 & 0.012 \\
    + WorldCup ($k=2$) & 0.997 & 0.978 & 8.250 & 0.011 & 0.993 & 0.978 & 8.500 & 0.011 & 0.993 & 0.983 & 8.375 & 0.012 & 0.997 & \textbf{0.984} & 9.500 & 0.012 & 1.000 & 0.977 & 10.94 & 0.012 \\
    \bottomrule[1.5pt]
    \end{tabular}}
    \label{Table: Layers Number}
\end{table*}

In addition, we examine the effect of the number of leaves $N$, as reported in Table~\ref{Table: Leaves Number}. The results indicate that setting $N = 2$ provides the best trade-off between message decoding accuracy and text quality.

\begin{table*}[h]
    \centering
    \caption{The result of Worldcup sampling across different number of leaves $N$.}
    \resizebox{\textwidth}{!}{\begin{tabular}{l|cccccccccccccccc}
    \toprule[1.5pt]
    \multicolumn{1}{l}{\multirow{2}{*}{\textbf{Model}}} & \multicolumn{15}{c}{\textsc{\textbf{C4 Dataset}}} \\

    & \multicolumn{4}{c}{leaves $N=1$} & \multicolumn{4}{c}{leaves $N=2$} & \multicolumn{4}{c}{leaves $N=3$} & \multicolumn{4}{c}{leaves $N=4$}\\
    
    \cmidrule(lr){2-5}
    \cmidrule(lr){6-9} 
    \cmidrule(lr){10-13}
    \cmidrule(lr){14-17}

    + Watermark & Best F1 $\uparrow$ & Bit Acc $\uparrow$ & PPL $\downarrow$ & Time (s) $\downarrow$ & Best F1 $\uparrow$ & Bit Acc $\uparrow$ & PPL $\downarrow$ & Time (s) $\downarrow$ & Best F1 $\uparrow$ & Bit Acc $\uparrow$ & PPL $\downarrow$ & Time (s) $\downarrow$ & Best F1 $\uparrow$ & Bit Acc $\uparrow$ & PPL $\downarrow$ & Time (s) $\downarrow$ \\
    \midrule

    \textsc{\textbf{Llama3-8B-Base}} \\
    + WorldCup (16 bit) & 0.670 & 0.495 & 5.500 & 0.011 & 0.998 & 0.985 & 6.719 & 0.011 & 1.000 & 0.998 & 15.38 & 0.011 & 1.000 & 1.000 & 53.75 & 0.012 \\
    + WorldCup (24 bit) & 0.672 & 0.492 & 5.500 & 0.012 & 0.997 & 0.958 & 6.781 & 0.012 & 1.000 & 0.988 & 15.88 & 0.012 & 1.000 & 0.994 & 60.00 & 0.012 \\
    + WorldCup (32 bit) & 0.670 & 0.493 & 5.500 & 0.012 & 0.995 & 0.925 & 6.875 & 0.013 & 1.000 & 0.972 & 15.63 & 0.013 & 1.000 & 0.991 & 61.75 & 0.013 \\
    + WorldCup (48 bit) & 0.668 & 0.469 & 5.469 & 0.014 & 0.990 & 0.875 & 6.547 & 0.013 & 1.000 & 0.933 & 16.25 & 0.014 & 1.000 & 0.968 & 58.00 & 0.014 \\
    \bottomrule[1.5pt]
    \end{tabular}}
    \label{Table: Leaves Number}
\end{table*}

\subsection{Different Activation Function} \label{Appendix: Activation Function}
We evaluate three activation functions in Eq.~\ref{Equation: entropy factor}, namely ReLU, Sigmoid, and Tanh. The results are reported in Table~\ref{Table: Different Activation Function}. Overall, the choice of activation function leads to comparable performance trends, and all variants exhibit a consistent trade-off between bit-level detection accuracy and text quality measured by perplexity. Specifically, ReLU exhibits relatively stronger detection performance due to its linear behavior for positive inputs; however, this gain is accompanied by a noticeable degradation in text quality. In contrast, Tanh and Sigmoid introduce nonlinear saturation, resulting in more conservative modulation and better preservation of generation quality. Since entropy values are non-negative, Tanh effectively normalizes them into the interval $[0,1]$, whereas Sigmoid maps them into a narrower range of $[0.5,1]$, which limits the dynamic range of modulation. Based on these considerations, we adopt the Tanh activation function in our method. Combined with a fixed scaling factor $\alpha=1.2$, this choice provides a balanced trade-off between detection accuracy and text quality, while allowing stable and interpretable control of watermark strength in practice.

\begin{table*}[htbp]
    \centering
    \caption{The comparison of different activation functions on C4 dataset.}
    \resizebox{\textwidth}{!}{\begin{tabular}{l|ccccccccccccccccc}
    \toprule[1.5pt]
    \multicolumn{1}{l}{\multirow{2}{*}{\textbf{Model}}} & \multicolumn{15}{c}{\textsc{\textbf{C4 Dataset}}} \\

    & \multicolumn{3}{c}{16 bits} & \multicolumn{3}{c}{24 bits} & \multicolumn{3}{c}{32 bits} & \multicolumn{3}{c}{48 bits} & \multicolumn{3}{c}{64 bits} \\
    
    \cmidrule(lr){2-4}
    \cmidrule(lr){5-7} 
    \cmidrule(lr){8-10}
    \cmidrule(lr){11-13}
    \cmidrule(lr){14-16}
    
    + Watermark & Best F1 $\uparrow$ & Bit Acc $\uparrow$ & PPL $\downarrow$ & Best F1 $\uparrow$ & Bit Acc $\uparrow$ & PPL $\downarrow$ & Best F1 $\uparrow$ & Bit Acc $\uparrow$ & PPL $\downarrow$ & Best F1 $\uparrow$ & Bit Acc $\uparrow$ & PPL $\downarrow$ & Best F1 $\uparrow$ & Bit Acc $\uparrow$ & PPL $\downarrow$ \\
    \midrule

    \textsc{\textbf{Llama3-8B-Base}} \\
    + WorldCup (ReLU, $\alpha=0.8$) & 1.000 & 0.989 & 10.75 & 1.000 & 0.974 & 10.94 & 1.000 & 0.959 & 10.56 & 0.997 & 0.921 & 10.50 & 0.995 & 0.885 & 11.06 \\
    + WorldCup (ReLU, $\alpha=1$) & 1.000 & 0.991 & 11.44 & 1.000 & 0.987 & 11.72 & 1.000 & 0.971 & 12.00 & 1.000 & 0.936 & 12.19 & 0.997 & 0.899 & 12.38 \\
    + WorldCup (Sigmoid, $\alpha=1$) & 0.997 & 0.988 & 9.719 & 1.000 & 0.970 & 9.813 & 1.000 & 0.956 & 9.500 & 0.990 & 0.913 & 9.719 & 0.997 & 0.870 & 9.500 \\
    + WorldCup (Sigmoid, $\alpha=1.2$) & 1.000 & 0.990 & 11.06 & 0.997 & 0.979 & 11.25 & 1.000 & 0.963 & 11.44 & 1.000 & 0.925 & 11.63 & 1.000 & 0.891 & 11.44 \\
    + WorldCup (Tanh, $\alpha=1$) & 1.000 & 0.984 & 9.500 & 0.995 & 0.966 & 9.813 & 0.995 & 0.951 & 9.625 & 0.992 & 0.914 & 9.875 & 0.997 & 0.871 & 9.625 \\
    + WorldCup (Tanh, $\alpha=1.2$) & 0.997 & 0.991 & 10.84 & 0.995 & 0.976 & 10.75 & 0.995 & 0.956 & 10.25 & 0.997 & 0.918 & 10.94 & 0.995 & 0.889 & 10.66 \\
    \textsc{\textbf{Gemma2-9B-Base}} \\
    + WorldCup (ReLU, $\alpha=0.8$) & 0.997 & 0.960 & 8.375 & 0.993 & 0.930 & 8.500 & 0.990 & 0.911 & 8.188 & 0.985 & 0.871 & 8.375 & 0.983 & 0.822 & 8.000 \\
    + WorldCup (ReLU, $\alpha=1$) & 0.987 & 0.967 & 9.250 & 0.992 & 0.955 & 9.063 & 0.990 & 0.924 & 9.125 & 0.992 & 0.886 & 9.406 & 0.982 & 0.841 & 9.188 \\
    + WorldCup (Sigmoid, $\alpha=1$) & 0.990 & 0.968 & 8.375 & 0.992 & 0.940 & 7.750 & 0.992 & 0.914 & 7.938 & 0.977 & 0.867 & 8.375 & 0.980 & 0.827 & 8.250 \\
    + WorldCup (Sigmoid, $\alpha=1.2$) & 1.000 & 0.968 & 9.406 & 0.993 & 0.957 & 9.313 & 0.992 & 0.924 & 9.313 & 0.990 & 0.879 & 9.813 & 0.987 & 0.835 & 9.625 \\
    + WorldCup (Tanh, $\alpha=1$) & 0.995 & 0.962 & 7.750 & 0.995 & 0.934 & 7.688 & 0.992 & 0.901 & 7.875 & 0.977 & 0.853 & 7.750 & 0.972 & 0.813 & 7.875 \\
    + WorldCup (Tanh, $\alpha=1.2$) & 0.993 & 0.970 & 8.844 & 0.990 & 0.941 & 8.500 & 0.982 & 0.916 & 8.688 & 0.992 & 0.869 & 8.500 & 0.988 & 0.826 & 8.500 \\
    \bottomrule[1.5pt]
    \end{tabular}}
    \label{Table: Different Activation Function}
\end{table*}

\subsection{Key Generation Hyperparameters} \label{Appendix: Key Generation Hyperparameters}

During watermark generation, both the $\textit{no\_repeat\_ngram\_size}$ and the window size significantly affect multi-bit watermark performance, since the random seed for embedding is derived from hashing tokens within a context window. If text diversity is too low (e.g., $\textit{no\_repeat\_ngram\_size} = 0$), identical seeds are repeatedly used, leading to uneven bit allocation across tokens and poor decoding accuracy (Table \ref{Table: No Repeat Ngram Size}). However, increasing diversity does not always help: setting $\textit{no\_repeat\_ngram\_size} = 1$ severely degrades fluency, as reflected by high perplexity. Balancing decoding accuracy and text quality, we set $\textit{no\_repeat\_ngram\_size} = 4$. Moreover, window size also influences robustness by determining the extent of context used in hashing. Empirically, larger windows reduce robustness under attacks (Table~\ref{Table: Llama3 Window Size} and Table~\ref{Table: Gemma2 Window Size}). We therefore use a window size of 2 to achieve a favorable trade-off between robustness and fluency.

\begin{table*}[htbp]
    \centering
    \caption{The effect of different $\text{no\_repeat\_ngram\_size}$ on multi-bit watermarking methods.}
    \resizebox{\textwidth}{!}{\begin{tabular}{l|ccccccccccccccc}
    \toprule[1.5pt]
    \multicolumn{1}{c|}{\multirow{3}{*}{\textbf{Watermark}}} & \multicolumn{15}{c}{\textsc{\textbf{Llama3-8B-Base}}} \\

    & \multicolumn{3}{c}{no\_repeat\_ngram\_size=0} & \multicolumn{3}{c}{no\_repeat\_ngram\_size=1} & \multicolumn{3}{c}{no\_repeat\_ngram\_size=2} & \multicolumn{3}{c}{no\_repeat\_ngram\_size=3} & \multicolumn{3}{c}{no\_repeat\_ngram\_size=4} \\
    
    \cmidrule(lr){2-4}
    \cmidrule(lr){5-7} 
    \cmidrule(lr){8-10}
    \cmidrule(lr){11-13}
    \cmidrule(lr){14-16}
    
    & Diversity $\uparrow$ & Bit Acc $\uparrow$ & Perplexity $\downarrow$ & Diversity $\uparrow$& Bit Acc $\uparrow$ & Perplexity $\downarrow$ & Diversity $\uparrow$& Bit Acc $\uparrow$ & Perplexity $\downarrow$ & Diversity $\uparrow$& Bit Acc $\uparrow$ & Perplexity $\downarrow$ & Diversity $\uparrow$ & Bit Acc $\uparrow$ & Perplexity $\downarrow$ \\
    \midrule

    \faToggleOff BiMark & 5.356 & 0.947 & 6.375 & 20.00 & 1.000 & 45.25 & 18.03 & 0.993 & 15.63 & 8.890 & 0.994 & 12.56 & 7.804 & 0.992 & 11.81 \\

    \faToggleOn WorldCup ($k=1$) & 5.036 & 0.885 & 3.070 & 20.00 & 1.000 & 34.25 & 17.88 & 0.987 & 9.500 & 7.715 & 0.981 & 7.344 & 7.202 & 0.978 & 6.625 \\

    \faToggleOn WorldCup ($k=2$) & 6.110 & 0.947 & 7.750 & 20.00 & 1.000 & 62.75 & 18.92 & 0.996 & 18.88 & 8.794 & 0.996 & 15.38 & 7.929 & 0.994 & 14.25 \\
    
    \bottomrule[1.5pt]
    \end{tabular}}
    \label{Table: No Repeat Ngram Size}
\end{table*}

\begin{table}[H]
    \centering
    \caption{The effect of window size $c$ across our framework WorldCup on LLaMA3-8B-Base.}
    \resizebox{\textwidth}{!}{\begin{tabular}{l|cccccccccccccccc}
    \toprule[1.5pt]
    \large{Watermark} & \multicolumn{16}{c}{\textsc{\textbf{Llama3-8B-Base}}} \\

    & \multicolumn{4}{c}{window\_size\ $c=1$} & \multicolumn{4}{c}{window\_size \ $c=2$} & \multicolumn{4}{c}{window\_size\ $c=3$} & \multicolumn{4}{c}{window\_size\ $c=4$} \\
    
    \cmidrule(lr){2-5}
    \cmidrule(lr){6-9} 
    \cmidrule(lr){10-13}
    \cmidrule(lr){14-17}
    
    \large{+ Attack} & TPR $\uparrow$ & FPR $\downarrow$ & F1 $\uparrow$ & Bit Acc $\uparrow$ & TPR $\uparrow$ & FPR $\downarrow$ & F1 $\uparrow$ & Bit Acc $\uparrow$ & TPR $\uparrow$ & FPR $\downarrow$ & F1 $\uparrow$ & Bit Acc $\uparrow$ & TPR $\uparrow$ & FPR $\downarrow$ & F1 $\uparrow$ & Bit Acc $\uparrow$ \\
    \midrule

    \makecell[l]{\textbf{\textsc{WorldCup ($k=1$)}}} & 0.995 & 0.005 & 0.995 & 0.951 & 1.000 & 0.000 & 1.000 & 0.983 & 0.995 & 0.000 & 0.997 & 0.984 & 0.990 & 0.000 & 0.995 & 0.987 \\

    + Word-D (ratio=0.2) & 0.985 & 0.020 & 0.983 & 0.907 & 0.960 & 0.000 & 0.980 & 0.933 & 0.930 & 0.005 & 0.961 & 0.911 & 0.895 & 0.020 & 0.935 & 0.873 \\

    + Word-S-DICT (ratio=0.2) & 0.965 & 0.025 & 0.970 & 0.895 & 0.930 & 0.005 & 0.961 & 0.908 & 0.960 & 0.065 & 0.948 & 0.868 & 0.845 & 0.070 & 0.883 & 0.842 \\

    + Word-S-BERT (ratio=0.3) & 0.970 & 0.025 & 0.972 & 0.893 & 0.950 & 0.000 & 0.974 & 0.924 & 0.930 & 0.030 & 0.949 & 0.906 & 0.900 & 0.020 & 0.938 & 0.871 \\

    + Copy-Paste (CP-3-30\%) & 0.880 & 0.025 & 0.924 & 0.886 & 0.910 & 0.000 & 0.953 & 0.926 & 0.925 & 0.015 & 0.954 & 0.931 & 0.885 & 0.015 & 0.932 & 0.913 \\

    + Translation (en-zh) & 0.990 & 0.975 & 0.668 & 0.605 & 0.990 & 0.960 & 0.671 & 0.603 & 0.985 & 0.955 & 0.670 & 0.563 & 0.980 & 0.965 & 0.666 & 0.538 \\

    + Rephrase (GPT-4o) & 0.870 & 0.250 & 0.821 & 0.742 & 0.735 & 0.325 & 0.714 & 0.729 & 0.885 & 0.630 & 0.704 & 0.657 & 0.805 & 0.565 & 0.679 & 0.634 \\

    + Dipper-1 (lex=0, order=0) & 0.945 & 0.020 & 0.962 & 0.881 & 0.940 & 0.035 & 0.952 & 0.904 & 0.910 & 0.045 & 0.931 & 0.887 & 0.885 & 0.020 & 0.929 & 0.873 \\

    + Dipper-2 (lex=20, order=20) & 0.895 & 0.110 & 0.893 & 0.793 & 0.865 & 0.180 & 0.846 & 0.772 & 0.765 & 0.130 & 0.807 & 0.738 & 0.770 & 0.295 & 0.746 & 0.698 \\

    \midrule

    \makecell[l]{\textbf{\textsc{WorldCup ($k=2$)}}} & 0.995 & 0.020 & 0.988 & 0.974 & 1.000 & 0.000 & 1.000 & 0.991 & 1.000 & 0.000 & 1.000 & 0.993 & 1.000 & 0.000 & 1.000 & 0.994 \\

    + Word-D (ratio=0.2) & 0.995 & 0.055 & 0.971 & 0.941 & 0.975 & 0.000 & 0.987 & 0.949 & 0.975 & 0.010 & 0.982 & 0.933 & 0.980 & 0.035 & 0.973 & 0.905 \\

    + Word-S-DICT (ratio=0.2) & 0.980 & 0.050 & 0.966 & 0.920 & 0.975 & 0.020 & 0.977 & 0.942 & 0.940 & 0.050 & 0.945 & 0.909 & 0.925 & 0.035 & 0.944 & 0.888 \\

    + Word-S-BERT (ratio=0.3) & 0.995 & 0.055 & 0.971 & 0.923 & 0.980 & 0.035 & 0.973 & 0.940 & 0.945 & 0.045 & 0.950 & 0.922 & 0.975 & 0.060 & 0.958 & 0.897 \\
    
    + Copy-Paste (CP-3-30\%) & 0.920 & 0.055 & 0.932 & 0.909 & 0.925 & 0.045 & 0.939 & 0.940 & 0.950 & 0.010 & 0.969 & 0.955 & 0.900 & 0.010 & 0.942 & 0.936 \\

    + Translation (en-zh) & 0.850 & 0.625 & 0.687 & 0.676 & 0.990 & 0.960 & 0.671 & 0.636 & 0.960 & 0.905 & 0.670 & 0.589 & 0.995 & 0.975 & 0.670 & 0.582 \\

    + Rephrase (GPT-4o) & 0.725 & 0.080 & 0.803 & 0.783 & 0.850 & 0.330 & 0.780 & 0.735 & 0.830 & 0.425 & 0.736 & 0.693 & 0.810 & 0.570 & 0.681 & 0.666 \\

    + Dipper-1 (lex=0, order=0) & 0.970 & 0.060 & 0.956 & 0.890 & 0.970 & 0.035 & 0.968 & 0.911 & 0.945 & 0.055 & 0.945 & 0.903 & 0.860 & 0.035 & 0.908 & 0.883 \\

    + Dipper-2 (lex=20, order=20) & 0.860 & 0.080 & 0.887 & 0.798 & 0.825 & 0.165 & 0.829 & 0.790 & 0.845 & 0.390 & 0.756 & 0.719 & 0.765 & 0.330 & 0.730 & 0.693 \\
    
    \bottomrule[1.5pt]
    \end{tabular}}
    \label{Table: Llama3 Window Size}
\end{table}

\begin{table*}[htbp]
    \centering
    \caption{The effect of window size $c$ across our framework WorldCup on Gemma2-9B-Base.}
    \resizebox{\textwidth}{!}{\begin{tabular}{l|cccccccccccccccc}
    \toprule[1.5pt]
    \large{Watermark} & \multicolumn{16}{c}{\textsc{\textbf{Gemma2-9B-Base}}} \\

    & \multicolumn{4}{c}{window\_size\ $c=1$} & \multicolumn{4}{c}{window\_size \ $c=2$} & \multicolumn{4}{c}{window\_size\ $c=3$} & \multicolumn{4}{c}{window\_size\ $c=4$} \\
    
    \cmidrule(lr){2-5}
    \cmidrule(lr){6-9} 
    \cmidrule(lr){10-13}
    \cmidrule(lr){14-17}
    
    \large{+ Attack} & TPR $\uparrow$ & FPR $\downarrow$ & F1 $\uparrow$ & Bit Acc $\uparrow$ & TPR $\uparrow$ & FPR $\downarrow$ & F1 $\uparrow$ & Bit Acc $\uparrow$ & TPR $\uparrow$ & FPR $\downarrow$ & F1 $\uparrow$ & Bit Acc $\uparrow$ & TPR $\uparrow$ & FPR $\downarrow$ & F1 $\uparrow$ & Bit Acc $\uparrow$ \\
    \midrule

    \makecell[l]{\textbf{\textsc{WorldCup ($k=1$)}}} & 0.995 & 0.015 & 0.990 & 0.909 & 0.995 & 0.010 & 0.993 & 0.943 & 0.995 & 0.000 & 0.997 & 0.973 & 1.000 & 0.005 & 0.998 & 0.974 \\

    + Word-D (ratio=0.2) & 0.955 & 0.030 & 0.962 & 0.872 & 0.875 & 0.085 & 0.893 & 0.868 & 0.860 & 0.050 & 0.901 & 0.848 & 0.830 & 0.105 & 0.858 & 0.813 \\

    + Word-S-DICT (ratio=0.2) & 0.945 & 0.075 & 0.936 & 0.853 & 0.890 & 0.165 & 0.866 & 0.832 & 0.945 & 0.235 & 0.867 & 0.827 & 0.810 & 0.170 & 0.818 & 0.784 \\

    + Word-S-BERT (ratio=0.3) & 0.930 & 0.025 & 0.951 & 0.851 & 0.925 & 0.140 & 0.896 & 0.843 & 0.830 & 0.055 & 0.881 & 0.847 & 0.815 & 0.150 & 0.830 & 0.828 \\

    + Copy-Paste (CP-3-30\%) & 0.875 & 0.030 & 0.919 & 0.836 & 0.880 & 0.030 & 0.921 & 0.874 & 0.885 & 0.025 & 0.927 & 0.879 & 0.885 & 0.040 & 0.919 & 0.884 \\

    + Translation (en-zh) & 0.990 & 0.970 & 0.669 & 0.590 & 1.000 & 0.990 & 0.669 & 0.580 & 0.995 & 0.990 & 0.667 & 0.552 & 0.995 & 1.000 & 0.664 & 0.516 \\

    + Rephrase (GPT-4o) & 0.790 & 0.280 & 0.763 & 0.712 & 0.760 & 0.335 & 0.726 & 0.701 & 0.800 & 0.460 & 0.708 & 0.681 & 0.920 & 0.750 & 0.689 & 0.647 \\

    + Dipper-1 (lex=0, order=0) & 0.925 & 0.070 & 0.927 & 0.781 & 0.825 & 0.030 & 0.889 & 0.795 & 0.819 & 0.055 & 0.874 & 0.817 & 0.760 & 0.070 & 0.831 & 0.798 \\

    + Dipper-2 (lex=20, Dipper=20) & 0.840 & 0.140 & 0.848 & 0.716 & 0.790 & 0.175 & 0.804 & 0.704 & 0.845 & 0.480 & 0.727 & 0.688 & 0.880 & 0.640 & 0.698 & 0.655 \\

    \midrule

    \makecell[l]{\textbf{\textsc{WorldCup ($k=2$)}}} & 0.970 & 0.020 & 0.975 & 0.947 & 0.995 & 0.000 & 0.997 & 0.968 & 0.995 & 0.000 & 0.997 & 0.983 & 0.995 & 0.000 & 0.997 & 0.984 \\

    + Word-D (ratio=0.2) & 0.950 & 0.085 & 0.934 & 0.909 & 0.935 & 0.045 & 0.944 & 0.905 & 0.915 & 0.060 & 0.927 & 0.887 & 0.905 & 0.065 & 0.919 & 0.858 \\

    + Word-S-DICT (ratio=0.2) & 0.905 & 0.060 & 0.921 & 0.877 & 0.845 & 0.045 & 0.894 & 0.882 & 0.860 & 0.060 & 0.896 & 0.869 & 0.855 & 0.110 & 0.870 & 0.824 \\

    + Word-S-BERT (ratio=0.3) & 0.910 & 0.060 & 0.924 & 0.896 & 0.850 & 0.045 & 0.897 & 0.878 & 0.865 & 0.060 & 0.899 & 0.867 & 0.815 & 0.110 & 0.847 & 0.830 \\
    
    + Copy-Paste (CP-3-30\%) & 0.910 & 0.050 & 0.929 & 0.886 & 0.895 & 0.010 & 0.939 & 0.900 & 0.895 & 0.015 & 0.937 & 0.909 & 0.910 & 0.010 & 0.948 & 0.905 \\

    + Translation (en-zh) & 0.995 & 1.000 & 0.664 & 0.641 & 0.995 & 0.995 & 0.666 & 0.602 & 0.990 & 0.970 & 0.669 & 0.588 & 0.975 & 0.945 & 0.668 & 0.537 \\
    
    + Rephrase (GPT-4o) & 0.795 & 0.255 & 0.776 & 0.756 & 0.825 & 0.575 & 0.688 & 0.709 & 0.935 & 0.795 & 0.685 & 0.690 & 0.965 & 0.875 & 0.680 & 0.634 \\

    + Dipper-1 (lex=0, order=0) & 0.905 & 0.165 & 0.874 & 0.805 & 0.825 & 0.045 & 0.882 & 0.825 & 0.795 & 0.065 & 0.855 & 0.803 & 0.785 & 0.095 & 0.835 & 0.775 \\

    + Dipper-2 (lex=20, Dipper=20) & 0.890 & 0.290 & 0.817 & 0.715 & 0.865 & 0.400 & 0.764 & 0.710 & 0.840 & 0.545 & 0.704 & 0.649 & 0.875 & 0.645 & 0.694 & 0.626 \\
    
    \bottomrule[1.5pt]
    \end{tabular}}
    \label{Table: Gemma2 Window Size}
\end{table*}

\subsection{Text Diversity Analysis}
We further compute Log Diversity to assess textual diversity. The results show that WorldCup ($k = 1$) exhibits lower diversity than the unwatermarked baseline and other methods, indicating a higher degree of repetition (Table~\ref{Table: Log Diversity}). This observation helps explain its lower PPL, as the metric can be influenced by repeated high-probability n-grams. However, we emphasize that the generated text under WorldCup ($k = 1$) remains generally fluent and coherent, rather than degenerating into meaningless repetition. Therefore, the improvement in PPL is not merely an artifact, but also partially reflects a shift toward safer, high-probability generations.

\begin{table*}[htbp]
    \centering
    \caption{The diversity analysis of unwatermarked texts and different watermarked texts.}
    \resizebox{\textwidth}{!}{\begin{tabular}{l|cccccccccccccccc}
    \toprule[1.5pt]
    \multicolumn{1}{l}{\multirow{2}{*}{\textbf{Model}}} & \multicolumn{16}{c}{\textsc{\textbf{C4 Dataset}}} \\

    & \multicolumn{4}{c}{16 bits} & \multicolumn{4}{c}{24 bits} & \multicolumn{4}{c}{32 bits} & \multicolumn{4}{c}{48 bits}\\
    
    \cmidrule(lr){2-5}
    \cmidrule(lr){6-9} 
    \cmidrule(lr){10-13}
    \cmidrule(lr){14-17}
    
    + Watermark & Bit Acc $\uparrow$ & PPL $\downarrow$ & STS $\uparrow$ & Diversity $\uparrow$ & Bit Acc $\uparrow$ & PPL $\downarrow$ & STS $\uparrow$ & Diversity $\uparrow$ & Bit Acc $\uparrow$ & PPL $\downarrow$ & STS $\uparrow$ & Diversity $\uparrow$ & Bit Acc $\uparrow$ & PPL $\downarrow$ & STS $\uparrow$ & Diversity $\uparrow$ \\
    \midrule

    \textsc{\textbf{LLaMA3-8B-Base}} \\
    + Human & - & 9.188 & 1.000 & 8.499 & - & 9.188 & 1.000 & 8.499 & - & 9.188 & 1.000 & 8.499 & - & 9.188 & 1.000 & 8.499 \\
    + Unwatermarked & - & 11.25 & 0.443 & 7.729 & - & 11.06 & 0.443 & 7.736 & - & 11.25 & 0.442 & 7.653 & - & 11.34 & 0.437 & 7.730 \\ 
    + BiMark & 0.983 & 13.38 & 0.433 & 7.533 & 0.962 & 13.38 & 0.439 & 7.676 & 0.952 & 13.81 & 0.441 & 7.572 & 0.890 & 13.81 & 0.424 & 7.626\\
    + StealthInk & 0.963 & 13.19 & 0.435 & 8.047 & 0.931 & 12.94 & 0.436 & 8.009 & 0.910 & 13.19 & 0.437 & 7.900 & 0.866 & 13.19 & 0.436 & 7.913 \\
    + WorldCup ($k=1$) & 0.980 & 7.500 & 0.441 & 6.783 & 0.952 & 7.625 & 0.438 & 6.833 & 0.918 & 7.625 & 0.442 & 6.918 & 0.860 & 7.625 & 0.446 & 6.810 \\
    + WorldCup ($k=2$) & 0.988 & 12.56 & 0.437 & 7.647 & 0.971 & 12.19 & 0.441 & 7.796 & 0.959 & 12.38 & 0.442 & 7.629 & 0.925 & 12.37 & 0.439 & 7.736 \\
    
    \bottomrule[1.5pt]
    \end{tabular}}
    \label{Table: Log Diversity}
\end{table*}

\subsection{Counting-based Decoding vs. Confidence-aware Decoding} \label{Appendix: Decoding}

To rigorously demonstrate that the proposed confidence-aware decoding outperforms conventional counting-based decoding, we compare the decoding accuracy of WorldCup under both strategies, as reported in Table~\ref{Table: Different Decoding}. The results clearly show that confidence-aware decoding consistently achieves higher accuracy across different embedded message bit lengths. This improvement fundamentally stems from its ability to mitigate the adverse influence of low-entropy tokens and instead rely more heavily on high-entropy tokens, which provide more reliable statistical evidence for decoding. This observation is also consistent with conclusions drawn in prior zero-bit watermarking studies, such as EWD~\citep{lu-etal-2024-entropy}, further validating the effectiveness of entropy-aware decoding strategies.

\begin{table*}[htbp]
    \centering
    \caption{The comparison of counting-based decoding and confidence-aware decoding.}
    \resizebox{\textwidth}{!}{\begin{tabular}{l|ccccccccccccccc}
    \toprule[1.5pt]
    \multicolumn{1}{l}{\multirow{2}{*}{\textbf{Model}}} & \multicolumn{15}{c}{\textsc{\textbf{C4 Dataset}}} \\

    & \multicolumn{3}{c}{16 bits} & \multicolumn{3}{c}{24 bits} & \multicolumn{3}{c}{32 bits} & \multicolumn{3}{c}{48 bits} & \multicolumn{3}{c}{64 bits}\\
    
    \cmidrule(lr){2-4}
    \cmidrule(lr){5-7} 
    \cmidrule(lr){8-10}
    \cmidrule(lr){11-13}
    \cmidrule(lr){14-16}
    
    + Watermark & Best F1 $\uparrow$ & Bit Acc $\uparrow$ & PPL $\downarrow$ & Best F1 $\uparrow$ & Bit Acc $\uparrow$ & PPL $\downarrow$ & Best F1 $\uparrow$ & Bit Acc $\uparrow$ & PPL $\downarrow$ & Best F1 $\uparrow$ & Bit Acc $\uparrow$ & PPL $\downarrow$ & Best F1 $\uparrow$ & Bit Acc $\uparrow$ & PPL $\downarrow$ \\
    \midrule

    \textsc{\textbf{Llama3-8B-Base}} \\
    + WorldCup ($k=1$, counting-based) & 0.982 & 0.971 & 6.688 & 0.985 & 0.933 & 6.625 & 0.966 & 0.897 & 6.969 & 0.965 & 0.844 & 6.750 & 0.925 & 0.771 & 6.594 \\
    + WorldCup ($k=1$, confidence-aware) & 0.998 & 0.984 & 6.625 & 0.997 & 0.961 & 6.750 & 0.995 & 0.930 & 7.000 & 0.992 & 0.875 & 6.563 & 0.990 & 0.813 & 6.547 \\
    + WorldCup ($k=2$, counting-based) & 0.931 & 0.978 & 11.00 & 0.926 & 0.959 & 11.16 & 0.908 & 0.925 & 11.06 & 0.883 & 0.886 & 11.44 & 0.889 & 0.843 & 11.44 \\
    + WorldCup ($k=2$, confidence-aware) & 1.000 & 0.990 & 11.00 & 1.000 & 0.982 & 11.25 & 1.000 & 0.962 & 11.16 & 1.000 & 0.925 & 11.44 & 1.000 & 0.887 & 11.16 \\
    
    \bottomrule[1.5pt]
    \end{tabular}}
    \label{Table: Different Decoding}
\end{table*}

\begin{table}[H]
    \centering
    \caption{The results on larger LLMs including Mixtral-8x7B-Instruct and LLaMA3-70B-Base.}
    \resizebox{\textwidth}{!}{\begin{tabular}{l|cccccccccccc}
    \toprule[1.5pt]
    \multicolumn{1}{l}{\multirow{2}{*}{\textbf{Model}}} & \multicolumn{12}{c}{\textsc{\textbf{C4 Dataset}}} \\

    & \multicolumn{3}{c}{16 bits} & \multicolumn{3}{c}{24 bits} & \multicolumn{3}{c}{32 bits} & \multicolumn{3}{c}{48 bits}\\
    
    \cmidrule(lr){2-4}
    \cmidrule(lr){5-7} 
    \cmidrule(lr){8-10}
    \cmidrule(lr){11-13}
    
    + Watermark & Best F1 $\uparrow$ & Bit Acc $\uparrow$ & STS $\uparrow$ & Best F1 $\uparrow$ & Bit Acc $\uparrow$ & STS $\uparrow$ & Best F1 $\uparrow$ & Bit Acc $\uparrow$ & STS $\uparrow$ & Best F1 $\uparrow$ & Bit Acc $\uparrow$ & STS $\uparrow$ \\
    \midrule

    \textsc{\textbf{Mixtral-8x7B-IT-v0.1}} \\
    + MPAC & 0.992 & 0.956 & 0.577 & 0.969 & 0.905 & 0.564	& 0.972 & 0.885 & 0.577	& 0.956 & 0.840 & 0.581 \\
    + WorldCup ($k=1$) & 0.990 & 0.957 & 0.599	& 0.987 & 0.933 & 0.593 & 0.985 & 0.891 & 0.583	& 0.985 & 0.846 & 0.590 \\
    + WorldCup ($k=2$) & 0.990 & 0.968 & 0.585 &	0.992 & 0.943 & 0.579 & 0.992 & 0.921 & 0.579	& 0.985 & 0.868 & 0.575 \\

    \midrule

    \textsc{\textbf{Llama3-70B-Base}} \\
    + MPAC & 0.980 & 0.973 & 0.531	& 0.982 & 0.948 & 0.546	& 0.956 & 0.916 & 0.538 & 0.940 & 0.861 & 0.553 \\
    + WorldCup ($k=1$) & 0.987 & 0.971 & 0.554 & 0.987 & 0.944 & 0.574 & 0.982 & 0.910 & 0.566	& 0.977 & 0.866 & 0.572 \\
    + WorldCup ($k=2$) & 0.990 & 0.977 & 0.551	& 0.987 & 0.956 & 0.540 & 0.983 & 0.936 & 0.551	& 0.977 & 0.891 & 0.528 \\
    
    \bottomrule[1.5pt]
    \end{tabular}}
    \label{Table: Larger Models}
\end{table}

\subsection{Larger Models}
We add Semantic Textual Similarity (STS) as an additional semantic quality metric, computed using sentence embeddings between the generated text and its corresponding reference text. We also conduct experiments on larger models (e.g., Mixtral-8x7B-Instruct and LLaMA3-70B-Base). The additional results (Table~\ref{Table: Larger Models}) remain consistent with our main findings: WorldCup maintains strong overall performance, while STS shows trends highly consistent with perplexity, further confirming good semantic preservation. Moreover, we note that entropy is not inherently tied to model size or capability. Experiments on other large models exhibit stable behavior across different entropy settings, demonstrating the scalability and robustness of our framework.

\subsection{Algorithm}

\begin{algorithm}[H]
\caption{WorldCup Message Decoding (Confidence-Aware)}
\label{Algorithm: WorldCup Message Decoding}
\begin{algorithmic}[1]

\STATE \textbf{Input:} generated text sequence $y_{1:T}$, hash function $h$, watermarking key $\xi$, total message length $b$, $2^k$-ary and $g$-value function families $\{\mathbf{g}_j,\mathbf{\bar g}_j\}_{j=1}^k$

\STATE Compute number of message symbols $B \leftarrow b / k$
\STATE Initialize decoded message symbols $\hat{\mathbf{m}} \in \{0,\ldots,2^k-1\}^B$
\STATE Using preceding context and hash function $h$ to compute random seed $r_t$

\STATE Use $(r_t,\xi)$ to map each token $y_t$ to a message position $p_t \in \{0,\ldots,B-1\}$

\STATE Group tokens by message position: $\mathcal{S}_p = \{y_t \mid p_t = p\}$

\FOR{$p = 0\ \text{to}\ B-1$}

    \FOR{$j = 1\ \text{to}\ k$}
        \STATE Compute confidence scores by averaging over the group: 
        
        $s_j^p \leftarrow \frac{1}{m|\mathcal{S}_p|}\sum_{y_t \in \mathcal{S}_p} \sum_{\ell=1}^mg_{\ell}^{(j)}(y_t, r_t), \
        \bar s_j^p \leftarrow \frac{1}{m|\mathcal{S}_p|}\sum_{y_t \in \mathcal{S}_p} \sum_{\ell=1}^m\bar g_{\ell}^{(j)}(y_t, r_t)$
    \ENDFOR

    \STATE Decode the $2^k$-ary message symbol: $\hat m_p \leftarrow \sum_{j=1}^k 2^{k-j}\,
    \mathbb{I}\!\left(s_j^p < \bar s_j^p\right)$
\ENDFOR

\STATE Assemble decoded symbols into message sequence: $\hat{\mathbf{m}} \leftarrow (\hat m_0, \hat m_1, \ldots, \hat m_{B-1})$

\STATE \textbf{Return} decoded message symbols $\hat{\mathbf{m}}$

\end{algorithmic}
\end{algorithm}

%% file: 07_Appendix-Impact_Statement.tex
\section{Impact Statement} \label{Appendix: Impact Statement}
This paper presents WorldCup, a multi-bit watermarking framework for large language models that enables reliable content attribution and integrity verification. By supporting robust message embedding while preserving text quality, WorldCup promotes accountable and transparent LLM deployment. At the same time, we acknowledge potential risks such as privacy concerns or misuse for surveillance and encourage responsible research and governance to ensure watermarking serves transparency and user rights.

%% file: 07_Appendix-Limitations.tex
\section{Limitations} \label{Appendix: Limitations}
While WorldCup achieves strong empirical performance by leveraging token redundancy, several aspects remain to be further explored. First, although multi-bit embedding per token is shown to be effective, we do not provide a formal analysis of its optimality. A theoretical understanding of token-level information capacity could further strengthen the framework. Second, we leave a comprehensive study of unforgeability in the multi-bit setting to future work. Compared to zero-bit schemes, multi-bit watermarking is generally harder to forge, since an attacker must recover both the bit assignment and the decoding rule. Developing a more systematic analysis of such attacks would provide additional insights into robustness.

%% file: 07_Appendix-Case_Study.tex
\section{Case Study}

\definecolor{prompt}{HTML}{71cdfd}
\definecolor{natural}{HTML}{a3db64}
\definecolor{unwatermarked}{HTML}{f0d724}
\definecolor{watermarked}{HTML}{ff6400}

\subsection{Llama3-8B-Base Examples}

\tcbset{colframe = prompt!30, colbacktitle=prompt!30, coltitle=black, colback=prompt!30, fonttitle = \bfseries}
\begin{tcolorbox}[title = {Prompt (C4 Dataset)}]
Applications for entrance in the Fall of 2019 are now open. We will begin to evaluate applications on January 15, 2019. If your application is not
\end{tcolorbox}

\tcbset{colframe = natural!30, colbacktitle=natural!30, coltitle=black, colback=natural!30, fonttitle = \bfseries}
\begin{tcolorbox}[title = {Natural Text (C4 Dataset)}]
received by then, it may not be evaluated. Please note that we do not typically admit for Spring or Summer admission. Offers are usually made between February and April 15. All applicants are considered for department support via research assistantships, teachings assistantships, and merit-based scholarships. Virtually all graduate students are supported for the academic year and in many cases during the summer term as well. Once the Graduate Admissions Committee has made a final decision on your application they will notify you via the email address that is on your application. There is no minimum GRE or GPA requirement. However, to be competitive for assistantships and scholarships, students need a GPA of at least 3.0, a combined verbal and quantitative score on the GRE of at least 300, and an analytical writing score on the GRE of at least 3.5. The average GRE score for students that were accepted into our program for Fall 2018 was 316.
\end{tcolorbox}

\tcbset{colframe = unwatermarked!30, colbacktitle=unwatermarked!30, coltitle=black, colback=unwatermarked!30, fonttitle = \bfseries}
\begin{tcolorbox}[title = {Unwatermarked Text (Llama3-8B-Base)}] 
complete by that date we will not be able to guarantee you an interview. Click here to begin an application. Please visit the requirements tab above for details about how to complete an application. The deadline for submitting applications for Fall, 2019 is March 31, 2019. A maximum of two interviews a week will be held every Tuesday at 1:00 p.m. and 3:00 p.m., and Wednesday at 9:00 a.m. and 1:00p.m. You may not request interview dates. There is no interview in the Spring. Interviewers are assigned by the Office of Admission. You will not be interviewed by anyone that you request for an interview.
\end{tcolorbox}

\tcbset{colframe = watermarked!30, colbacktitle=watermarked!30, coltitle=black, colback=watermarked!30, fonttitle = \bfseries}
\begin{tcolorbox}[title = {Watermarked Text\ (Message: 1101000010110000\ \ Bit Accuracy: 1.0\ \ k=2)}]
complete or received before January 15, you will be placed in the second round of applications. Please be aware there is limited or no amount of spots for second round applications. Please be aware that all our students for the U.S. and Canada are in class by April 1, for non-residents this is 2 weeks later. We do not keep our spots open for the full year. If you cannot get into Canada by April 1 you cannot join our Program. We will not make arrangements for you to join the Program at a later date. For spots that become available, even if we have already started the Program, only students who have the ability to be in Canada by April 13 will be admitted.
\end{tcolorbox}

\tcbset{colframe = prompt!30, colbacktitle=prompt!30, coltitle=black, colback=prompt!30, fonttitle = \bfseries}
\begin{tcolorbox}[title = {Prompt (C4 Dataset)}]
Independent Bank Corp. (nasdaq global select market:INDB), parent of Rockland Trust Company, today announced 2018 fourth quarter net income of
\end{tcolorbox}

\tcbset{colframe = natural!30, colbacktitle=natural!30, coltitle=black, colback=natural!30, fonttitle = \bfseries}
\begin{tcolorbox}[title = {Natural Text (C4 Dataset)}]
\$29.9 million, or \$1.07 per diluted share, compared to net income of \$33.0 million, or \$1.20 per diluted share, reported in the prior quarter of 2018. Excluding merger and acquisition expenses incurred in the fourth and third quarters of 2018 related to the MNB Bancorp (MNB) merger which closed in November 2018 and the Blue Hills Bancorp (BHB) merger announced on September 20, 2018, operating net income was \$35.9 million, or \$1.29 per diluted share during the fourth quarter of 2018 compared to \$34.9 million, or \$1.27 per diluted share during the third quarter of 2018. Full year net income was \$121.6 million, or \$4.40 on a diluted earnings per share basis, an increase of \$34.4 million, or 39.5\%, as compared to the prior year. In addition, full year operating net income was \$129.8.
\end{tcolorbox}

\tcbset{colframe = unwatermarked!30, colbacktitle=unwatermarked!30, coltitle=black, colback=unwatermarked!30, fonttitle = \bfseries}
\begin{tcolorbox}[title = {Unwatermarked Text (Llama3-8B-Base)}] 
\$35.8 million, or \$0.78 per diluted share, an increase of 2.7\% and 3.3\% from the \$34.9 million, or \$1.02 per diluted share, reported for the 2017 fourth quarter. For the 2018 full year, net income increased 20.1\% to \$137.4 million, or \$3.03 per diluted share, compared to \$114.3 million, or \$2.88 per diluted share, for the 2018 year ended December 31, 2017. The 2018 fourth quarters and full year included \$5.6 million in total credit loss expense, compared to the corresponding prior year amounts of \$4.0 million and \$16.6 million. The 2017 fourth quarters and full years included \$3.6 million and \$6.3 million, respectively, in additional expense for the tax effect of deferred tax liabilities associated with the impact of the change in U.S. federal corporate tax rates from 35\% to 21\% as a result of the Tax Cuts and Jobs Act. “During the quarter we completed a successful acquisition of the 17 branches and other deposits from Eastern Bank.
\end{tcolorbox}

\tcbset{colframe = watermarked!30, colbacktitle=watermarked!30, coltitle=black, colback=watermarked!30, fonttitle = \bfseries}
\begin{tcolorbox}[title = {Watermarked Text\ (Message: 001111111010010000101010\ \ Bit Accuracy: 0.75\ \ k=2)}]
\$26.6 million, or \$0.79 per share, compared to \$23.5 million, or \$1.40 per share, in the prior year’s fourth quarter. For the full year 2018, net income was \$101.7 million, or \$3.06 per share, compared with \$85.2 million, or \$5.29 per share, in 2017. The Company’s 2018 performance was impacted by the non-cash charge of \$10.7 million, which was added back to net income in the calculation of earnings per share, to establish a valuation allowance against its deferred tax assets. This charge was partially offset by a \$6.1 million, or \$10.1 million on an after-tax basis, increase in net interest income due to lower interest expense. Pre-tax, pre-provision net revenue increased 15.1\%, driven by growth in loans and deposits, higher non-interest income and lower non-interest expense.
\end{tcolorbox}

\newpage

\subsection{Gemma2-9B-Base Examples}

\tcbset{colframe = prompt!30, colbacktitle=prompt!30, coltitle=black, colback=prompt!30, fonttitle = \bfseries}
\begin{tcolorbox}[title = {Prompt (C4 Dataset)}]
Our no. 1 sneaker of 2014 just got a serious upgrade in the form of the adidas Pure Boost ZG Prime. We first
\end{tcolorbox}

\tcbset{colframe = natural!30, colbacktitle=natural!30, coltitle=black, colback=natural!30, fonttitle = \bfseries}
\begin{tcolorbox}[title = {Natural Text (C4 Dataset)}]
previewed this sneaker in January, and now the next-gen of Pure Boost has arrived. This model mixes a full-length Boost midsole with a layer of EVA on the outsole to produce what adidas calls a \"floating sensation,\" hence the Zero Gravity's; (ZG) part of its name. That's not the only key feature here, as this version of the Pure Boost ZG also features a Primeknit upper for a cozy and adaptive sock-like fit. It's then finished off with leather tabs at the tongue and heel, adding a premium touch to an otherwise sporty silhouette. If for some reason this Primeknit variation isn't quite your cup of tea—or out of your budget—there's also a less expensive mesh-based version dropping this summer. The adidas Pure Boost ZG Prime is now available in black.
\end{tcolorbox}

\tcbset{colframe = unwatermarked!30, colbacktitle=unwatermarked!30, coltitle=black, colback=unwatermarked!30, fonttitle = \bfseries}
\begin{tcolorbox}[title = {Unwatermarked Text (Gemma2-9B-Base)}] 
laid eyes on the Adidas PureBoost ZG Prime at the recent 2015 PGA Show where the model made its debut on the feet of former World No. 1 and 3-time Major champion, Ernie Els. For the 2016 season, Ernie Els has teamed up with Adidas Golf and will don the PureBoost ZG on his feet during the Major season. It's no secret that adidas Pure Boost technology is the best when it comes to running. The adidas PureBoost ZG and PureBoost ZG Primes boast a flexible woven upper that provides a sock-like fit with minimal movement. adidas Pureboost ZG Prime are made with Primeknit upper, featuring woven technology for flexible fit. adidas Pure Boost ZG. Adidas Pure Boost ZG/Primes are engineered with Primeknit upper, featuring flexible material and delivering comfortable fit. adidas has now released the PureBoost Z and ZG Primes. Adidas PureBoost ZG will also launch this winter in six color options. adidas just released the PureBoost Prime, and now the ZG model is set to make its debut in the near future.
\end{tcolorbox}

\tcbset{colframe = watermarked!30, colbacktitle=watermarked!30, coltitle=black, colback=watermarked!30, fonttitle = \bfseries}
\begin{tcolorbox}[title = {Watermarked Text\ (Message: 1101101011101010\ \ Bit Accuracy: 1.0\ \ k=2)}]
saw the Pure Boost ZG back in July, and one of our editors was lucky enough to try them out during the 2015 World Championships in Beijing, China. The adidas Pure Boost Zg Prime is an update to the most popular running shoe of 2004, the Pure Boost Zg. The Zg Prime has a more streamlined upper and a wider base for a more comfortable fit. The new upper is made of 2 layers of a breathable material, with a 3rd layer of support at the medial side of the shoe. The adidas pure boost zg prime is the latest in a long line of running shoes from adidas. It features a Boost midsole and a Primeknit upper. The shoe also features an EVA midsole for cushioning and a Torsion System for stability. The adidas Zg Prime is a running shoe that features an Air Max unit, a Boost midsole, and a Primeknt upper. The shoe is available in two colorways: Black/White/Grey and White/Black/Grey. The black version features a suede upper with a leather overlay, while the white version features a mesh upper with a leather liner. Both shoes feature a Boost midsole with an Air Max unit.
\end{tcolorbox}

\tcbset{colframe = prompt!30, colbacktitle=prompt!30, coltitle=black, colback=prompt!30, fonttitle = \bfseries}
\begin{tcolorbox}[title = {Prompt (C4 Dataset)}]
A child in Florida has become the first person to die of the flu this season, according to state health officials. State epidemiologists say the
\end{tcolorbox}

\tcbset{colframe = natural!30, colbacktitle=natural!30, coltitle=black, colback=natural!30, fonttitle = \bfseries}
\begin{tcolorbox}[title = {Natural Text (C4 Dataset)}]
child had not been vaccinated and was otherwise healthy before getting sick with the flu. The child, who tested positive for influenza B, died sometime during the week of Sept. 30, although privacy concerns prevent officials from saying exactly where, CBS affiliate WTSP-TV reports. Last flu season, 183 children in the U.S. died from flu or flu-related causes. That's the most since the Centers for Disease Control and Prevention (CDC) began keeping these records in 2004. Overall, an estimated 80,000 Americans died from flu last season. CBS News medical contributor Dr. Tara Narula says this latest news should be a wake-up call to parents to get their children vaccinated. "What this is a strong clear message to parents about the importance of vaccination," she told "CBS This Morning." "This vaccine is safe. It is the most effective tool we have. And we know of the pediatric deaths last year, 80 percent were in kids who were unvaccinated.
\end{tcolorbox}

\tcbset{colframe = unwatermarked!30, colbacktitle=unwatermarked!30, coltitle=black, colback=unwatermarked!30, fonttitle = \bfseries}
\begin{tcolorbox}[title = {Unwatermarked Text (Gemma2-9B-Base)}] 
child died from influenza B. That's different from the type of flu A that caused death and illness in children last year. The CDC says 17 children died of the flu in the 2009-10 flu season, 11 of them from the H1N1 strain, or swine flu. Last year 18 children in 10 states died of swine flu, according to the AAP. That's in addition to eight children who died from the flu last year.
\end{tcolorbox}

\tcbset{colframe = watermarked!30, colbacktitle=watermarked!30, coltitle=black, colback=watermarked!30, fonttitle = \bfseries}
\begin{tcolorbox}[title = {Watermarked Text\ (Message: 011001110110011011100110\ \ Bit Accuracy: 0.958\ \ k=2)}]
child died of flu complications, but did not specify the age of the child or where in the state they lived. It's unclear if the child had been vaccinated against the flu. A spokesperson for the Florida Department of Health said the state does not release information about the individual flu cases. The U.S. Centers for Disease Control and Prevention currently reports that the flu has spread to 46 states, and has killed at least 11 people so far this season. According to the CDC, the flu has hit people hardest 50 to 64 years old. The CDC also reports that in the last week, 45 people out of every 100,000 have sought medical treatment for flu-like symptoms. That's up from 26 the week before. Health officials say this year's flu strain is particularly. The CDC says 75 percent of people who die from the flu each year are 65 or older. Some experts say this year'll be flu season. But they say there's still time to get vaccinated against the flu, and they urge people to do so.
\end{tcolorbox}